\documentclass[10pt,twocolumn,letterpaper]{article}

\usepackage[pagenumbers]{cvpr}

\newcommand{\NN}{\mathcal{N}}
\usepackage{bm}
\DeclareMathOperator*{\argmax}{arg\,max}

\makeatletter
\@ifundefined{E}{}{}
\@ifundefined{KL}{}{}
\@ifundefined{Tr}{}{}
\makeatother

\providecommand{\NN}{\mathcal{N}}
\providecommand{\bx}{\mathbf{x}}
\providecommand{\eps}{\bm{\epsilon}}

\usepackage{subcaption}

\makeatletter

\renewcommand\p@subfigure{} 

\DeclareCaptionLabelFormat{subfiglabel}{Figure~#1#2}

\makeatother

\newcommand{\ours}{Ours}
\usepackage[accsupp]{axessibility}  

\definecolor{cvprblue}{rgb}{0.21,0.49,0.74}
\usepackage[table]{xcolor}
\usepackage{contour}
\usepackage{tikz}

\definecolor{softpink}{RGB}{255,182,193}

\contourlength{0.15em}

\newcommand{\pfat}{\contour{black}{\textcolor{softpink}{\textbf{\textsf{p}}}}}

\usepackage[pagebackref,breaklinks,colorlinks,allcolors=cvprblue]{hyperref}

\DeclareRobustCommand{\rchi}{{\mathpalette\irchi\relax}}
\newcommand{\irchi}[2]{\raisebox{\depth}{$#1\chi$}}
\newcommand{\myparagraph}[1]{\vspace{1pt}\noindent{\bf{#1}}}

\usepackage{comicneue}
\newcommand{\comicfont}{\comicneue}

\def\confName{CVPR}
\def\confYear{2026}

\title{It's Never Too Late:\\ Noise Optimization for Collapse Recovery in Trained Diffusion Models}

\author{Anne Harrington$^{1*}$ \quad A. Sophia Koepke$^{1,2,3*}$ \quad  Shyamgopal Karthik$^{2}$\\[0.5ex] 
Trevor Darrell$^{1}$ \quad Alexei A. Efros$^{1}$ \\[0.5ex] 
\small{$^{1}$ UC Berkeley \quad $^{2}$University of Tübingen, Tübingen AI Center  \quad $^{3}$TU Munich, MCML} \quad \small{[$^{*}$Equal contribution]}
}

\begin{document}
\twocolumn[{
\renewcommand\twocolumn[1][]{#1} %

\maketitle
\begin{center}
    \includegraphics[width=\textwidth]{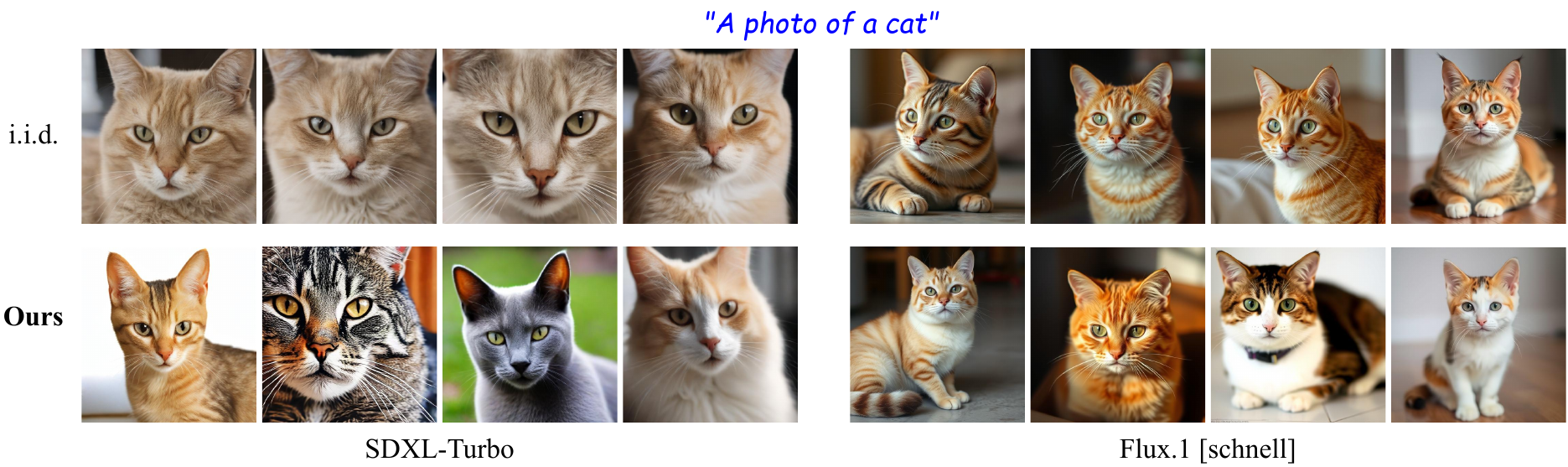}
\vspace{-1.8em}
    \captionof{figure}{Repeatedly sampling from text-to-image models using a fixed text prompt produces surprisingly little visual variation (top row) in both Stable Diffusion SDXL-Turbo~\cite{sdxlturbo} (left) and Flux.1 [schnell]~\cite{flux} (right). Our approach (bottom row) directly optimizes the initial noise to recover from mode collapse, producing diverse outputs.}
    \label{fig:teaser}
    \vspace{0.8em}
\end{center}
}]

\begin{abstract}
Contemporary text-to-image models exhibit a surprising degree of mode collapse, as can be seen when sampling several images given the same text prompt. 
Previous work has attempted to address this issue by steering the model using guidance mechanisms, or by generating a large pool of candidates and refining them. In this work, we take a different direction and aim for diversity in generations via noise optimization. Specifically, we show that a simple noise optimization objective can mitigate mode collapse while preserving the fidelity of the base model. 
We also analyze the frequency characteristics of the noise and show that alternative noise initializations with different frequency profiles can improve both optimization and search.
Our experiments demonstrate that noise optimization yields superior results in terms of generation quality and diversity. Code is available at: \href{https://github.com/anneharrington/divgen}{https://github.com/anneharrington/divgen}
\end{abstract}
\vspace{-1em}
\section{Introduction}

Diffusion models can generate stunning images, yet, when asked to create multiple outputs given a fixed prompt, they often produce nearly identical results over and over again across different random seeds. \Cref{fig:teaser} illustrates this issue, with the top row showing strikingly similar generations (e.g.\ of a cat). For many tasks, we need not only generation quality but also a diversity in outputs that capture the range of possible images per prompt.

At the same time, inference-time scaling has become widespread in diffusion models. The key premise of this line of work is to utilize additional compute during inference to tackle challenging problems which could not otherwise be successfully solved. In the context of diffusion models, inference-time scaling has been used with great success to improve prompt adherence~\cite{doodl,reno,novack2024dittodiffusioninferencetimetoptimization} and personalization~\cite{dreambooth,hyperdreambooth}.

Based on these insights, several inference-time approaches for improving the diversity of images generated with diffusion models have emerged. One popular approach has been to utilize guidance strategies to steer the model towards generating varied samples~\cite{particle,singh2024negative,sadat2023cads}.
Alternatively, generating a large number of candidates and iteratively pruning them to optimize for increasing variety has recently shown success~\cite{gi}. 
This highlights that the initial noise inputs can play a crucial role in obtaining varied sets of generated images, if you are willing to ``roll the dice'' enough times. But what if, instead of just waiting for some random seed to yield a generated image with specific properties, we were able to directly optimize the input noise to satisfy desired properties~\cite{reno}?

In this paper, we design an end-to-end noise optimization strategy to maximize the diversity in sets of generated images, as described in Sec.~\ref{sec:method}. Specifically, we sample a batch of initial noise samples. We then directly optimize these by minimizing a pairwise similarity metric that drives samples apart. Our method outperforms prior works across multiple diffusion models and benchmarks (see Sec.~\ref{sec:experiments}). We demonstrate that we can flexibly select different optimization objectives that facilitate diversity in generated outputs (e.g.\ DINOv2~\cite{dinov2}, LPIPS~\cite{lpips}, DreamSim~\cite{fu2023dreamsim}). Further, we also investigate the usage of set-level diversity objectives such as Determinantal Point Processes (DPP)~\cite{elfeki2019gdpp} and Vendi Score~\cite{friedman2022vendi} and find that they are more suitable to provide increased variation backed by user studies.

In addition, we analyze how the initial noise evolves during optimization and specifically how this impacts different frequency bands (Sec. \ref{subsec:noise}). Inspired by these observations, we explore boosting low-frequency components in the noise initialization, using pink noise, to increase output diversity. Pink noise initializations consistently improve the diversity of generated samples not only for our approach, but also the baselines we compare to for all evaluated models.

Overall, we introduce an end-to-end noise optimization framework that achieves superior output diversity compared to prior methods. Our framework supports the flexible selection of diversity optimization objectives. We find set-level diversity objectives to be most effective and identify low-frequency noise components as key drivers of diverse image generation.

\section{Related Work}
\myparagraph{Inference-Time Scaling.} By allocating additional compute at inference time, test-time scaling enables models to address more challenging problems. Beyond scaling denoising steps in diffusion models, test-time techniques improve generation quality by finding better initial noise or refining intermediate states, often guided by pre-trained reward models. These methods fall into two categories: search-based approaches~\cite{ma2025inference,uehara2025rewardguidediterativerefinementdiffusion,uehara2025inferencetimealignmentdiffusionmodels,imageselect} that evaluate multiple candidates, and optimization-based approaches~\cite{doodl,dflow,novack2024dittodiffusioninferencetimetoptimization,dno,initno,tang2024inference} that iteratively refine noise or latents through gradient descent. In the context of increasing the diversity in the outputs of the generative model, \citet{gi} proposed an efficient search strategy using intermediate generations as a proxy for the final images. Differently, in this work, we demonstrate that an end-to-end noise optimization strategy along with changing the noise initialization achieves superior performance on the quality-diversity tradeoff.

\myparagraph{Guidance Mechanisms.} Drawing from the success of classifier-free guidance (CFG) mechanisms~\cite{cfg,chung2024cfg++} in steering diffusion models towards desired objectives, several variations have been proposed to either improve the effectiveness of CFG~\cite{pag,pag2,kwon2025tcfg}, or reduce its computational complexity~\cite{noiserefine,intervalguidance,autocfg}. To increase the diversity when multiple outputs are sampled, several alternatives have been proposed~\cite{kirchhof2024shielded, singh2024negative}, including the usage of particle guidance~\cite{particle} and DPP~\cite{kulesza2012determinantal,morshed2025diverseflow}. These methods use guidance mechanisms to balance the tradeoff between quality and diversity~\cite{sadat2023cads,sadat2024eliminating,ifriqi2025entropy}. 
Unlike guidance methods that steer the model toward a particular target through modified conditioning, SliderSpace~\cite{gandikota2025sliderspace} encodes semantic directions of variation into LoRA~\cite{lora} weights, enabling user-controlled diversity without inference-time overhead.
Different to those approaches, we directly optimize the initial noise with target diversity and quality objectives.

\myparagraph{Prompt Augmentations.} Improving controllability in generation by modifying the textual conditioning input rather than the diffusion dynamics~\cite{mokady2023null,hertz2022prompt} has also been a popular direction. These methods try to explicitly improve quality and/or diversity using LLMs to rewrite prompts for diffusion models~\cite{manas2024improving,ashutosh2025llms}. Our approach is orthogonal to these methods: while prompt refinements improve the semantic conditioning, some variations in the output space cannot be captured easily by text alone.

\myparagraph{Effect of Initial Noise in Generation.} Several works have explored the controllability of the generation process through initial noise~\cite{initno,sundaram2024cocono,reno,Samuel2023SeedSelect}. It has been observed that specific noise seeds control global behavior~\cite{xu2025good}.
However, the most popular approach is to utilize best-of-n sampling approaches~\cite{dalle, vqvae2,imageselect,pickscore,ma2025inference} or direct noise optimization approaches. In this work, we show that directly optimizing the initial noise can be used as an effective tool to improve the diversity of generations in pre-trained diffusion models. Furthermore, we demonstrate that directly altering the frequency patterns of the initial noise itself impacts the diversity of outputs. This is motivated by an analysis of our noise optimization process and prior work demonstrating that low frequency information at initialization can enhance video diffusion~\cite{wu2024freeinit} and determine object placement in text-to-image models~\cite{ban2024crystal}.

\begin{figure}[t]
    \centering
\includegraphics[width=0.95\linewidth]{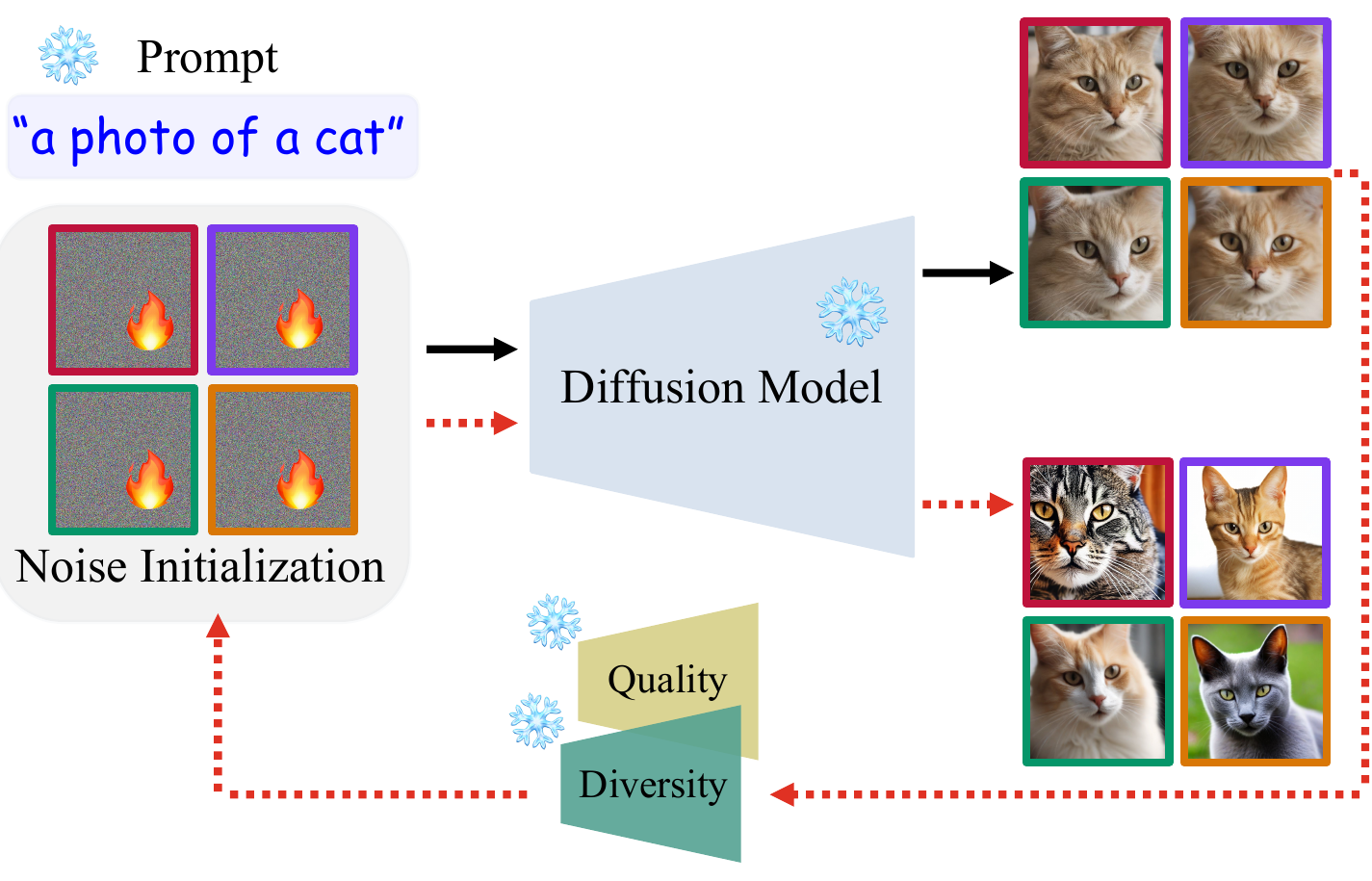}
\vspace{-1em}
\caption{Noise optimization pipeline for diverse image generation. We optimize the noise initialization to increase visual diversity given a fixed text prompt and diffusion model. Starting from $\text{i.i.d.}$\ noise samples, we generate a set of images. Using a diversity objective (e.g.\ DINOv2 dissimilarity) and optionally a quality reward (e.g.\ HPSv2), we update the noise to produce output images that capture more diversity per text prompt. Our method supports optimizing over a variety of objective ensembles.}
    \label{fig:pipeline}
\end{figure}

\section{Collapse Recovery in Diffusion Models}
\label{sec:method}
\myparagraph{Preliminaries.} Diffusion models
interpolate between noise $z = \mathbf{x}_0 \sim \mathcal{N}(0, \mathbf{I})$ and data $\mathbf{x}_1 \sim p_{data}$, such that
$\mathbf{x}_t = \alpha_t \mathbf{x}_0 + \sigma_t \mathbf{x}_1,$
where $\alpha_t$ decreases and $\sigma_t$ increases with $t \in [0,1]$. A generative model $g_\theta(z, c)$ is obtained by simulating a learned differential equation from $\mathbf{x}_0 \sim p_0$, conditioned on a prompt $c$, to produce an image $\mathbf{x}$.

Test-time optimization techniques aim to improve pre-trained generative models on a per-sample basis at inference. A popular gradient-based strategy is test-time noise optimization~\cite{doodl,dflow,novack2024dittodiffusioninferencetimetoptimization,dno,initno,tang2024inference}. Given a pre-trained generator $g_\theta$ (e.g.\ a diffusion or flow matching model), this approach optimizes the initial noise $\mathbf{x}_0$ for each generated instance. The objective is to find an improved $\mathbf{x}_0^\star$ that maximizes a given reward $r(g_\theta(\mathbf{x}_0))$, subject to regularization and can be formulated as
\begin{equation}
\label{eq:noise-opt}
\mathbf{x}_0^\star = \argmax_{\mathbf{x}_0}(r(g_\theta(\mathbf{x}_0)) - \mathrm{reg}(\mathbf{x}_0)),
\end{equation}
where $\mathrm{reg}(\mathbf{x}_0)$ is a regularization term designed to keep $\bx_0^\star$ within a high-density region of the prior noise distribution $p_0$. These methods are designed to improve the quality of a single sample~\cite{reno}, as opposed to our objective of increasing the diversity in multiple generated outputs. We build on this approach for achieving this goal.

\myparagraph{Increasing Diversity through Noise Optimization.}
We propose a noise optimization approach for collapse recovery as shown in Fig.~\ref{fig:pipeline}. We start with a text-to-image diffusion model that takes as input a noise initialization and text prompt. From these inputs, we generate an output set of images. We then compute diversity and quality scores over the images using objective functions such as DINOv2~\cite{dinov2} and image rewards such as HPSv2~\cite{hpsv2}. We use these scores to backpropagate to the initial noise, optimizing for higher output diversity without quality degradation. We iteratively optimize the noise until a user-specified compute budget or diversity/quality threshold is met. Critically, we keep the starting prompt, diffusion model, and objective/reward models frozen and only update the noise. This enables us to increase diversity without altering the model or user input.

Formally, given a prompt $c$, we draw a batch
$\mathcal{B}=\{\mathbf{x}_0^{(i)}\}_{i=1}^{B}$ with
$\mathbf{x}_0^{(i)}\!\sim\!\mathcal{N}(\mathbf{0},\mathbf{I})$ and generate
$\mathbf{x}^{(i)}=g_\theta(\mathbf{x}_0^{(i)},c)$. We optimize the batch
to jointly increase (i) sample-level quality via a reward
$r_s(\mathbf{x}^{(i)},c)$ such as CLIPScore, and (ii) batch-level diversity
via a statistic $v_\mathcal{B}$ computed from pairwise or set-based features
(e.g.~using DINOv2). We minimize
\begin{align}
\label{eq:batch-loss}
\mathcal{L}(\mathcal{B}) &= -\,\frac{\lambda_{q}}{B}\sum_{i=1}^{B} r_s\!\left(\mathbf{x}^{(i)},c\right) \nonumber \\
&\quad - \lambda_{\mathrm{div}}\, v_\mathcal{B}
\;+\; \lambda_{\mathrm{reg}}\,\frac{1}{B}\sum_{i=1}^{B} \mathrm{reg}\!\left(\mathbf{x}_0^{(i)}\right),
\end{align}
where $\lambda_q,\lambda_{\mathrm{div}},\lambda_{\mathrm{reg}}\!\ge\!0$ balance the three terms. The diversity statistic aggregates global feature distances, or patch-level distances for $P$ patches:
\begin{equation}\label{eq:patchwise_distance}
v_\mathcal{B}=\frac{1}{P}\sum_{p=1}^{P}\frac{2}{B(B-1)}\sum_{1\le i<j\le B} d\!\left(f_p(\mathbf{x}^{(i)}),f_p(\mathbf{x}^{(j)})\right),
\end{equation}
with $f_p$ a patch embedding and $d$ a distance metric (e.g.\ cosine distance). Beyond pairwise distances for diversity, we can utilize DPP or Vendi Score on top of these pairwise similarity kernels which provide more meaningful set-level diversity metrics. 
To keep initial noises in high-density regions of the prior we regularize their norm. Writing $\eps^{(i)}\equiv\bx_0^{(i)}$ and $r^{(i)}=\|\eps^{(i)}\|$, the radius $r$ follows a $\smash{\rchi^d}$ law under $\NN(0,\mathbf{I})$. Following \citet{Samuel2023NAO,Samuel2023SeedSelect} and \citet{dflow}, we maximize the log-likelihood of $r$, whose unnormalized log-density is
\begin{equation}
K(\eps)=(d-1)\log\|\eps\|-\tfrac{1}{2}\|\eps\|^2.
\end{equation}
Similar to recent works~\cite{Samuel2023NAO,dflow,reno}, we implement this as a penalty $\mathrm{reg}(\bx_0^{(i)})=-K(\eps^{(i)})$, which encourages $\|\bx_0^{(i)}\|$ to match the $\smash{\rchi^d}$ profile of the Gaussian prior and prevents drift to unlikely radii. We optimize $\{\mathbf{x}_0^{(i)}\}$ by backpropagating through the
frozen sampler $g_\theta$. To control the quality--diversity tradeoff
without adding extra loss terms, we use two threshold-based mechanisms:
optimization stops once the batch diversity reaches a target
$v_\mathcal{B}\!\ge\!\tau_\mathcal{D}$, and individual samples whose
quality reward $r_s$ drops below a threshold $\tau_s$ are reverted to
their last latent state above the threshold (used for Flux.1~[schnell]
in Sec.~\ref{sec:experiments}; see Supplementary Tab.~1 for threshold
values). Optimization also halts once a compute budget is exhausted.

\begin{figure*}[t]
    \centering
    {\textit{\color{blue}{\comicfont \textbf{"A photo of a dog"}}}\\[1ex]}
    \begin{tabular}{p{0.48\textwidth} p{0.48\textwidth}}
        \includegraphics[width=1\linewidth]{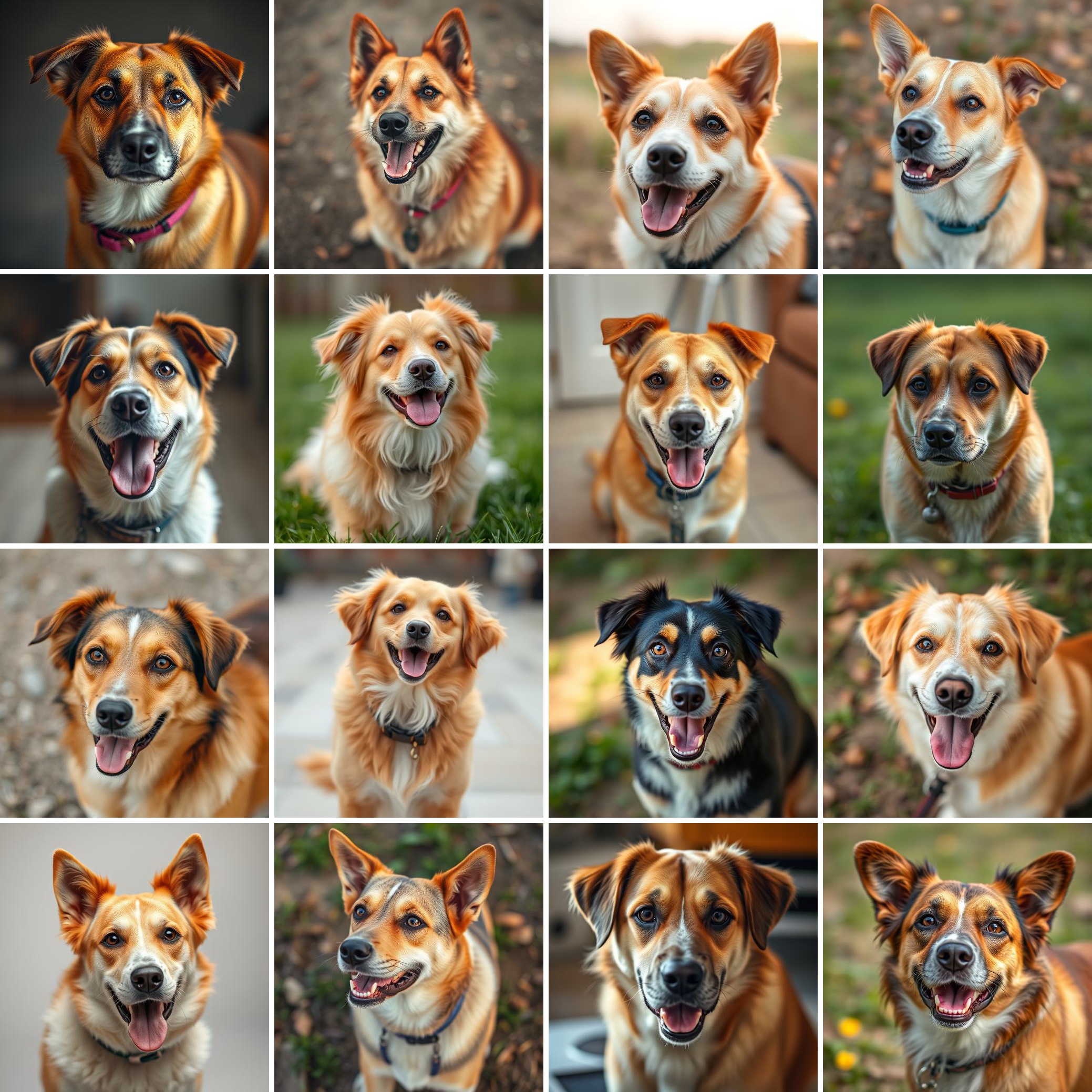} &
        \includegraphics[width=1\linewidth]{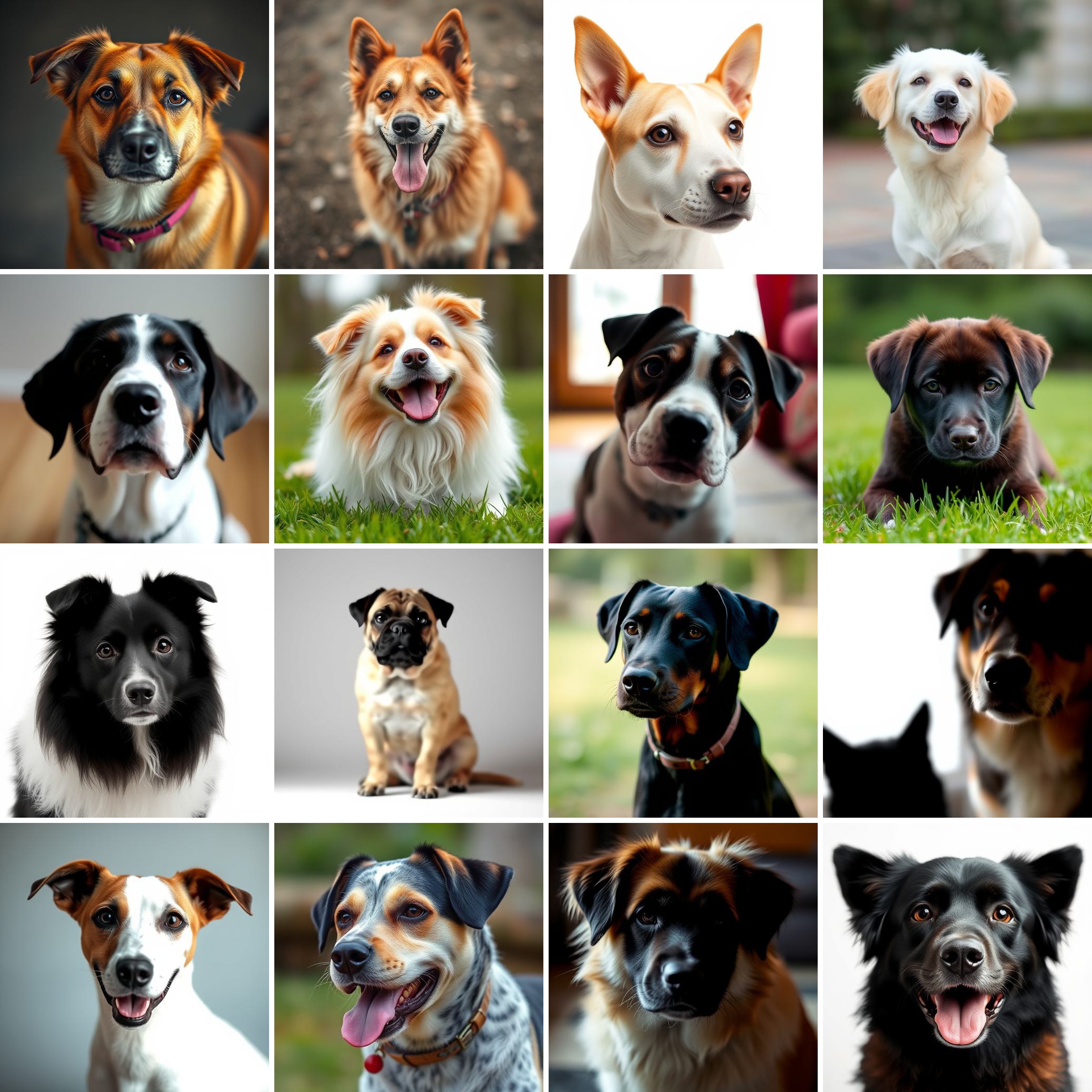}
    \end{tabular}
    \hfill
    \begin{tabular}{p{0.48\textwidth} p{0.48\textwidth}}
        \centering $\text{i.i.d.}$ &
        \centering \textbf{Ours} \\
    \end{tabular}
\vspace{-1.3em}
    \caption{Our method sequentially generates large, diverse image sets. For Flux.1 [schnell], our optimization yields improved diversity of generated image sets (right) compared to $\text{i.i.d.}$ sampling (left). Additional examples are included in the Supplementary (Fig.~4).}
    \label{fig:sequential_generation}
\end{figure*}

\begin{table*}[h!]
\centering

\caption{Output diversity and image-text alignment results on GenEval for our proposed method with SDXL-Turbo using white and pink noise initialization ($\alpha=0.2$). Output diversity for our method and the baselines ($\text{i.i.d.}$ and \cite{gi}) is measured with averaged pairwise DINOv2 \cite{dinov2}, DreamSim \cite{fu2023dreamsim}, and LPIPS \cite{lpips} scores. For our method, we optimize using DINOv2 for diversity and CLIP \cite{clip} for quality.}
\vspace{-0.5em}
\resizebox{\textwidth}{!}{%
\begin{tabular}{lcccc|cccc}
  \toprule
  &  \multicolumn{4}{c}{\textbf{White Noise}} & \multicolumn{4}{c}{\textbf{Pink Noise}} \\
  \cmidrule(lr){2-5} \cmidrule(lr){6-9}
  \textbf{Method} &  DINO & DreamSim & LPIPS & CLIPScore & DINO & DreamSim & LPIPS & CLIPScore \\
  \midrule
 $\text{i.i.d.}$  &  0.588$_{\pm 0.083}$ & 0.249$_{\pm 0.089}$ & 0.642$_{\pm 0.059}$ & 0.335$_{\pm 0.031}$  &  0.642$_{\pm 0.068}$ & 0.305$_{\pm 0.090}$ & 0.729$_{\pm 0.052}$ & 0.328$_{\pm 0.031}$  \\
  \citet{gi}  & 0.705$_{\pm 0.065}$ & 0.331$_{\pm 0.098}$ & 0.682$_{\pm 0.055}$ & 0.333$_{\pm 0.028}$ & 0.749$_{\pm 0.054}$ & 0.392$_{\pm 0.100}$ & 0.757$_{\pm 0.048}$ & 0.323$_{\pm 0.028}$ \\
  \ours\ &  0.784$_{\pm 0.026}$ & 0.411$_{\pm 0.102}$ & 0.767$_{\pm 0.052}$ & 0.349$_{\pm 0.029}$  &  0.786$_{\pm 0.028}$ & 0.427$_{\pm 0.095}$ & 0.811$_{\pm 0.044}$ & 0.341$_{\pm 0.029}$  \\
  \bottomrule
  \bottomrule

\end{tabular}%
}
\vspace{-1em}
\label{tab:main_table}
\end{table*}

\myparagraph{Batched and Sequential Optimization.} 
Our framework supports both batched and sequential optimization. In the batched setting, as in Tab.~\ref{tab:main_table}, we follow \cite{gi} and jointly optimize a set of 4 images per prompt. However, our approach readily scales to much larger diverse sets which is a significant advantage over batch-only methods like \cite{gi}. By generating one image at a time, each image can be optimized to differ from previous outputs. Thus, we avoid the memory overhead of simultaneously processing many candidates, enabling efficient generation of large diverse sets. We show examples of this in \cref{fig:sequential_generation}. Details on hyperparameters can be found in the Supplementary (Sec.~1.2). 

 \begin{figure*}[t]
    \centering
    \includegraphics[width=\textwidth]{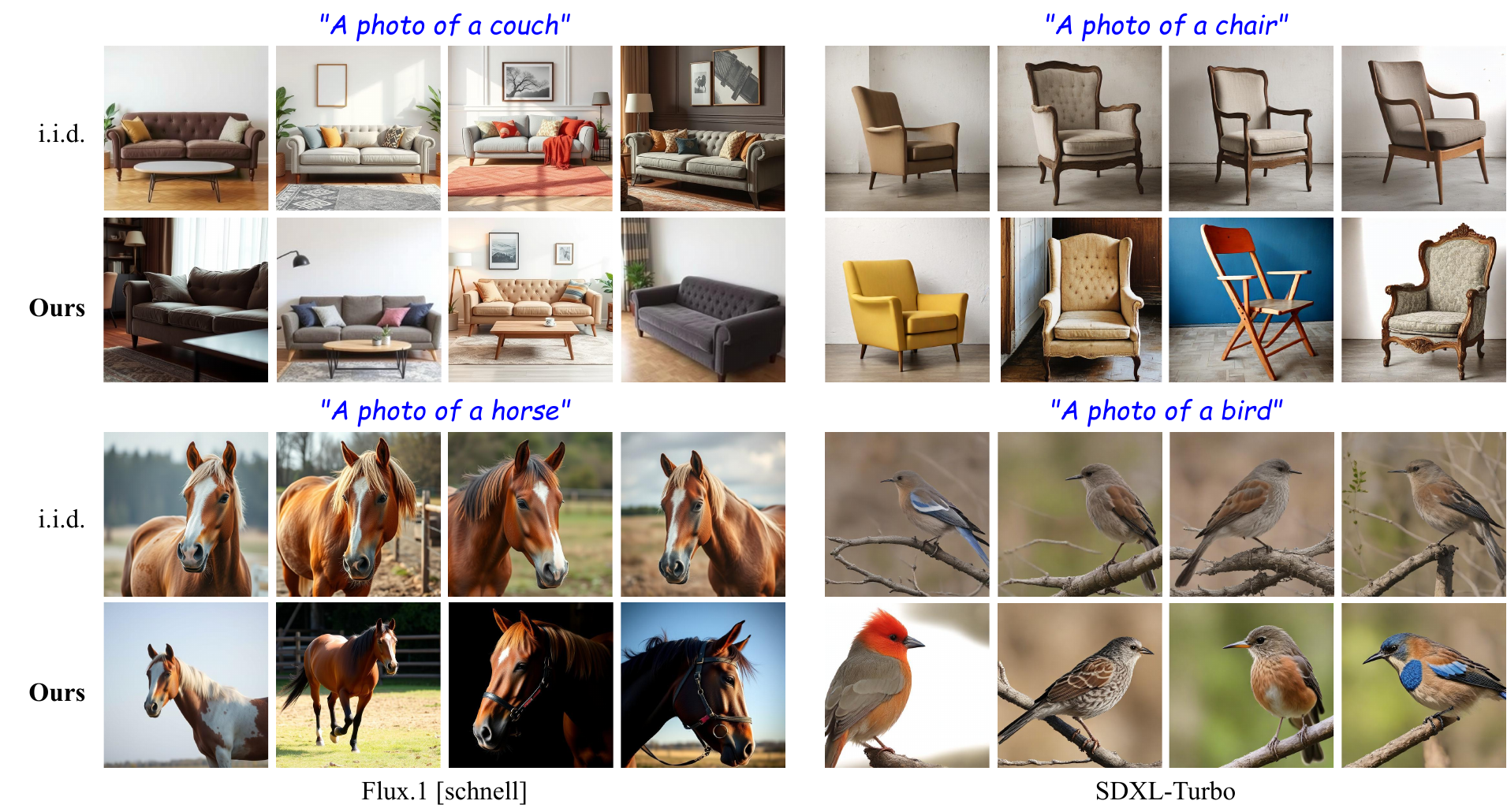}
\vspace{-1.5em}
    \caption{Diverse image generation with Flux.1 [schnell] and SDXL-Turbo using white noise initialization (optimized with DPP~\cite{elfeki2019gdpp} and HPSv2~\cite{hpsv2} objectives). We observe that our method improves diversity in color, orientation, and style compared to $\text{i.i.d.}$ sampling. Additional qualitative results can be found in the Supplementary (Fig. 5).}
    \label{fig:example_generations}
\end{figure*}

\begin{table*}[h]
\centering
\caption{Output diversity (DreamSim, Vendi) and image quality (HPSv2) on GenEval using white and pink ($\alpha=0.2$) noise initialization. We optimize with the DPP diversity objective and HPSv2 quality reward.}
\resizebox{\textwidth}{!}{%
\setlength{\tabcolsep}{12pt}%
\begin{tabular}{lccc|ccc}
\toprule
 & \multicolumn{3}{c}{\textbf{White Noise}} & \multicolumn{3}{c}{\textbf{Pink Noise}} \\
\cmidrule(lr){2-4} \cmidrule(lr){5-7}
Method & DreamSim & Vendi & HPSv2 & DreamSim & Vendi & HPSv2 \\ 
\midrule
\multicolumn{7}{l}{\textbf{SDXL-Turbo}} \\
\midrule
$\text{i.i.d.}$  &
0.262$_{\pm 0.094}$ & 2.000$_{\pm 0.513}$ & 0.284$_{\pm 0.030}$ & 0.296$_{\pm 0.091}$ & 2.136$_{\pm 0.531}$ & 0.285$_{\pm 0.028}$ \\ 
\citet{gi}
& 0.336$_{\pm 0.100}$ & 2.769$_{\pm 0.630}$ & 0.273$_{\pm 0.029}$ & 0.403$_{\pm 0.101}$ & 3.028$_{\pm 0.594}$ & 0.273$_{\pm 0.026}$ \\ 
\ours &
0.457$_{\pm 0.110}$ & 4.000$_{\pm 0.000}$ & 0.292$_{\pm 0.025}$ & 0.474$_{\pm 0.099}$ & 4.000$_{\pm 0.000}$ & 0.288$_{\pm 0.025}$ \\ 
\midrule
\multicolumn{7}{l}{\textbf{Flux.1 [schnell]}} \\
\midrule
$\text{i.i.d.}$ & 0.307$_{\pm 0.100}$ & 2.013$_{\pm 0.490}$ & 0.304$_{\pm 0.025}$ & 0.362$_{\pm 0.100}$ & 2.207$_{\pm 0.502}$ & 0.297$_{\pm 0.023}$  \\
\citet{gi}
& 0.413$_{\pm0.105}$ & 2.473$_{\pm0.554}$ & 0.296$_{\pm0.023}$ & 0.462$_{\pm 0.100}$ & 2.977$_{\pm 0.560}$ & 0.290$_{\pm 0.024}$ \\
\ours &
 0.446$_{\pm 0.116}$ & 2.753$_{\pm 0.587}$ & 0.293$_{\pm 0.025}$ & 0.495$_{\pm 0.097}$ & 3.038$_{\pm 0.523}$ & 0.279$_{\pm 0.024}$ \\
\bottomrule
\bottomrule
\end{tabular}%
}
\label{tab:flux_table}
\vspace{-1em}
\end{table*}

\begin{figure}[t]
    \centering
    \includegraphics[width=\linewidth]{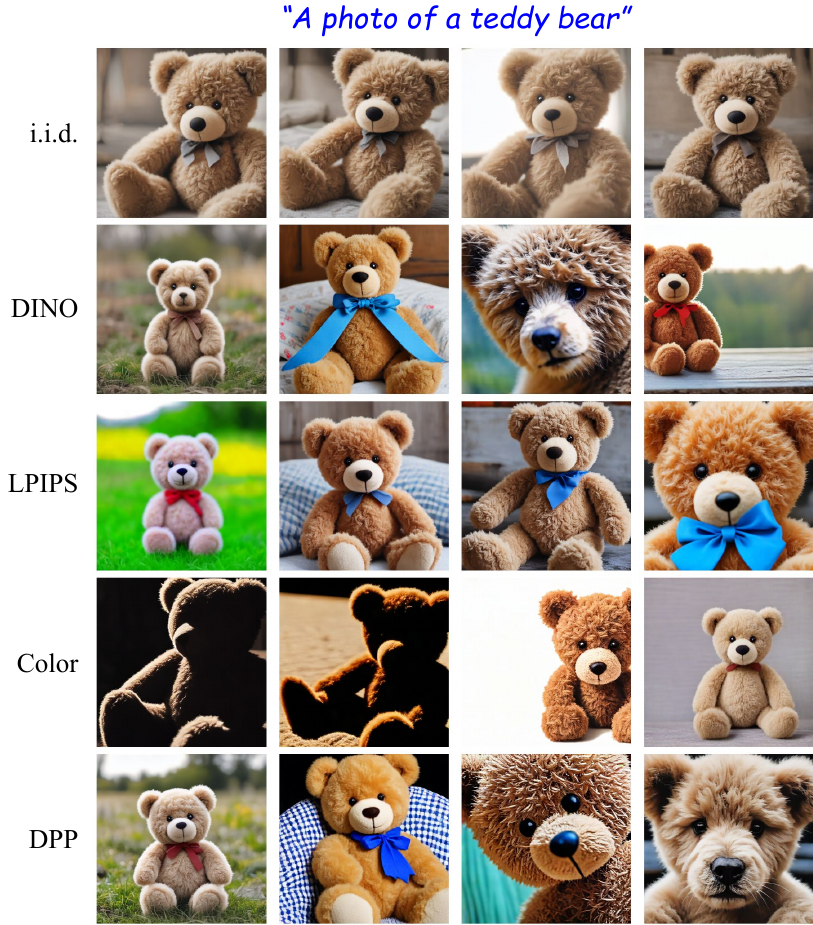}
\vspace{-1em}
    \caption{
    Different optimization objectives produce distinct image sets (SDXL-Turbo). The top row shows $\text{i.i.d.}$ samples, and the rest show our method optimizing with DINOv2~\cite{dinov2},  LPIPS~\cite{lpips}, Color Histogram~\cite{torralba200880}, and DPP~\cite{elfeki2019gdpp} diversity objectives. Quantitative results and additional example generations can be found in Tab.~7 and Fig.~13,14 in the Supplementary.}
    \label{fig:objective}
\end{figure}

\begin{figure}[t]
\centering
\vspace{-1em}
\includegraphics[width=0.95\columnwidth]{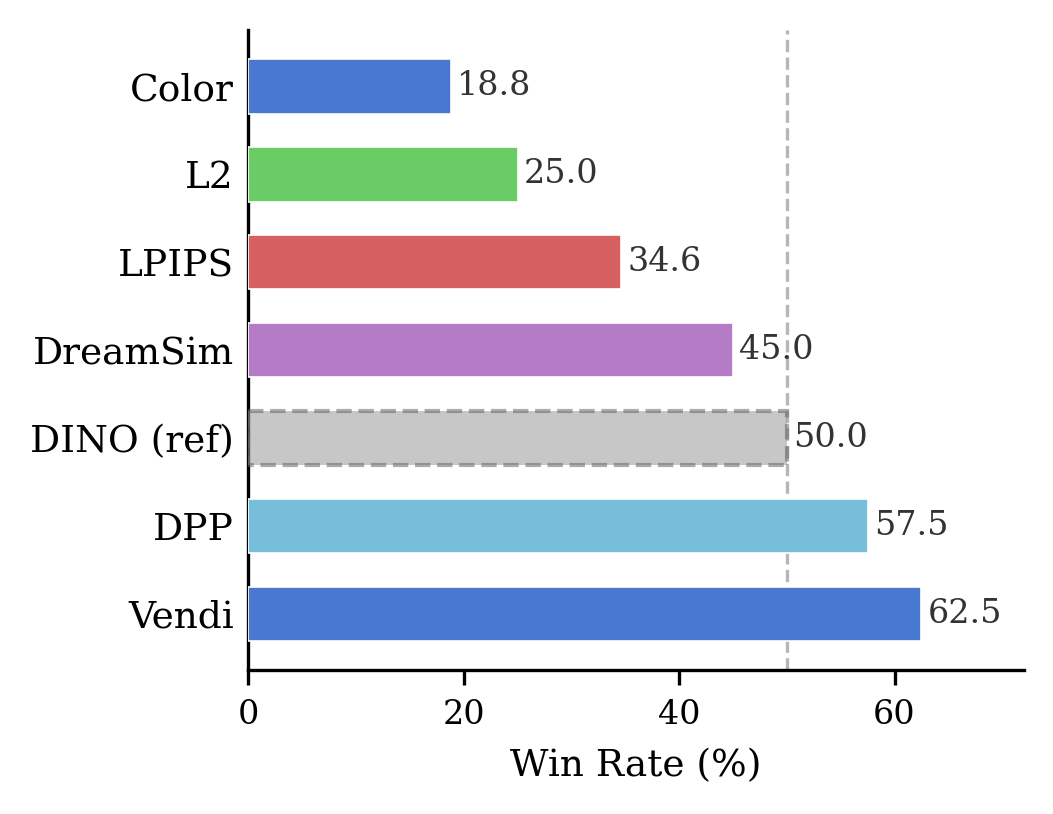}
\vspace{-1em}
\caption{Human preference win rates of our method across diversity objectives for SDXL-Turbo. We use DINOv2 \cite{dinov2} as our reference objective and compare all other metric results against it, including Color Histogram~\cite{torralba200880}, L2 distance, LPIPS~\cite{lpips}, DreamSim~\cite{fu2023dreamsim},  DPP \cite{elfeki2019gdpp}, and Vendi~\cite{friedman2022vendi}. We see that images optimized with set-level objectives (DPP, Vendi) are preferred by users.}
\label{tab:div_win_rate_main}
\vspace{-1em}
\end{figure}

 \begin{figure*}[t]
    \centering
    \includegraphics[width=\textwidth]{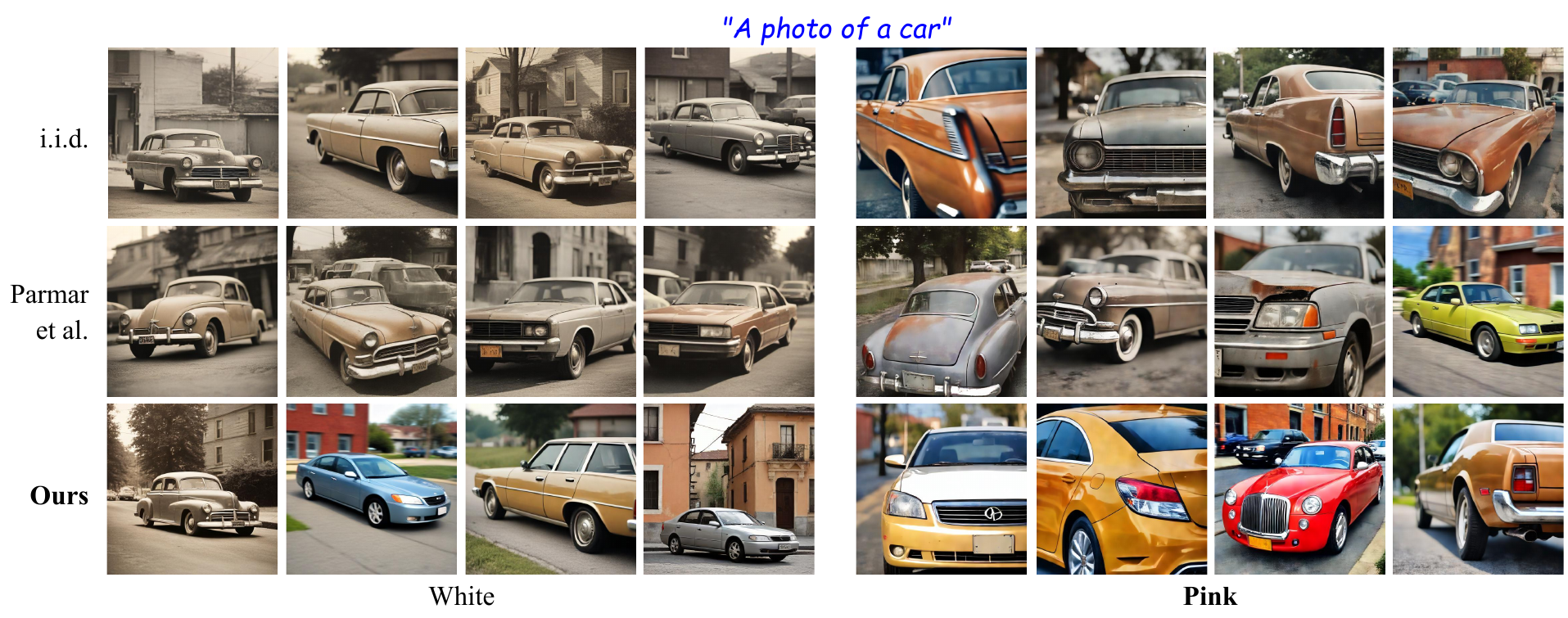}
    \vspace{-1.8em}
    \caption{Diverse image generation with SDXL-Turbo using white and pink noise initialization. Optimizing DPP~\cite{elfeki2019gdpp} and HPSv2~\cite{hpsv2}, we observe that our method improves diversity in color, pose, and style compared to $\text{i.i.d.}$ sampling. In addition, using pink noise simply at inference time without any optimization increases diversity for both $\text{i.i.d.}$ sampling and \cite{gi}. Additional examples can be found in the Supplementary (Fig.~3).}
\vspace{-0.5em}
    \label{fig:example_generations_pink_white}
\end{figure*}

\myparagraph{Sampling Initial Noise.}
Diffusion models commonly initialize the denoising process with white Gaussian noise where the power spectral density is constant across all frequencies. However, natural images have a $1/f$ power spectrum: lower frequencies have more power than higher frequencies ~\cite{field1987relations, simoncelli2001natural, torralba2003statistics}. Motivated by this, we explore alternative noise initialization strategies that align more closely with statistical properties of natural images.

In particular, we consider \textit{pink noise} initialization where we apply spectral filtering in the frequency domain.
For this, $z_{\text{white}} \sim \mathcal{N}(0, \mathbf{I})$ is transformed to the frequency domain using a 2D Fast Fourier Transform (FFT):
\begin{equation}
    \hat{z}^f = \text{FFT2D}(z_{\text{white}}).
\end{equation}
For each frequency component at position $(u, v)$, we compute the radial frequency $f_{u,v} = \sqrt{u^2 + v^2}$.
We then apply power-scaling by reweighing the FFT amplitudes with
$f_{u,v}^{-\alpha/2}$, which yields a $1/f^{\alpha}$ power spectrum:
\begin{equation}
\label{eq:alpha_eq}
\hat{z}^f_{\text{pink}}(u,v) \;=\; \hat{z}^{f}(u,v)\,\cdot\,\frac{1}{f_{u,v}^{\alpha/2}}.
\end{equation}
We then transform this back to the spatial domain by applying an inverse 2D FFT:
\begin{equation}
    \hat{z}_{\text{pink}} = \text{IFFT2D}(\hat{z}^f_{\text{pink}}),
\end{equation}
before normalization to unit variance, $z_{\text{pink}} = \hat{z}_{\text{pink}}/\sigma$,
where $\sigma$ is the empirical standard deviation. Since the IFFT of
the spectrally filtered signal has approximately zero mean, this
recovers the white-noise variance while preserving the $1/f^{\alpha}$
power profile.

\newlength{\imgheight}
\begin{figure*}[t]
\centering
\settoheight{\imgheight}{\includegraphics[width=0.42\textwidth]{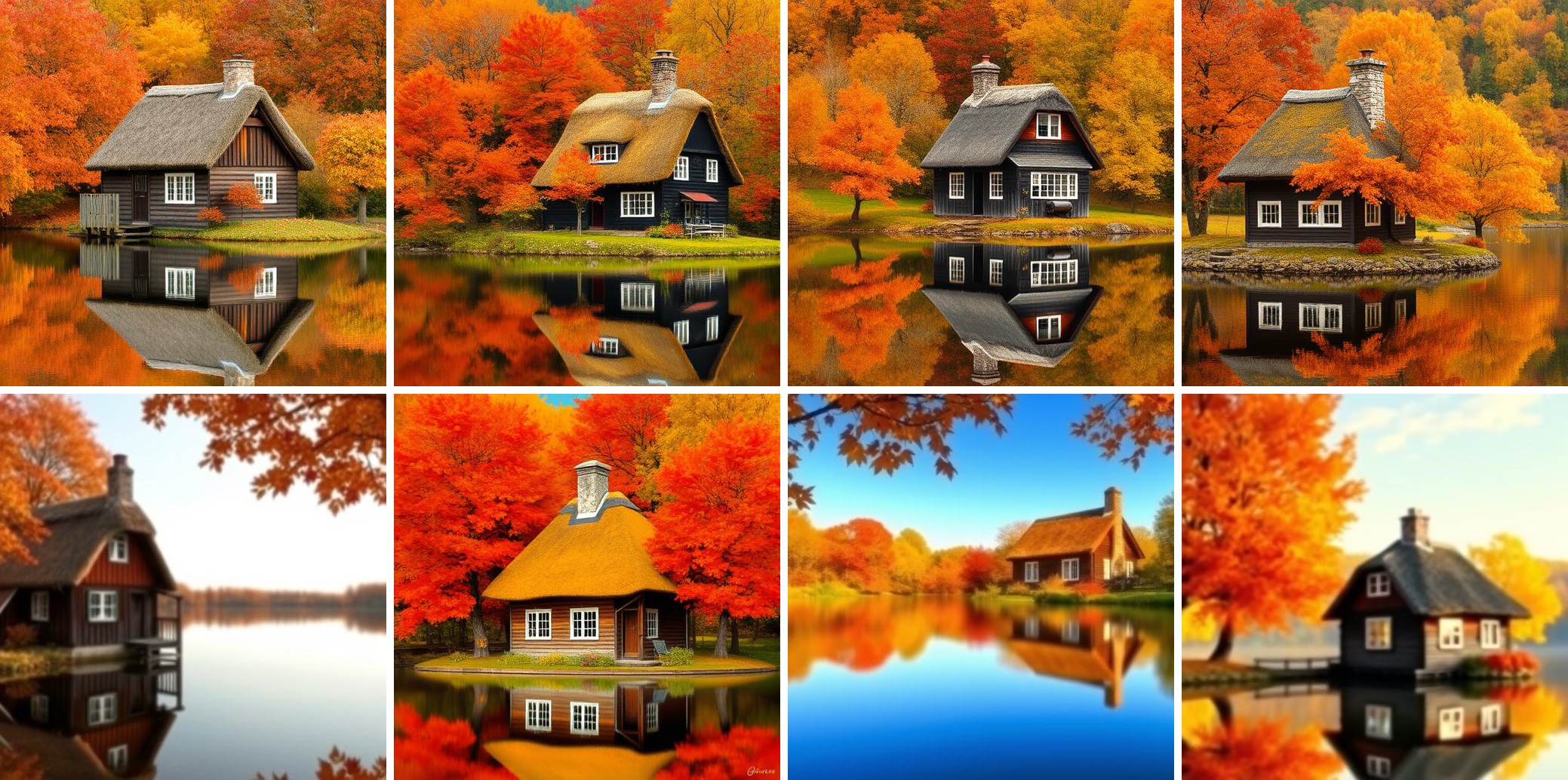}}

\begin{minipage}[c]{0.46\textwidth}
\textit{\small{\textcolor{blue}{\comicfont \textbf{``A picturesque autumn scene where a quaint cottage with a thatched roof sits beside a tranquil lake, surrounded by trees with leaves in vibrant shades of orange, red, and yellow. The cottage's wooden exterior is complemented by white-framed windows, and a stone chimney rises above the roofline. The lake reflects the warm fall colors, creating a mirror image of the foliage and the small structure on its calm surface.''}}}}
\end{minipage}
\hfill
\begin{minipage}[c]{0.04\textwidth}
\centering
\vbox to 0.5\imgheight{\vss \scriptsize $\text{i.i.d.}$\vss}%
\vbox to 0.5\imgheight{\vss \scriptsize \textbf{Ours}\vss}%
\end{minipage}
\hfill
\begin{minipage}[c]{0.42\textwidth}
\includegraphics[width=\linewidth]{figs/comparison_init_ours_00775.jpg}
\end{minipage}
\vspace{-0.5em}
\caption{Our method increases diversity on complex prompts. Optimizing noise for Flux.1 [schnell] on DPG-Bench prompts, we observe that we can increase diversity even when the prompt is highly specified. }
\label{fig:iid_vs_ours}
\vspace{-1em}
\end{figure*}

\begin{figure}[t]
        \centering
  \includegraphics[width=\linewidth, trim={6 6 8 6}, clip]{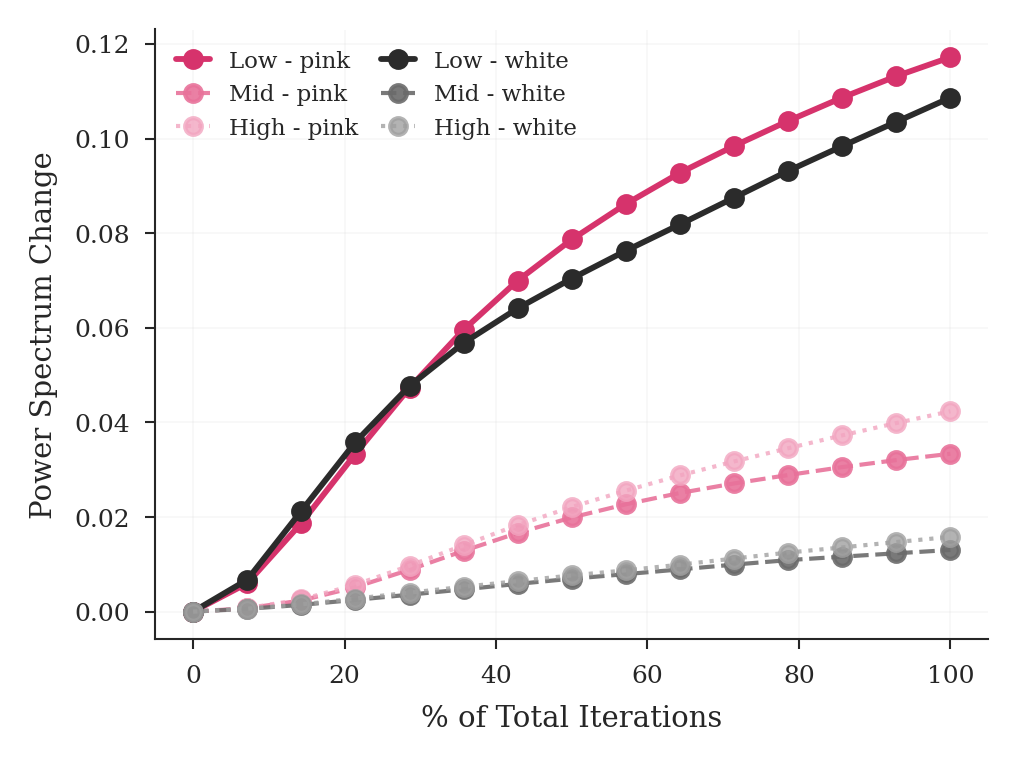}
  \vspace{-1.5em}
    \caption{Noise change across different power spectrum bins throughout optimization iterations with SDXL-Turbo averaged over all prompts of GenEval~\cite{ghosh2023geneval}. While optimizing the starting noise, we observe the largest changes occur in the lowest third of the spectrum (low freq). This trend holds for both white and pink noise initializations.}
        \label{fig:noise_frequency_evolution}
        \vspace{-1em}
\end{figure}

\section{Experiments}
\label{sec:experiments}
For each prompt, we sample a batch of $4$ noise initializations and generate the corresponding $4$ candidate images. We then compute an averaged pairwise diversity objective together with an image-level reward, and use both to optimize the initial noises so that the final set exhibits high visual diversity while preserving image quality (\cref{fig:pipeline}). As diversity objectives, we consider patchwise DINOv2 (\cref{eq:patchwise_distance}), DreamSim~\cite{fu2023dreamsim}, LPIPS~\cite{lpips}, Color histogram distance, and a low-resolution pixel L2 measure that uses 32$\times$32 features, following \cite{torralba200880}. We also evaluate DPP~\cite{kulesza2012determinantal} and Vendi~\cite{friedman2022vendi} scores computed with a DINOv2 [CLS] kernel which has recently been shown to align well with human judgements~\cite{albuquerque2025benchmarking}. To assess image quality and prompt alignment, we report CLIPScore~\cite{hessel2021clipscore,clip} and HPSv2~\cite{hps,hpsv2}, and provide standard deviations across test samples.

Our experiments cover popular step-distilled samplers including SDXL-Turbo~\cite{sdxlturbo} and a larger 10B+ parameter model Flux.1 [schnell]~\cite{flux}. Additional model results for SANA-Sprint~\cite{sanasprint} and PixArt-$\alpha$-DMD~\cite{pixartalpha} can be found in the Supplementary (Tab.~4, 5). These models all rely on ODE-based sampling. While SDE-based sampling could further increase diversity, we focus our work on these distilled samplers. The full noise optimization procedure runs on a single A100 or H100 GPU. Additional details are provided in the Supplementary (Sec.~2).

\myparagraph{Baseline Comparisons.}
We compare our test-time optimization approach to sampling from $\text{i.i.d.}$ noise, and to \cite{gi}, which has been shown to outperform previous guidance-based methods~\cite{particle,sadat2023cads, singh2024negative}. Comparisons to guidance methods can be found in the Supplementary (Tab.~8). Following \cite{gi}, we set the initial set size to $64$ and select $4$ diverse outputs using their default objectives.

We show generation diversity and image-text alignment results for text-to-image generation on GenEval~\cite{ghosh2023geneval} in \cref{tab:main_table}.
Here, we optimize CLIPScore~\cite{hessel2021clipscore} for image-text alignment and pairwise cosine similarity scores with DINOv2 following prior work~\cite{gi}. 
 Our noise optimization demonstrates substantial improvements over $\text{i.i.d.}$ sampled noise initializations and \cite{gi}. Optimizing the noise gives direct control over the quality-diversity tradeoff, allowing us to flexibly balance our objectives or use additional different diversity and image quality optimization objectives. To generate the results in \cref{tab:main_table}, we halted the optimization when reaching preset thresholds (CLIPScore comparable to \cite{gi}, or DINO diversity one standard deviation above \cite{gi}). 
 To evaluate diversity, we include scores for metrics that are not used in optimization and show strong generalization to unused metrics. In addition, our user study (Tab.~9 in the Supplementary) confirms that human judgments of diversity are aligned with the reported metrics.

\myparagraph{Effect of Different Diversity Optimization Objectives.}
We examine how different diversity objectives influence both the variety of generated outputs and the quality of individual samples. Qualitative examples are shown in \cref{fig:objective}. The Supplementary contains quantitative results (Tab.~8) and additional example generations (Fig.~13,14). Using SDXL-Turbo on the GenEval prompts, we compare several objectives that aim to increase visual diversity without sacrificing image-text alignment or image quality. Across all settings, our noise optimization maintains image quality while producing clear gains in visual diversity. Each objective best improves its own metric, but others improve as well, indicating that diversity in one feature space transfers to others and that our optimization increases diversity without harming perceptual quality. Additionally, we conduct a user study that compares the pairwise DINO similarity metric with other diversity metrics in \cref{tab:div_win_rate_main}. We provide details about the user study setup in the Supplementary (Sec.~6). We observe that image sets obtained with Vendi Score~\cite{friedman2022vendi} and DPP~\cite{elfeki2019gdpp} as diversity objectives are preferred by users. The main advantage of these set-level objectives is that they cannot be increased by simply making one single image very different, which would boost the average pairwise score. 

\myparagraph{Quantitative Results.} Informed by our user study and baseline comparison experiments,  we optimize SDXL-Turbo and Flux.1 [schnell] using DPP for diversity and HPSv2~\cite{hpsv2} for quality. We show results for these settings on GenEval in \cref{tab:flux_table}. We observe that we are able to completely saturate Vendi diversity scores to $4.0$ and even slightly improve the HPSv2 quality score. We also see substantial gains on the DreamSim score. See Supplementary Tab.~1 for more details on hyperparameter choice.

For Flux.1 [schnell], we similarly see large gains in Vendi and DreamSim scores compared to $\text{i.i.d.}$ sampling and \cite{gi} (\cref{tab:flux_table}), with a slight drop in the HPSv2 score. We observed that Flux.1 [schnell] is more sensitive to hyperparameter choice than SDXL-Turbo, particularly with balancing the quality reward. 
During optimization, we revert to the last latent when the HPSv2 score drops below a threshold (See Supplementary Tab.~1 for threshold values). 
More dynamic optimization schedules and additional image quality constraints similar to \cite{reno} could be explored in the future to push diversity higher while maintaining image quality.

\myparagraph{Qualitative Generation Examples.}
\cref{fig:example_generations} showcases the effectiveness of our noise optimization approach in generating images with improved variety compared to image sets generated from $\text{i.i.d.}$-sampled noise initializations.
Here, we use the DPP diversity objective (corresponding to \cref{tab:flux_table}). We consistently see increased diversity of object shapes, poses, colors and backgrounds while maintaining alignment to the input prompt.

\myparagraph{Diversity on Complex Prompts.}
Our method also improves diversity on complex prompts. In addition to GenEval, we evaluate our method on DPG-Bench~\cite{hu2024ella} which consists of detailed prompts averaging 67 words. We optimize with Flux.1 [schnell]~\cite{flux_url} using a DPP DINO diversity objective. 
We observe that complex prompts still lead to mode collapse, and that optimizing for diversity is effective (\cref{fig:iid_vs_ours}). Here, our approach also achieves substantial diversity gains (DreamSim, Vendi) while maintaining image quality (Supplementary Tab.~6).

\subsection{Noise Initialization}
\label{subsec:noise}
\myparagraph{Noise Evolution.} We analyze how the optimization modifies the initial noise, which is white Gaussian to start. In particular, we examine changes across frequency bands of the noise power spectrum, shown in \cref{fig:noise_frequency_evolution}. We compute the spectrum via a Fourier Transform on the raw noise latents and track how it evolves over the course of optimization. For interpretability, we divide the spectrum into three equally sized frequency bins and measure the change in each bin relative to the initial noise. 

We observe that the majority of the change occurs in the lowest frequency bin, corresponding to the bottom third of the spectrum. Low-frequency components show noticeably larger shifts than mid- or high-frequency components. This indicates that the optimization primarily acts on the low-frequency structure of the noise, with higher frequencies remaining relatively stable throughout the process. Interestingly, we also observe that the final optimized noise remains Gaussian (i.e.\ follows a standard normal distribution with mean 0 and standard deviation 1), but it is no longer spectrally white due to shifts in the low frequencies.

\myparagraph{Pink Noise Initialization.} As the majority of noise changes across iterations occur in the low-frequency range, we explore pink noise initializations as they are more likely to cover different regions of the noise space in terms of low noise frequencies which appears to be critical for generating diverse images. 
The 1/f frequency distribution inherent in pink noise allocates greater power to lower frequencies, aligning well with the observed optimization dynamics. 
The increased diversity in generated images from pink noise initializations is confirmed by our quantitative results in \cref{tab:main_table} and \cref{tab:flux_table}, and example generations from pink noise in \cref{fig:example_generations_pink_white}. Interestingly, using pink noise initializations also results in higher diversity in output generations for  $\text{i.i.d.}$-sampling and \cite{gi} while only slightly reducing the image-text alignment as measured by HPSv2. See Tab.~1  in the Supplementary for details on optimization hyperparameters. For additional qualitative examples, see Supplementary  Fig.~2 and 3.

\subsection{Scaling Behavior with White and Pink Noise}
\label{subsec:scaling}
A crucial aspect of inference-time scaling is to obtain the best possible improvements for the downstream task given additional compute. We observe that our noise optimization approach can outperform \cite{gi} in terms of image diversity with just a few iterations (Fig.~1 in the Supplementary).  
We optimized the noise for different initializations (i.e.\ $\alpha$ values in \cref{eq:alpha_eq}), using the experimental settings from \cref{tab:main_table}.

With white noise initializations, our approach requires 9 and 12 iterations to reach higher diversity scores than \cite{gi} with an initial pool size of 64 and 128 samples respectively. For $\alpha=0.2$, we require only 6 / 8 iterations to outperform \cite{gi} with initial pool size 64 / 128.
Our approach with pink noise initialization ($\alpha=0.2$) requires 12 / 15 iterations to yield more diverse images than \cite{gi} with similar initialization. For SDXL-Turbo, our approach takes 0.345 s per iteration (with DINO, CLIP objectives) on a single A100 GPU. That means in the pink noise setting, our method takes 2.07 s in total vs.\ 11.20 s for \cite{gi}. See Supplementary Sec.~2 for additional details on computational cost.

Higher $\alpha$ values generally lead to higher diversity scores. However, the image quality decreases with noise exponents $\alpha>0.2$ (see Fig.~11 in the Supplementary).

\section{Conclusion}
In this work, we investigated the critical impact of initial noise on the diversity of diffusion model outputs. We proposed an end-to-end noise optimization approach for maximizing variation across generated samples which allows the flexible selection of diversity optimization objectives. Our noise evolution analysis further inspired a simple yet effective strategy of using pink noise initializations, which consistently enhances the variety of outputs across models and baselines. Our experiments demonstrate that our approach offers a general solution for generating diverse images that significantly outperforms prior methods.

\paragraph{Acknowledgements:}
This work was in part supported by the BMFTR (FKZ: 16IS24060), the DFG (SFB 1233, project number: 276693517), NSF IIS-2403305, and ONR MURI. AH was supported by NDSEG Graduate Fellowship. This research utilized compute resources at the Tübingen Machine Learning Cloud.

{
    \small
    \bibliographystyle{ieeenat_fullname}
    \bibliography{main}

\begin{thebibliography}{72}
\providecommand{\natexlab}[1]{#1}
\providecommand{\url}[1]{\texttt{#1}}
\expandafter\ifx\csname urlstyle\endcsname\relax
  \providecommand{\doi}[1]{doi: #1}\else
  \providecommand{\doi}{doi: \begingroup \urlstyle{rm}\Url}\fi

\bibitem[Ahn et~al.(2024{\natexlab{a}})Ahn, Cho, Min, Jang, Kim, Kim, Park,
  Jin, and Kim]{pag}
Donghoon Ahn, Hyoungwon Cho, Jaewon Min, Wooseok Jang, Jungwoo Kim, SeonHwa
  Kim, Hyun~Hee Park, Kyong~Hwan Jin, and Seungryong Kim.
\newblock Self-rectifying diffusion sampling with perturbed-attention guidance.
\newblock In \emph{ECCV}, 2024{\natexlab{a}}.

\bibitem[Ahn et~al.(2024{\natexlab{b}})Ahn, Kang, Lee, Min, Kim, Jang, Cho,
  Paul, Kim, Cha, et~al.]{noiserefine}
Donghoon Ahn, Jiwon Kang, Sanghyun Lee, Jaewon Min, Minjae Kim, Wooseok Jang,
  Hyoungwon Cho, Sayak Paul, SeonHwa Kim, Eunju Cha, et~al.
\newblock A noise is worth diffusion guidance.
\newblock \emph{arXiv preprint arXiv:2412.03895}, 2024{\natexlab{b}}.

\bibitem[Ahn et~al.(2025)Ahn, Kang, Lee, Kim, Min, Jang, Lee, Paul, Hong, and
  Kim]{pag2}
Donghoon Ahn, Jiwon Kang, Sanghyun Lee, Minjae Kim, Jaewon Min, Wooseok Jang,
  Sangwu Lee, Sayak Paul, Susung Hong, and Seungryong Kim.
\newblock Fine-grained perturbation guidance via attention head selection.
\newblock \emph{arXiv preprint arXiv:2506.10978}, 2025.

\bibitem[Albuquerque et~al.(2025)Albuquerque, Ktena, Wiles, Kaji{\'c},
  Rannen-Triki, Vasconcelos, and Nematzadeh]{albuquerque2025benchmarking}
Isabela Albuquerque, Ira Ktena, Olivia Wiles, Ivana Kaji{\'c}, Amal
  Rannen-Triki, Cristina Vasconcelos, and Aida Nematzadeh.
\newblock Benchmarking diversity in image generation via attribute-conditional
  human evaluation.
\newblock \emph{arXiv preprint arXiv:2511.10547}, 2025.

\bibitem[Ashutosh et~al.(2025)Ashutosh, Gandelsman, Chen, Misra, and
  Girdhar]{ashutosh2025llms}
Kumar Ashutosh, Yossi Gandelsman, Xinlei Chen, Ishan Misra, and Rohit Girdhar.
\newblock Llms can see and hear without any training.
\newblock \emph{arXiv preprint arXiv:2501.18096}, 2025.

\bibitem[Ban et~al.(2024)Ban, Wang, Zhou, Gong, Hsieh, and
  Cheng]{ban2024crystal}
Yuanhao Ban, Ruochen Wang, Tianyi Zhou, Boqing Gong, Cho-Jui Hsieh, and Minhao
  Cheng.
\newblock The crystal ball hypothesis in diffusion models: Anticipating object
  positions from initial noise.
\newblock \emph{arXiv preprint arXiv:2406.01970}, 2024.

\bibitem[Ben-Hamu et~al.(2024)Ben-Hamu, Puny, Gat, Karrer, Singer, and
  Lipman]{dflow}
Heli Ben-Hamu, Omri Puny, Itai Gat, Brian Karrer, Uriel Singer, and Yaron
  Lipman.
\newblock D-flow: Differentiating through flows for controlled generation.
\newblock In \emph{ICML}, 2024.

\bibitem[Bi{\'n}kowski et~al.(2018)Bi{\'n}kowski, Sutherland, Arbel, and
  Gretton]{binkowski2018demystifying}
Miko{\l}aj Bi{\'n}kowski, Danica~J Sutherland, Michael Arbel, and Arthur
  Gretton.
\newblock Demystifying mmd gans.
\newblock \emph{arXiv preprint arXiv:1801.01401}, 2018.

\bibitem[Chen et~al.(2024)Chen, Yu, Ge, Yao, Xie, Wu, Wang, Kwok, Luo, Lu,
  et~al.]{pixartalpha}
Junsong Chen, Jincheng Yu, Chongjian Ge, Lewei Yao, Enze Xie, Yue Wu, Zhongdao
  Wang, James Kwok, Ping Luo, Huchuan Lu, et~al.
\newblock Pixart-alpha: Fast training of diffusion transformer for
  photorealistic text-to-image synthesis.
\newblock In \emph{ICLR}, 2024.

\bibitem[Chen et~al.(2025)Chen, Xue, Zhao, Yu, Paul, Chen, Cai, Xie, and
  Han]{sanasprint}
Junsong Chen, Shuchen Xue, Yuyang Zhao, Jincheng Yu, Sayak Paul, Junyu Chen,
  Han Cai, Enze Xie, and Song Han.
\newblock Sana-sprint: One-step diffusion with continuous-time consistency
  distillation.
\newblock \emph{arXiv preprint arXiv:2503.09641}, 2025.

\bibitem[Chung et~al.(2024)Chung, Kim, Park, Nam, and Ye]{chung2024cfg++}
Hyungjin Chung, Jeongsol Kim, Geon~Yeong Park, Hyelin Nam, and Jong~Chul Ye.
\newblock Cfg++: Manifold-constrained classifier free guidance for diffusion
  models.
\newblock \emph{arXiv preprint arXiv:2406.08070}, 2024.

\bibitem[Corso et~al.(2023)Corso, Xu, De~Bortoli, Barzilay, and
  Jaakkola]{particle}
Gabriele Corso, Yilun Xu, Valentin De~Bortoli, Regina Barzilay, and Tommi
  Jaakkola.
\newblock Particle guidance: non-iid diverse sampling with diffusion models.
\newblock \emph{arXiv preprint arXiv:2310.13102}, 2023.

\bibitem[Elfeki et~al.(2019)Elfeki, Couprie, Riviere, and
  Elhoseiny]{elfeki2019gdpp}
Mohamed Elfeki, Camille Couprie, Morgane Riviere, and Mohamed Elhoseiny.
\newblock Gdpp: Learning diverse generations using determinantal point
  processes.
\newblock In \emph{ICML}, 2019.

\bibitem[Eyring et~al.(2024)Eyring, Karthik, Roth, Dosovitskiy, and
  Akata]{reno}
Luca Eyring, Shyamgopal Karthik, Karsten Roth, Alexey Dosovitskiy, and Zeynep
  Akata.
\newblock Reno: Enhancing one-step text-to-image models through reward-based
  noise optimization.
\newblock \emph{NeurIPS}, 2024.

\bibitem[Field(1987)]{field1987relations}
David~J Field.
\newblock Relations between the statistics of natural images and the response
  properties of cortical cells.
\newblock \emph{Journal of the Optical Society of America A}, 4\penalty0 (12),
  1987.

\bibitem[Friedman and Dieng(2022)]{friedman2022vendi}
Dan Friedman and Adji~Bousso Dieng.
\newblock The vendi score: A diversity evaluation metric for machine learning.
\newblock \emph{arXiv preprint arXiv:2210.02410}, 2022.

\bibitem[Fu et~al.(2023)Fu, Tamir, Sundaram, Chai, Zhang, Dekel, and
  Isola]{fu2023dreamsim}
Stephanie Fu, Netanel Tamir, Shobhita Sundaram, Lucy Chai, Richard Zhang, Tali
  Dekel, and Phillip Isola.
\newblock Dreamsim: Learning new dimensions of human visual similarity using
  synthetic data.
\newblock \emph{arXiv preprint arXiv:2306.09344}, 2023.

\bibitem[Gandikota et~al.(2025)Gandikota, Wu, Zhang, Bau, Shechtman, and
  Kolkin]{gandikota2025sliderspace}
Rohit Gandikota, Zongze Wu, Richard Zhang, David Bau, Eli Shechtman, and Nick
  Kolkin.
\newblock Sliderspace: Decomposing the visual capabilities of diffusion models.
\newblock In \emph{ICCV}, 2025.

\bibitem[Ghosh et~al.(2023)Ghosh, Hajishirzi, and Schmidt]{ghosh2023geneval}
Dhruba Ghosh, Hanna Hajishirzi, and Ludwig Schmidt.
\newblock Geneval: An object-focused framework for evaluating text-to-image
  alignment.
\newblock \emph{NeurIPS}, 2023.

\bibitem[Guo et~al.(2024)Guo, Liu, Cui, Li, Yang, and Huang]{initno}
Xiefan Guo, Jinlin Liu, Miaomiao Cui, Jiankai Li, Hongyu Yang, and Di Huang.
\newblock Initno: Boosting text-to-image diffusion models via initial noise
  optimization.
\newblock In \emph{CVPR}, 2024.

\bibitem[Hertz et~al.(2022)Hertz, Mokady, Tenenbaum, Aberman, Pritch, and
  Cohen-Or]{hertz2022prompt}
Amir Hertz, Ron Mokady, Jay Tenenbaum, Kfir Aberman, Yael Pritch, and Daniel
  Cohen-Or.
\newblock Prompt-to-prompt image editing with cross attention control.
\newblock \emph{arXiv preprint arXiv:2208.01626}, 2022.

\bibitem[Hessel et~al.(2021)Hessel, Holtzman, Forbes, Bras, and
  Choi]{hessel2021clipscore}
Jack Hessel, Ari Holtzman, Maxwell Forbes, Ronan~Le Bras, and Yejin Choi.
\newblock Clipscore: A reference-free evaluation metric for image captioning.
\newblock In \emph{EMNLP}, 2021.

\bibitem[Heusel et~al.(2017)Heusel, Ramsauer, Unterthiner, Nessler, and
  Hochreiter]{heusel2017gans}
Martin Heusel, Hubert Ramsauer, Thomas Unterthiner, Bernhard Nessler, and Sepp
  Hochreiter.
\newblock Gans trained by a two time-scale update rule converge to a local nash
  equilibrium.
\newblock \emph{NeurIPS}, 2017.

\bibitem[Ho and Salimans(2022)]{cfg}
Jonathan Ho and Tim Salimans.
\newblock Classifier-free diffusion guidance.
\newblock \emph{arXiv preprint arXiv:2207.12598}, 2022.

\bibitem[Hu et~al.(2022)Hu, yelong shen, Wallis, Allen-Zhu, Li, Wang, Wang, and
  Chen]{lora}
Edward~J Hu, yelong shen, Phillip Wallis, Zeyuan Allen-Zhu, Yuanzhi Li, Shean
  Wang, Lu Wang, and Weizhu Chen.
\newblock Lo{RA}: Low-rank adaptation of large language models.
\newblock In \emph{ICLR}, 2022.

\bibitem[Hu et~al.(2024)Hu, Wang, Fang, Fu, Cheng, and Yu]{hu2024ella}
Xiwei Hu, Rui Wang, Yixiao Fang, Bin Fu, Pei Cheng, and Gang Yu.
\newblock Ella: Equip diffusion models with llm for enhanced semantic
  alignment.
\newblock \emph{arXiv preprint arXiv:2403.05135}, 2024.

\bibitem[Huang et~al.(2023)Huang, Sun, Xie, Li, and Liu]{huang2023t2icompbench}
Kaiyi Huang, Kaiyue Sun, Enze Xie, Zhenguo Li, and Xihui Liu.
\newblock T2i-compbench: A comprehensive benchmark for open-world compositional
  text-to-image generation.
\newblock \emph{NeurIPS}, 2023.

\bibitem[Ifriqi et~al.(2025)Ifriqi, Romero-Soriano, Drozdzal, Verbeek, and
  Alahari]{ifriqi2025entropy}
Tariq~Berrada Ifriqi, Adriana Romero-Soriano, Michal Drozdzal, Jakob Verbeek,
  and Karteek Alahari.
\newblock Entropy rectifying guidance for diffusion and flow models.
\newblock \emph{NeurIPS}, 2025.

\bibitem[Karras et~al.(2024)Karras, Aittala, Kynk{\"a}{\"a}nniemi, Lehtinen,
  Aila, and Laine]{autocfg}
Tero Karras, Miika Aittala, Tuomas Kynk{\"a}{\"a}nniemi, Jaakko Lehtinen, Timo
  Aila, and Samuli Laine.
\newblock Guiding a diffusion model with a bad version of itself.
\newblock \emph{NeurIPS}, 2024.

\bibitem[Karthik et~al.(2023)Karthik, Roth, Mancini, and Akata]{imageselect}
Shyamgopal Karthik, Karsten Roth, Massimiliano Mancini, and Zeynep Akata.
\newblock If at first you don't succeed, try, try again: Faithful
  diffusion-based text-to-image generation by selection.
\newblock \emph{arXiv preprint arXiv:2305.13308}, 2023.

\bibitem[Karunratanakul et~al.(2024)Karunratanakul, Preechakul, Aksan, Beeler,
  Suwajanakorn, and Tang]{dno}
Korrawe Karunratanakul, Konpat Preechakul, Emre Aksan, Thabo Beeler, Supasorn
  Suwajanakorn, and Siyu Tang.
\newblock Optimizing diffusion noise can serve as universal motion priors.
\newblock In \emph{CVPR}, 2024.

\bibitem[Kirchhof et~al.(2024)Kirchhof, Thornton, B{\'e}thune, Ablin, Ndiaye,
  and Cuturi]{kirchhof2024shielded}
Michael Kirchhof, James Thornton, Louis B{\'e}thune, Pierre Ablin, Eugene
  Ndiaye, and Marco Cuturi.
\newblock Shielded diffusion: Generating novel and diverse images using sparse
  repellency.
\newblock \emph{arXiv preprint arXiv:2410.06025}, 2024.

\bibitem[Kirstain et~al.(2023)Kirstain, Polyak, Singer, Matiana, Penna, and
  Levy]{pickscore}
Yuval Kirstain, Adam Polyak, Uriel Singer, Shahbuland Matiana, Joe Penna, and
  Omer Levy.
\newblock Pick-a-pic: An open dataset of user preferences for text-to-image
  generation.
\newblock \emph{NeurIPS}, 2023.

\bibitem[Kulesza et~al.(2012)Kulesza, Taskar, et~al.]{kulesza2012determinantal}
Alex Kulesza, Ben Taskar, et~al.
\newblock Determinantal point processes for machine learning.
\newblock \emph{Foundations and Trends in Machine Learning}, 2012.

\bibitem[Kwon et~al.(2025)Kwon, Jeong, Hsiao, Uh, et~al.]{kwon2025tcfg}
Mingi Kwon, Jaeseok Jeong, Yi~Ting Hsiao, Youngjung Uh, et~al.
\newblock Tcfg: Tangential damping classifier-free guidance.
\newblock In \emph{CVPR}, 2025.

\bibitem[Kynk{\"a}{\"a}nniemi et~al.(2024)Kynk{\"a}{\"a}nniemi, Aittala,
  Karras, Laine, Aila, and Lehtinen]{intervalguidance}
Tuomas Kynk{\"a}{\"a}nniemi, Miika Aittala, Tero Karras, Samuli Laine, Timo
  Aila, and Jaakko Lehtinen.
\newblock Applying guidance in a limited interval improves sample and
  distribution quality in diffusion models.
\newblock \emph{NeurIPS}, 2024.

\bibitem[Labs(2024)]{flux_url}
Black~Forest Labs.
\newblock Flux.
\newblock \url{https://github.com/black-forest-labs/flux}, 2024.

\bibitem[Labs(2025)]{flux}
Black~Forest Labs.
\newblock Flux.1 kontext: Flow matching for in-context image generation and
  editing in latent space.
\newblock \emph{arXiv preprint arXiv:2506.15742}, 2025.

\bibitem[Lin et~al.(2014)Lin, Maire, Belongie, Hays, Perona, Ramanan,
  Doll{\'a}r, and Zitnick]{mscoco}
Tsung-Yi Lin, Michael Maire, Serge Belongie, James Hays, Pietro Perona, Deva
  Ramanan, Piotr Doll{\'a}r, and C~Lawrence Zitnick.
\newblock Microsoft coco: Common objects in context.
\newblock In \emph{ECCV}, 2014.

\bibitem[Ma et~al.(2025)Ma, Tong, Jia, Hu, Su, Zhang, Yang, Li, Jaakkola, Jia,
  et~al.]{ma2025inference}
Nanye Ma, Shangyuan Tong, Haolin Jia, Hexiang Hu, Yu-Chuan Su, Mingda Zhang,
  Xuan Yang, Yandong Li, Tommi Jaakkola, Xuhui Jia, et~al.
\newblock Inference-time scaling for diffusion models beyond scaling denoising
  steps.
\newblock \emph{arXiv preprint arXiv:2501.09732}, 2025.

\bibitem[Ma{\~n}as et~al.(2024)Ma{\~n}as, Astolfi, Hall, Ross, Urbanek,
  Williams, Agrawal, Romero-Soriano, and Drozdzal]{manas2024improving}
Oscar Ma{\~n}as, Pietro Astolfi, Melissa Hall, Candace Ross, Jack Urbanek,
  Adina Williams, Aishwarya Agrawal, Adriana Romero-Soriano, and Michal
  Drozdzal.
\newblock Improving text-to-image consistency via automatic prompt
  optimization.
\newblock \emph{arXiv preprint arXiv:2403.17804}, 2024.

\bibitem[Mokady et~al.(2023)Mokady, Hertz, Aberman, Pritch, and
  Cohen-Or]{mokady2023null}
Ron Mokady, Amir Hertz, Kfir Aberman, Yael Pritch, and Daniel Cohen-Or.
\newblock Null-text inversion for editing real images using guided diffusion
  models.
\newblock In \emph{CVPR}, 2023.

\bibitem[Morshed and Boddeti(2025)]{morshed2025diverseflow}
Mashrur~M Morshed and Vishnu Boddeti.
\newblock Diverseflow: Sample-efficient diverse mode coverage in flows.
\newblock In \emph{CVPR}, 2025.

\bibitem[Novack et~al.(2024)Novack, McAuley, Berg-Kirkpatrick, and
  Bryan]{novack2024dittodiffusioninferencetimetoptimization}
Zachary Novack, Julian McAuley, Taylor Berg-Kirkpatrick, and Nicholas~J Bryan.
\newblock Ditto: Diffusion inference-time t-optimization for music generation.
\newblock \emph{arXiv preprint arXiv:2401.12179}, 2024.

\bibitem[Oquab et~al.(2023)Oquab, Darcet, Moutakanni, Vo, Szafraniec, Khalidov,
  Fernandez, Haziza, Massa, El-Nouby, et~al.]{dinov2}
Maxime Oquab, Timoth{\'e}e Darcet, Th{\'e}o Moutakanni, Huy Vo, Marc
  Szafraniec, Vasil Khalidov, Pierre Fernandez, Daniel Haziza, Francisco Massa,
  Alaaeldin El-Nouby, et~al.
\newblock Dinov2: Learning robust visual features without supervision.
\newblock \emph{arXiv preprint arXiv:2304.07193}, 2023.

\bibitem[Parmar et~al.(2022)Parmar, Zhang, and Zhu]{parmar2021cleanfid}
Gaurav Parmar, Richard Zhang, and Jun-Yan Zhu.
\newblock On aliased resizing and surprising subtleties in gan evaluation.
\newblock In \emph{CVPR}, 2022.

\bibitem[Parmar et~al.(2025)Parmar, Patashnik, Ostashev, Wang, Aberman,
  Narasimhan, and Zhu]{gi}
Gaurav Parmar, Or Patashnik, Daniil Ostashev, Kuan-Chieh Wang, Kfir Aberman,
  Srinivasa Narasimhan, and Jun-Yan Zhu.
\newblock Scaling group inference for diverse and high-quality generation.
\newblock \emph{arXiv preprint arXiv:2508.15773}, 2025.

\bibitem[Radford et~al.(2021)Radford, Kim, Hallacy, Ramesh, Goh, Agarwal,
  Sastry, Askell, Mishkin, Clark, et~al.]{clip}
Alec Radford, Jong~Wook Kim, Chris Hallacy, Aditya Ramesh, Gabriel Goh,
  Sandhini Agarwal, Girish Sastry, Amanda Askell, Pamela Mishkin, Jack Clark,
  et~al.
\newblock Learning transferable visual models from natural language
  supervision.
\newblock In \emph{ICML}, 2021.

\bibitem[Ramesh et~al.(2021)Ramesh, Pavlov, Goh, Gray, Voss, Radford, Chen, and
  Sutskever]{dalle}
Aditya Ramesh, Mikhail Pavlov, Gabriel Goh, Scott Gray, Chelsea Voss, Alec
  Radford, Mark Chen, and Ilya Sutskever.
\newblock Zero-shot text-to-image generation.
\newblock In \emph{ICML}, 2021.

\bibitem[Razavi et~al.(2019)Razavi, Van~den Oord, and Vinyals]{vqvae2}
Ali Razavi, Aaron Van~den Oord, and Oriol Vinyals.
\newblock Generating diverse high-fidelity images with vq-vae-2.
\newblock \emph{NeurIPS}, 2019.

\bibitem[Ruiz et~al.(2023)Ruiz, Li, Jampani, Pritch, Rubinstein, and
  Aberman]{dreambooth}
Nataniel Ruiz, Yuanzhen Li, Varun Jampani, Yael Pritch, Michael Rubinstein, and
  Kfir Aberman.
\newblock Dreambooth: Fine tuning text-to-image diffusion models for
  subject-driven generation.
\newblock In \emph{CVPR}, 2023.

\bibitem[Ruiz et~al.(2024)Ruiz, Li, Jampani, Wei, Hou, Pritch, Wadhwa,
  Rubinstein, and Aberman]{hyperdreambooth}
Nataniel Ruiz, Yuanzhen Li, Varun Jampani, Wei Wei, Tingbo Hou, Yael Pritch,
  Neal Wadhwa, Michael Rubinstein, and Kfir Aberman.
\newblock Hyperdreambooth: Hypernetworks for fast personalization of
  text-to-image models.
\newblock In \emph{CVPR}, 2024.

\bibitem[Sadat et~al.(2023)Sadat, Buhmann, Bradley, Hilliges, and
  Weber]{sadat2023cads}
Seyedmorteza Sadat, Jakob Buhmann, Derek Bradley, Otmar Hilliges, and Romann~M
  Weber.
\newblock Cads: Unleashing the diversity of diffusion models through
  condition-annealed sampling.
\newblock \emph{arXiv preprint arXiv:2310.17347}, 2023.

\bibitem[Sadat et~al.(2024)Sadat, Hilliges, and Weber]{sadat2024eliminating}
Seyedmorteza Sadat, Otmar Hilliges, and Romann~M Weber.
\newblock Eliminating oversaturation and artifacts of high guidance scales in
  diffusion models.
\newblock In \emph{ICLR}, 2024.

\bibitem[Samuel et~al.(2023)Samuel, Ben-Ari, Darshan, Maron, and
  Chechik]{Samuel2023NAO}
Dvir Samuel, Rami Ben-Ari, Nir Darshan, Haggai Maron, and Gal Chechik.
\newblock Norm-guided latent space exploration for text-to-image generation.
\newblock \emph{NeurIPS}, 2023.

\bibitem[Samuel et~al.(2024)Samuel, Ben-Ari, Raviv, Darshan, and
  Chechik]{Samuel2023SeedSelect}
Dvir Samuel, Rami Ben-Ari, Simon Raviv, Nir Darshan, and Gal Chechik.
\newblock Generating images of rare concepts using pre-trained diffusion
  models.
\newblock In \emph{AAAI}, 2024.

\bibitem[Sauer et~al.(2023)Sauer, Lorenz, Blattmann, and Rombach]{sdxlturbo}
Axel Sauer, Dominik Lorenz, Andreas Blattmann, and Robin Rombach.
\newblock Adversarial diffusion distillation.
\newblock \emph{arXiv preprint arXiv:2311.17042}, 2023.

\bibitem[Simoncelli and Olshausen(2001)]{simoncelli2001natural}
Eero~P Simoncelli and Bruno~A Olshausen.
\newblock Natural image statistics and neural representation.
\newblock \emph{Annual review of neuroscience}, 2001.

\bibitem[Simonyan and Zisserman(2014)]{simonyan2014very}
Karen Simonyan and Andrew Zisserman.
\newblock Very deep convolutional networks for large-scale image recognition.
\newblock \emph{arXiv preprint arXiv:1409.1556}, 2014.

\bibitem[Singh et~al.(2024)Singh, Li, Shi, Krishna, Choi, Koh, Cohen, Gould,
  Zheng, and Zettlemoyer]{singh2024negative}
Jaskirat Singh, Lindsey Li, Weijia Shi, Ranjay Krishna, Yejin Choi, Pang~Wei
  Koh, Michael~F Cohen, Stephen Gould, Liang Zheng, and Luke Zettlemoyer.
\newblock Negative token merging: Image-based adversarial feature guidance.
\newblock \emph{arXiv preprint arXiv:2412.01339}, 2024.

\bibitem[Sundaram et~al.(2024)Sundaram, Pal, Chauhan, Agarwal, and
  Karanam]{sundaram2024cocono}
Aravindan Sundaram, Ujjayan Pal, Abhimanyu Chauhan, Aishwarya Agarwal, and
  Srikrishna Karanam.
\newblock Cocono: Attention contrast-and-complete for initial noise
  optimization in text-to-image synthesis.
\newblock \emph{arXiv preprint arXiv:2411.16783}, 2024.

\bibitem[Tang et~al.(2024)Tang, Peng, Tang, Hong, Wang, and
  Chang]{tang2024inference}
Zhiwei Tang, Jiangweizhi Peng, Jiasheng Tang, Mingyi Hong, Fan Wang, and
  Tsung-Hui Chang.
\newblock Inference-time alignment of diffusion models with direct noise
  optimization.
\newblock \emph{arXiv preprint arXiv:2405.18881}, 2024.

\bibitem[Torralba and Oliva(2003)]{torralba2003statistics}
Antonio Torralba and Aude Oliva.
\newblock Statistics of natural image categories.
\newblock \emph{Network: computation in neural systems}, 2003.

\bibitem[Torralba et~al.(2008)Torralba, Fergus, and Freeman]{torralba200880}
Antonio Torralba, Rob Fergus, and William~T Freeman.
\newblock 80 million tiny images: A large data set for nonparametric object and
  scene recognition.
\newblock \emph{TPAMI}, 2008.

\bibitem[Uehara et~al.(2025{\natexlab{a}})Uehara, Su, Zhao, Li, Regev, Ji,
  Levine, and Biancalani]{uehara2025rewardguidediterativerefinementdiffusion}
Masatoshi Uehara, Xingyu Su, Yulai Zhao, Xiner Li, Aviv Regev, Shuiwang Ji,
  Sergey Levine, and Tommaso Biancalani.
\newblock Reward-guided iterative refinement in diffusion models at test-time
  with applications to protein and dna design.
\newblock \emph{arXiv preprint arXiv:2502.14944}, 2025{\natexlab{a}}.

\bibitem[Uehara et~al.(2025{\natexlab{b}})Uehara, Zhao, Wang, Li, Regev,
  Levine, and Biancalani]{uehara2025inferencetimealignmentdiffusionmodels}
Masatoshi Uehara, Yulai Zhao, Chenyu Wang, Xiner Li, Aviv Regev, Sergey Levine,
  and Tommaso Biancalani.
\newblock Inference-time alignment in diffusion models with reward-guided
  generation: Tutorial and review.
\newblock \emph{arXiv preprint arXiv:2501.09685}, 2025{\natexlab{b}}.

\bibitem[Wallace et~al.(2023)Wallace, Gokul, Ermon, and Naik]{doodl}
Bram Wallace, Akash Gokul, Stefano Ermon, and Nikhil Naik.
\newblock End-to-end diffusion latent optimization improves classifier
  guidance.
\newblock In \emph{ICCV}, 2023.

\bibitem[Wu et~al.(2024)Wu, Si, Jiang, Huang, and Liu]{wu2024freeinit}
Tianxing Wu, Chenyang Si, Yuming Jiang, Ziqi Huang, and Ziwei Liu.
\newblock Freeinit: Bridging initialization gap in video diffusion models.
\newblock In \emph{ECCV}, 2024.

\bibitem[Wu et~al.(2023{\natexlab{a}})Wu, Hao, Sun, Chen, Zhu, Zhao, and
  Li]{hpsv2}
Xiaoshi Wu, Yiming Hao, Keqiang Sun, Yixiong Chen, Feng Zhu, Rui Zhao, and
  Hongsheng Li.
\newblock Human preference score v2: A solid benchmark for evaluating human
  preferences of text-to-image synthesis.
\newblock \emph{arXiv preprint arXiv:2306.09341}, 2023{\natexlab{a}}.

\bibitem[Wu et~al.(2023{\natexlab{b}})Wu, Sun, Zhu, Zhao, and Li]{hps}
Xiaoshi Wu, Keqiang Sun, Feng Zhu, Rui Zhao, and Hongsheng Li.
\newblock Better aligning text-to-image models with human preference.
\newblock In \emph{ICCV}, 2023{\natexlab{b}}.

\bibitem[Xu et~al.(2025)Xu, Zhang, and Shi]{xu2025good}
Katherine Xu, Lingzhi Zhang, and Jianbo Shi.
\newblock Good seed makes a good crop: Discovering secret seeds in
  text-to-image diffusion models.
\newblock In \emph{WACV}, 2025.

\bibitem[Zhang et~al.(2018)Zhang, Isola, Efros, Shechtman, and Wang]{lpips}
Richard Zhang, Phillip Isola, Alexei~A Efros, Eli Shechtman, and Oliver Wang.
\newblock The unreasonable effectiveness of deep features as a perceptual
  metric.
\newblock In \emph{CVPR}, 2018.

\end{thebibliography}
}

\clearpage

\setcounter{section}{0}
\setcounter{figure}{0}
\setcounter{table}{0}
\setcounter{equation}{0}

\twocolumn[{%
\renewcommand\twocolumn[1][]{#1}%
\begin{center}
    \vspace*{0.5em}
    {\Large\bfseries Supplementary Material:\par}
    \vspace{0.5em}
    {\Large\bfseries It's Never Too Late:\\ Noise Optimization for Collapse Recovery
    in Trained Diffusion Models\par}
    \vspace{1em}
\end{center}
}]

\section{Implementation Details}

\subsection{Optimization Objectives and Metrics}
\paragraph{Output Diversity.}
We use multiple diversity objectives that aim at generating a set of diverse images with diffusion models. In the following, we first describe the pairwise diversity metrics that we used. \\

\noindent \textbf{DINO.} This diversity objective and metric uses DINOv2~\cite{dinov2} patch features to measure perceptual diversity as defined in Eq. 3 in the main paper. Specifically, we compute the pairwise cosine distances (i.e.\ $d$ is the cosine distance) between patch features in different images. Lower values indicate similar images, and values closer to $1$ represent higher diversity. We also refer to this metric as ``Output variation (DINO)''.

\noindent \textbf{DreamSim.} We use pairwise DreamSim dissimilarity scores obtained with a DINO ViT-B/16 backbone that was trained to align with human perception~\cite{fu2023dreamsim}. Lower values indicate similar images, whereas values closer to $1$ correspond to more diversity in the outputs.

\noindent \textbf{LPIPS.} We use LPIPS~\cite{lpips} to quantify the dissimilarity between a pair of images with a VGG~\cite{simonyan2014very} backbone. Specifically, LPIPS computes a weighted sum of perceptual similarities across the outputs of all five convolutional blocks of VGG16. Values close to $0$ indicate similar images, whereas values closer to $1$ indicate higher diversity.

\noindent \textbf{Color Histogram.} We consider the pairwise color histogram distance between images. In particular, we calculate color histograms for each channel considering 32 bins. We use soft histograms with Gaussian kernels to ensure that this operation is differentiable. We then measure the pairwise L2 distance between the resulting color histograms of two images, and normalize this such that the final score is in the range $[0,1]$.

\noindent \textbf{L2.} Inspired by the image similarity used in \cite{torralba200880}, we use a low-resolution L2 distance between pairs of images. In particular, we resize the generated images to $32 \times 32$ and compute the L2 distance between the resulting 3072-dimensional vectors representing each image. We normalize this score to be in the range $[0,1]$. Higher values correspond to higher diversity. \\

In addition to the above-described averaged pairwise diversity objectives, we consider two set-based metrics.

\noindent \textbf{DPP.} 
We normalize the DINOv2 [CLS] token embeddings $\bar{f}_i$ for each image $x^{(i)}$. The normalized embeddings are used to construct a similarity kernel matrix $K_{s} = \bar{F} \bar{F}^T$ where $\bar{F} = [\bar{f}_1, \bar{f}_2, \ldots, \bar{f}_N]^T$, and $N$ the number of images. The kernel is symmetrized as $K_{sym} = (K_{s} + K_{s}^T)/2$ and augmented with $K \leftarrow K_{sym} + \epsilon I$ where $\epsilon = 10^{-6}$. The Determinantal Point Process (DPP) score~\cite{kulesza2012determinantal} is then computed as the log-determinant:
\begin{equation}
\mathcal{D}_{\text{DPP}} = \log \det(I + K).
\end{equation}
This score ranges between  $[0,\log(16)]$ for a set of four images, with $0$ indicating that all images are identical, and 2.77 stating that all images in the set are maximally diverse. 

\noindent \textbf{Vendi.} Starting with the same similarity kernel $K$ as in DPP, we compute its eigenvalue decomposition to obtain $\lambda_1, \lambda_2, \ldots, \lambda_N$. These eigenvalues are normalized to form a probability distribution $p_i = \lambda_i / \sum_{j=1}^N \lambda_j$. The Vendi score~\cite{friedman2022vendi} is defined as the exponential of the Shannon entropy of this distribution:
\begin{equation}
\mathcal{D}_{\text{Vendi}} = \exp\left(-\sum_{i=1}^N p_i \log(p_i + \delta)\right),
\end{equation}
where $\delta = 10^{-12}$ to prevent numerical issues.
This score is between $[1,4]$ for a set of four images, measuring the effective number of diverse images in the set. A score of $1$ signifies that all images are effectively similar, and $4$ shows that each image in the set is unique. 

\paragraph{Image Quality.}
We optimize image quality using CLIPScore and a human preference score. 

\noindent \textbf{CLIPScore.} Similar to \cite{reno}, we use a reward model that pushes the optimization process to preserve image quality and prompt relevance. Specifically, we use a pretrained CLIP~\cite{clip} ViT-B/32 model. \cite{gi} also used this model to ensure image quality and prompt following.

\noindent \textbf{HPSv2.} We use the HPSv2~\cite{hpsv2} metric as another image reward to maintain quality during optimization. It is based on a CLIP~\cite{clip} ViT-H/14 backbone.

For evaluation, in addition to measuring CLIPScore and HPSv2 we report PickScore~\cite{pickscore}, and for completeness we also report FID~\cite{heusel2017gans} on a subset of results. 

\noindent \textbf{PickScore.} We reserve PickScore~\cite{pickscore} as an independent quality metric since we do not use it during optimizations.  The metric uses a CLIP~\cite{clip} ViT-H/14 backbone fine-tuned on Pick-a-Pic user preference dataset. Higher PickScore values indicate better quality.

\noindent \textbf{FID.} For additional comparisons, we also evaluate quality with Fr\'{e}chet Inception Distance (FID)~\cite{heusel2017gans}. To obtain scores, we compare generated images to the COCO~\cite{mscoco} validation set which consists of 5000 images. 
Specifically, we compute FID using the clean-fid library~\cite{parmar2021cleanfid} with CLIP ViT-B/32~\cite{clip} features and its recommended ``clean'' preprocessing pipeline.
For each prompt in an evaluation set, we generate four samples. 
In our setting, images are generated from GenEval and DPG prompts that bear little resemblance to COCO images, making the reference distribution a suboptimal match. Furthermore, with only small image sets (e.g.\ 2212 generated images for GenEval), FID estimates are unreliable~\cite{binkowski2018demystifying}. We therefore mainly rely on per-image quality metrics (HPSv2, PickScore) that do not assume a matching reference distribution.

\subsection{Hyperparameter Choices}\label{sec:supp_hyperparams}
We use the SDXL-Turbo~\cite{sdxlturbo}, SANA-Sprint~\cite{sanasprint}, PixArt-$\alpha$-DMD~\cite{pixartalpha}, and Flux.1 [schnell]~\cite{flux} models in our experiments. For the majority of our experiments we show results for batched optimization. In this case for \textbf{$\text{i.i.d.}$ samples}, we randomly sample input noise and generate a set of four images in a model's default configuration without altering the four initial noises.

For sequential optimization (see main paper Fig. 3), we use 25 iterations, a learning rate of $3.0$, $\lambda_{div}=15$ for the DPP diversity objective, $\lambda_{q}=1$ for a HPSv2 quality reward, and gradient clipping of $0.15$.

All batched experiments use $\lambda_{\text{reg}} = 0.01$ (Eq.~2 main paper) and pink noise 
exponent $\alpha = 0.2$ unless otherwise noted. Table~\ref{tab:hparams_experiments} 
summarizes per-experiment settings; Table~\ref{tab:hparams_objectives} summarizes 
settings used in the diversity objective comparison (Sec.~4.1 main paper). For SDXL-Turbo, PixArt, and Flux.1 [schnell], we use image resolutions of $512 \times 512$, and $768 \times 768$ for SANA-Sprint. 

\paragraph{\citet{gi}.} We apply \cite{gi} to the SDXL-Turbo, SANA-Sprint, PixArt-$\alpha$, and Flux.1 [schnell] models. We use the default parameters that were used for Flux.1 [schnell]~\cite{flux,flux_url} in \cite{gi}, since this setting is closest to our setup with one-step / few-step models. However, for SDXL-Turbo and PixArt, we use image resolutions of $512 \times 512$. We use $768 \times 768$ for SANA-Sprint, and $512 \times 512$ for Flux.1 [schnell].

\begin{table*}[h]
\centering
\caption{Hyperparameters per experiment. $\lambda_{\text{div}}$ and $\lambda_q$ weight 
the diversity and quality terms in Eq.~2 (main paper). ``Revert'' indicates whether 
optimization reverts to the previous latent when HPSv2 drops below a threshold. For Flux.1 [schnell], white noise used no HPSv2 weighting, and pink noise used no HPSv2 weighting until iteration 20.}
\label{tab:hparams_experiments}
\setlength{\tabcolsep}{4pt}
\begin{tabular}{llllcccccl}
\toprule
\textbf{Table / Fig.} & \textbf{Model} & \textbf{Noise} & \textbf{Objective} 
  & $\lambda_{\text{div}}$ & $\lambda_q$ & \textbf{LR} & \textbf{Grad Clip} 
  & \textbf{Iter.} & \textbf{Revert} \\
\midrule
Tab.~1, 3, 4  & SDXL-Turbo       & white / pink & DINO + CLIP  & 80  & 50 & 10.0 & 0.1  & 100 & -- \\
Tab.~3, 4     & PixArt-$\alpha$   & white / pink & DINO + CLIP  & 80  & 50 & 10.0 & 0.1  & 100 & -- \\
Tab.~3, 4     & SANA-Sprint-1.6B  & white / pink & DINO + CLIP  & 25  & 10 & 10.0 & 0.1  & 100 & -- \\
\midrule
Tab.~2        & SDXL-Turbo       & white        & DPP + HPSv2  & 150 & 3  & 6.0  & 0.1  & 150 & -- \\
Tab.~2        & SDXL-Turbo       & pink         & DPP + HPSv2  & 150 & 3  & 6.0  & 0.1   & 150 & -- \\
Tab.~2        & Flux.1 [schnell] & white        & DPP + HPSv2  & 1.5 & -- & 6.0  & 0.1  & 80  & hard $< 0.31$ \\
Tab.~2        & Flux.1 [schnell] & pink         & DPP + HPSv2  & 5 & 5 & 2  & 0.15  & 80 & relative $< 0.05$ \\
\midrule
\multicolumn{10}{l}{\textit{Sequential generation}} \\
Fig.~3        & Flux.1 [schnell] & white        & DPP + HPSv2  & 15  & 1  & 3.0  & 0.15 & 25  & -- \\
\bottomrule
\end{tabular}
\end{table*}

\begin{table*}[h]
\centering
\caption{Hyperparameters used for the diversity objective comparison with
SDXL-Turbo and white noise initialization on GenEval ($\lambda_q = 10$,  LR 10.0, and grad. clip 0.1 for all objectives.)}
\label{tab:hparams_objectives}
\begin{tabular}{lcccc}
\toprule
\textbf{Objective} & $\lambda_{\text{div}}$ & \textbf{Max Iter.} 
  & \textbf{Stop Threshold} & \textbf{Threshold Type} \\
\midrule
DINO       & 50 & 100 & 0.9        & absolute       \\
DreamSim   & 70 & 50  & 0.9        & absolute       \\
LPIPS      & 60 & 60  & 0.9        & absolute       \\
Color Hist.& 60 & 60  & $4\times$  & relative to $\text{i.i.d.}$ \\
L2         & 60 & 60  & $2\times$  & relative to $\text{i.i.d.}$ \\
DPP        & 50 & 100 & $4\times$  & relative to $\text{i.i.d.}$ \\
Vendi      & 50 & 100 & $4\times$  & relative to $\text{i.i.d.}$ \\
\bottomrule
\end{tabular}
\end{table*}

\begin{figure}[t]
    \centering
    \includegraphics[width=\linewidth]{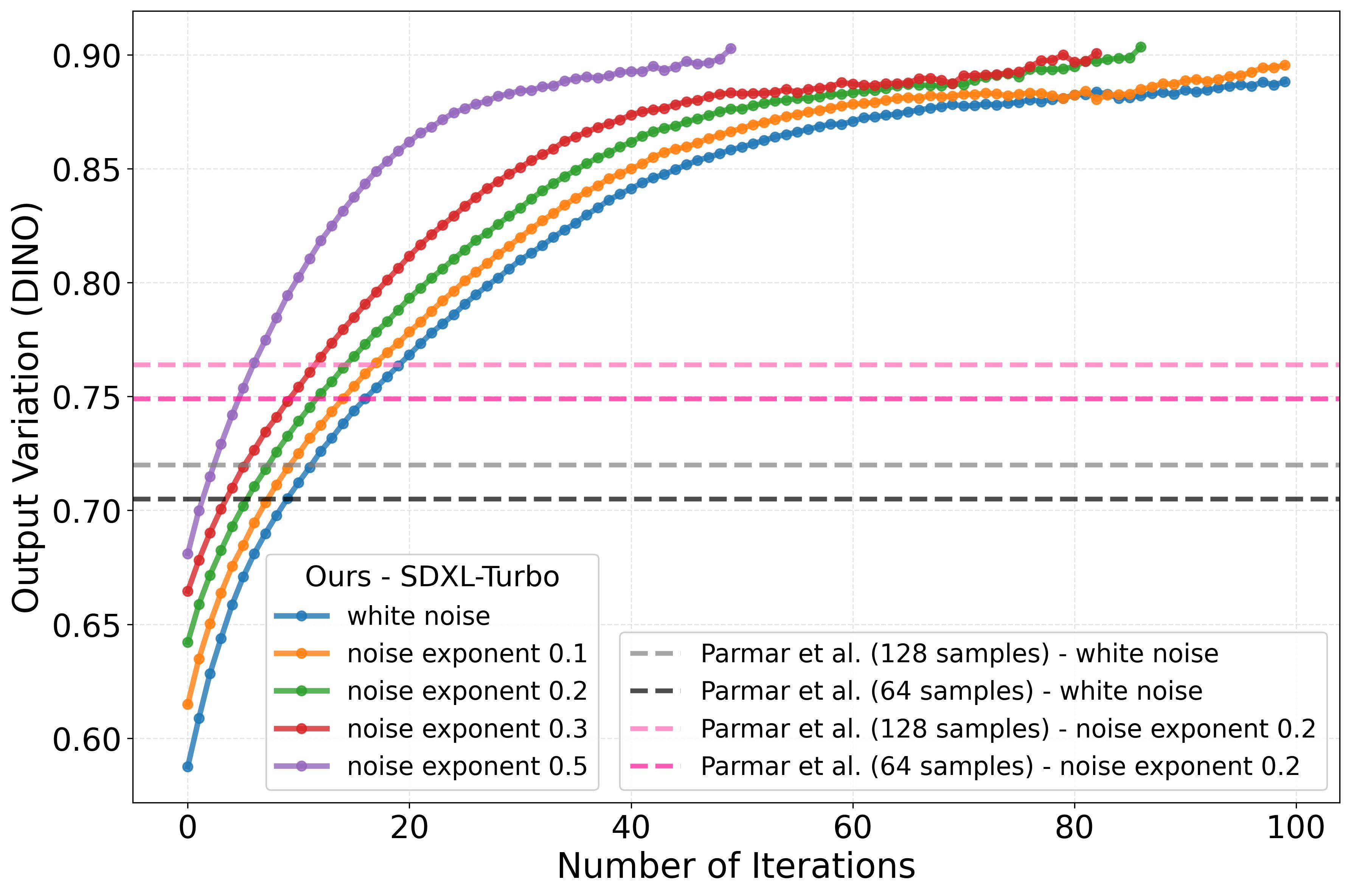}
    \vspace{-.15in}
    \caption{Output variation across optimization iterations for SDXL-Turbo with different noise initializations on GenEval. Higher noise exponents produce greater diversity. Dashed lines are baseline scores from \cite{gi} for white noise (gray/black) and pink noise with exponent $0.2$ (pink tones) using 64 and 128 samples. Our approach reaches higher diversity (output variation) than \cite{gi}, requiring only relatively few iterations to outperform \cite{gi}.}
    \label{fig:scaling_curve}
\end{figure}

\subsection{Datasets}
\noindent \textbf{GenEval~\cite{ghosh2023geneval}} is a text-to-image generation benchmark that evaluates models across 553 diverse prompts requiring understanding of complex compositional relationships. Unless mentioned otherwise, we report results across all prompts in the dataset.

\noindent \textbf{T2I-CompBench~\cite{huang2023t2icompbench}} tests compositional understanding in text-to-image models across eight distinct categories: color, shape, texture, spatial relationships, non-spatial attributes, complex compositions, 3D spatial reasoning, and numeracy. We select 50 random prompts per category, resulting in a set of 400 prompts.

\noindent \textbf{DPG-Bench~\cite{hu2024ella}} evaluates image generation on 1,065 long, detailed prompts with an average length of 67 words. We use this to assess our diversity optimization on highly detailed text prompts.

\section{Computational Cost}\label{sec:supp_computational_cost}

We measure the time per iteration on a single A100 80GB GPU in \cref{tab:compute_iters}. Numbers reported are an average over 100 iterations with the error reported over three different seeds. It takes less than 15 iterations to reach similar levels of diversity as \citet{gi} on GenEval~\cite{ghosh2023geneval} (see \cref{fig:scaling_curve}).

\begin{table}[h]
\caption{Time per iteration of our proposed optimization approach. We report time on a single A100 80GB in seconds using DPP and HPSv2 objectives.}
\centering
\begin{tabular}{lc}
\toprule
\textbf{Model} &  \textbf{Time per Iteration}\\
\midrule
SDXL-Turbo & 0.345$_{\pm 0.004}$  \\
Flux.1 [schnell] & 1.092$_{\pm 0.008}$ \\
\bottomrule
\bottomrule
\end{tabular}
\label{tab:compute_iters}
\end{table}

\section{Additional Experimental Results}

\begin{table*}[h]
\centering
\caption{Output diversity and image-text alignment results on GenEval and T2I-CompBench for our proposed method with the PixArt-$\alpha$, SANA-Sprint-1.6B, and SDXL-Turbo models using white noise initialization. Output diversity is measured with averaged pairwise DINO, DreamSim, and LPIPS scores.}
\vspace{-.08in}
\resizebox{\textwidth}{!}{%
\begin{tabular}{lcccc|cccc}
  \toprule
  &  \multicolumn{4}{c}{\textbf{GenEval}~\cite{ghosh2023geneval}} & \multicolumn{4}{c}{\textbf{T2I-CompBench}~\cite{huang2023t2icompbench}} \\
  \cmidrule(lr){2-5} \cmidrule(lr){6-9}
  \textbf{Method} &  DINO & DreamSim & LPIPS & CLIPScore & DINO & DreamSim & LPIPS & CLIPScore \\
  \midrule
   \textbf{PixArt-$\alpha$}~\cite{pixartalpha} & & &&&&&& \\ 
   \midrule%
   $\text{i.i.d.}$   & 0.431$_{\pm 0.094}$ & 0.182$_{\pm 0.080}$ & 0.474$_{\pm 0.119}$ &  0.326$_{\pm 0.030}$  & 0.469$_{\pm 0.084}$ & 0.188$_{\pm 0.069}$ & 0.512$_{\pm 0.099}$ & 0.326$_{\pm 0.027}$  \\
    \citet{gi}   &  0.559$_{\pm 0.091}$ & 0.246$_{\pm 0.094}$ & 0.569$_{\pm 0.107}$ & 0.327$_{\pm 0.028}$  &  0.590$_{\pm 0.078}$ & 0.256$_{\pm 0.088}$ & 0.593$_{\pm 0.088}$ & 0.328$_{\pm 0.027}$  \\
    \ours\ (DINO) &   0.695$_{\pm 0.063}$ & 0.335$_{\pm 0.107}$ & 0.664$_{\pm 0.089}$  & 0.337$_{\pm 0.026}$  & 0.716$_{\pm 0.060}$ & 0.331$_{\pm 0.102}$ & 0.674$_{\pm 0.072}$ & 0.335$_{\pm 0.023}$  \\
  \midrule
   \textbf{SANA-Sprint-1.6B}~\cite{sanasprint} &  & &&&&&& \\ 
   \midrule%
   $\text{i.i.d.}$   &  0.526$_{\pm 0.088}$ & 0.229$_{\pm 0.075}$ & 0.635$_{\pm 0.087}$  & 0.336$_{\pm 0.032}$ &  0.562$_{\pm 0.074}$ & 0.252$_{\pm 0.078}$ & 0.656$_{\pm 0.066}$ & 0.334$_{\pm 0.029}$ \\
   \citet{gi}   &  0.714$_{\pm 0.060}$ & 0.354$_{\pm 0.095}$ & 0.741$_{\pm 0.055}$ & 0.342$_{\pm 0.032}$  &  0.684$_{\pm 0.060}$ & 0.331$_{\pm 0.089}$ & 0.718$_{\pm 0.049}$ & 0.338$_{\pm 0.028}$  \\
    \ours\ (DINO) &  0.744$_{\pm 0.061}$ & 0.438$_{\pm 0.099}$ & 0.781$_{\pm 0.062}$ & 0.335$_{\pm 0.030}$ &  0.738$_{\pm 0.056}$ & 0.437$_{\pm 0.105}$ & 0.767$_{\pm 0.053}$ & 0.330$_{\pm 0.029}$ \\
    \midrule
 \textbf{SDXL-Turbo}~\cite{sdxlturbo} & & & &&&&& \\  
 \midrule%
 $\text{i.i.d.}$  &  0.588$_{\pm 0.083}$ & 0.249$_{\pm 0.089}$ & 0.642$_{\pm 0.059}$ & 0.335$_{\pm 0.031}$  &  0.586$_{\pm 0.079}$ & 0.244$_{\pm 0.077}$ & 0.634$_{\pm 0.056}$ & 0.332$_{\pm 0.029}$  \\
  \citet{gi}  & 0.705$_{\pm 0.065}$ & 0.331$_{\pm 0.098}$ & 0.682$_{\pm 0.055}$ & 0.333$_{\pm 0.028}$ & 0.701$_{\pm 0.063}$ & 0.329$_{\pm 0.087}$ & 0.680$_{\pm 0.048}$ & 0.334$_{\pm 0.029}$ \\
  \ours\ (DINO) &  0.784$_{\pm 0.026}$ & 0.411$_{\pm 0.102}$ & 0.767$_{\pm 0.052}$ & 0.349$_{\pm 0.029}$  &  0.799$_{\pm 0.021}$ & 0.424$_{\pm 0.085}$ & 0.764$_{\pm 0.056}$ & 0.351$_{\pm 0.027}$  \\
  \bottomrule
  \bottomrule

\end{tabular}%
}

\label{tab:supp_main_table}
\end{table*}

\begin{table*}[h]
\centering
\caption{Output diversity and image-text alignment results on GenEval and T2I-CompBench for our proposed method and pink noise initialization with the PixArt-$\alpha$, SANA-Sprint-1.6B, and SDXL-Turbo models. Output diversity is measured with averaged pairwise DINO, DreamSim, and LPIPS scores.}
\resizebox{\textwidth}{!}{%
\begin{tabular}{lccccc|cccc}
  \toprule
  & & \multicolumn{4}{c}{\textbf{GenEval}~\cite{ghosh2023geneval}} & \multicolumn{4}{c}{\textbf{T2I-CompBench}~\cite{huang2023t2icompbench}} \\
  \cmidrule(lr){3-6} \cmidrule(lr){7-10}
  \textbf{Method} & \textbf{Noise} & DINO & DreamSim & LPIPS & CLIPScore & DINO & DreamSim & LPIPS & CLIPScore \\
  \midrule
   \textbf{PixArt-$\alpha$}~\cite{pixartalpha} & & & &&&&&& \\ 
   \midrule%
   $\text{i.i.d.}$  & \pfat &  0.533$_{\pm 0.088}$ & 0.244$_{\pm 0.091}$ & 0.604$_{\pm 0.116}$ & 0.326$_{\pm 0.030}$  & 0.558$_{\pm 0.077}$ & 0.247$_{\pm 0.083}$ & 0.626$_{\pm 0.095}$ & 0.325$_{\pm 0.027}$ \\
  \citet{gi}  & \pfat &  0.664$_{\pm 0.074}$ & 0.319$_{\pm 0.104}$ & 0.684$_{\pm 0.094}$ & 0.323$_{\pm 0.029}$  &  0.675$_{\pm 0.066}$ & 0.326$_{\pm 0.095}$ & 0.692$_{\pm 0.077}$ & 0.324$_{\pm 0.026}$ \\
 
   \ours\ (DINO) & \pfat &  0.764$_{\pm 0.039}$ & 0.388$_{\pm 0.102}$ &  0.750$_{\pm 0.067}$ &  0.335$_{\pm 0.029}$  & 0.770$_{\pm 0.046}$ & 0.377$_{\pm 0.097}$ & 0.748$_{\pm 0.063}$ & 0.333$_{\pm 0.024}$  \\
    \midrule
   \textbf{SANA-Sprint-1.6B}~\cite{sanasprint} & & & &&&&&& \\ 
   \midrule%
   $\text{i.i.d.}$  & \pfat &  0.551$_{\pm 0.083}$ & 0.235$_{\pm 0.075}$ & 0.649$_{\pm 0.083}$ &  0.335$_{\pm 0.033}$ &  0.584$_{\pm 0.069}$ & 0.259$_{\pm 0.079}$ & 0.670$_{\pm 0.065}$ & 0.334$_{\pm 0.029}$ \\
 
  \citet{gi}  & \pfat &  0.737$_{\pm 0.053}$ & 0.369$_{\pm 0.093}$ & 0.767$_{\pm 0.050}$ & 0.341$_{\pm 0.032}$  & 0.705$_{\pm 0.056}$ & 0.346$_{\pm 0.090}$ & 0.736$_{\pm 0.048}$ & 0.335$_{\pm 0.028}$  \\
   \ours\ (DINO) & \pfat &  0.753$_{\pm 0.049}$ & 0.440$_{\pm 0.093}$ & 0.784$_{\pm 0.056}$ & 0.334$_{\pm 0.031}$  &  0.750$_{\pm 0.046}$ & 0.443$_{\pm 0.096}$ & 0.773$_{\pm 0.050}$ & 0.330$_{\pm 0.030}$ \\
    \midrule
 \textbf{SDXL-Turbo}~\cite{sdxlturbo} & & & &&&&&& \\  
 \midrule
   $\text{i.i.d.}$  & \pfat &  0.642$_{\pm 0.068}$ & 0.305$_{\pm 0.090}$ & 0.729$_{\pm 0.052}$ & 0.328$_{\pm 0.031}$  &  0.643$_{\pm 0.071}$ & 0.303$_{\pm 0.080}$ & 0.719$_{\pm 0.055}$ & 0.326$_{\pm 0.028}$  \\
  
  \citet{gi} %
  & \pfat & 0.749$_{\pm 0.054}$ & 0.392$_{\pm 0.100}$ & 0.757$_{\pm 0.048}$ & 0.323$_{\pm 0.028}$   &  0.742$_{\pm 0.055}$ & 0.391$_{\pm 0.088}$ & 0.751$_{\pm 0.049}$ & 0.328$_{\pm 0.027}$ \\
  
    \ours\ (DINO) & \pfat &  
    0.786$_{\pm 0.028}$ & 0.427$_{\pm 0.095}$ & 0.811$_{\pm 0.044}$ & 0.341$_{\pm 0.029}$
   &0.804$_{\pm 0.026}$ & 0.440$_{\pm 0.084}$ & 0.808$_{\pm 0.049}$ & 0.344$_{\pm 0.026}$   \\
  \bottomrule
  \bottomrule

\end{tabular}%
}

\label{tab:supp_pink_table}
\end{table*}

\textbf{More Model Comparisons.} We provide additional results for PixArt-$\alpha$ and SANA-Sprint-1.6B. We evaluate models on GenEval~\cite{ghosh2023geneval} and a subset of $50$ prompts per category of T2I-CompBench~\cite{huang2023t2icompbench}. Across models, we observe substantial diversity gains with minimal loss in image quality, as reported in \cref{tab:supp_main_table} and \cref{tab:supp_pink_table}. To obtain our results on PixArt-$\alpha$, SANA-Sprint-1.6B, we use the hyperparameters specified in \cref{sec:supp_hyperparams}. For the SDXL-Turbo results, we use the same settings as Tab. 1 in the main paper. For comparison to \cite{gi}, we optimize using DINOv2~\cite{dinov2} and CLIP~\cite{clip} across a batch of $4$ output images. All pink noise results are obtained with $\alpha=0.2$.

\noindent \textbf{Results for Different Diversity Objectives.} For the SDXL-Turbo model, we additionally evaluate the effect of different diversity objectives during optimization (\cref{tab:diversity_metrics}). Using the same hyperparameters in \cref{tab:hparams_objectives}, we report scores optimizing for diversity with DINOv2~\cite{dinov2}, DreamSim~\cite{fu2023dreamsim}, LPIPS~\cite{lpips}, Color Histogram~\cite{torralba200880}, L2 distance, DPP~\cite{elfeki2019gdpp}, and Vendi~\cite{friedman2022vendi}. For quality we use a CLIP objective. We observe that set-level objectives DPP and Vendi produce the best diversity scores, consistent with our user study in Fig.~6 of the main paper.
We show generation results that compare different diversity objectives in \cref{fig:supp_objectives_1} and \cref{fig:supp_objectives_2}. These visualizations correspond to the quantitative results in \cref{tab:diversity_metrics}. We can observe that our approach yields more diverse output image sets compared to \cite{gi} and generations from $\text{i.i.d.}$-sampled noise initializations across different diversity objectives. All generations are obtained from white noise initializations using the SDXL-Turbo model.

\begin{figure*}[t]
    \centering
    \includegraphics[trim=25 20 20 20, clip, width=0.92\textwidth]{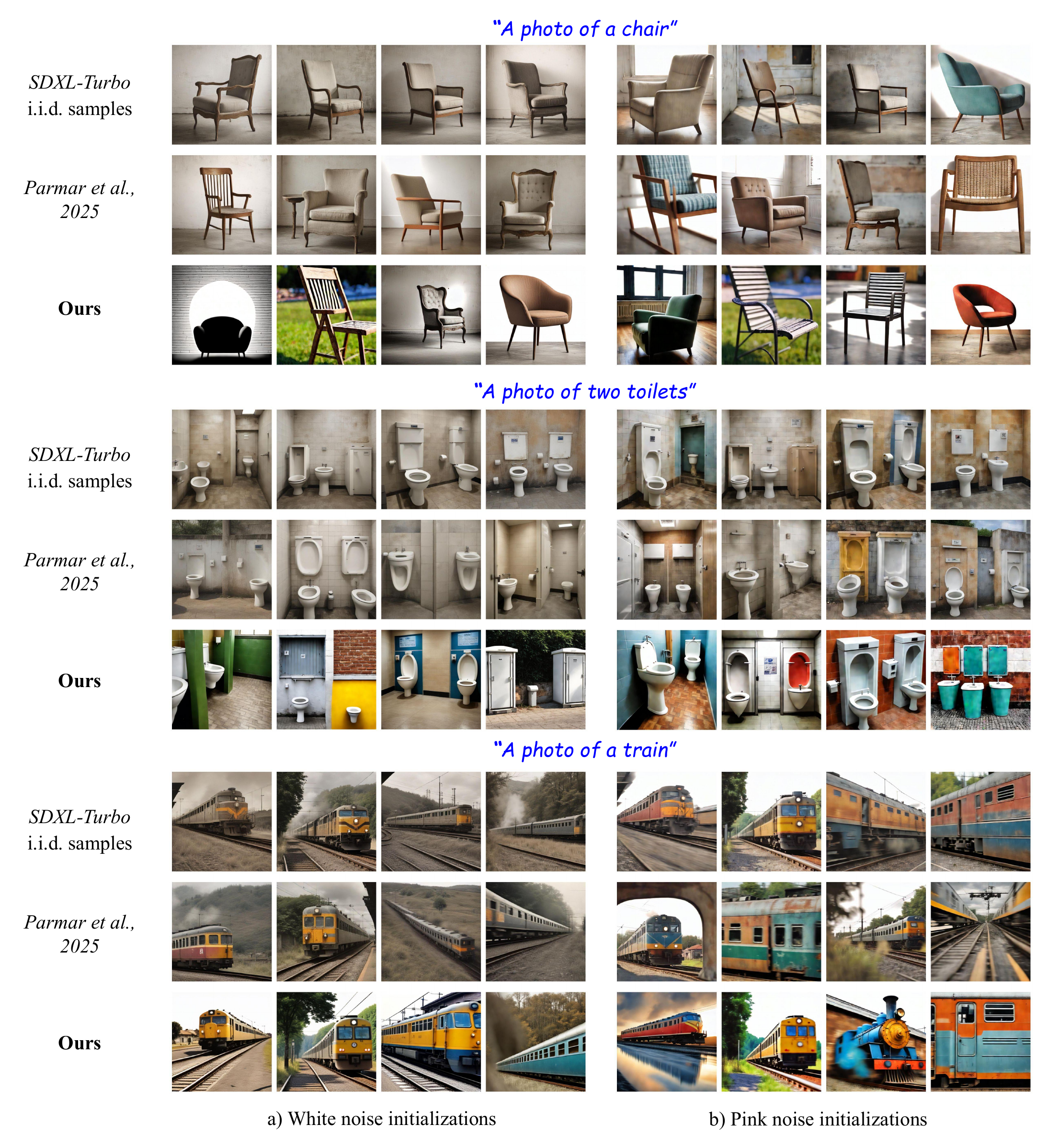}
    \caption{Image generations using our noise optimization approach for SDXL-Turbo yields improved diversity within generated image sets compared to $\text{i.i.d.}$ sampling and \cite{gi}. Pink noise initializations (b) give more diverse generations than standard white noise (a). Ours uses the DINO diversity objective (similar to \cref{tab:supp_main_table} and \cref{tab:supp_pink_table}).
    }
    \vspace{-0.5em}
    \label{fig:supp_example_generations}
\end{figure*}

\begin{figure*}[t]
    \centering
    \includegraphics[width=\textwidth]{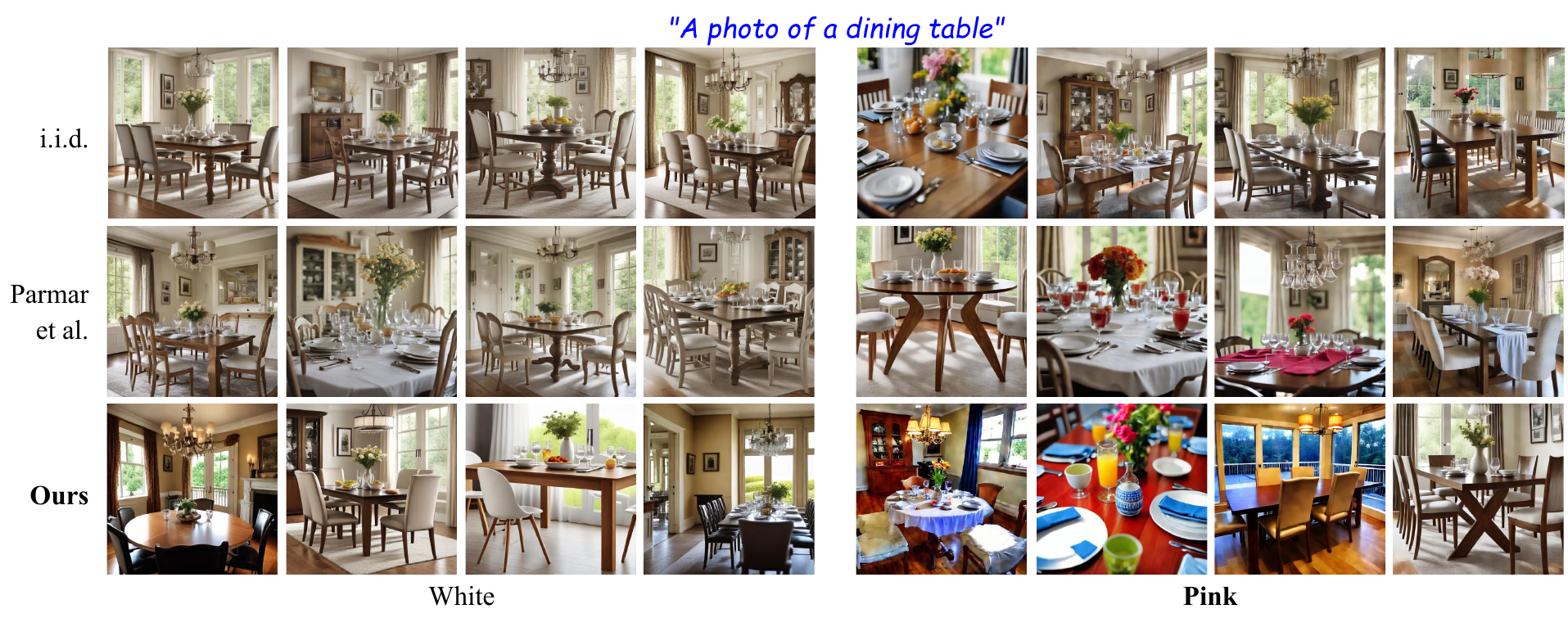}
    \vspace{-1.5em}
    \caption{Diverse image generation with SDXL-Turbo using white and pink noise initialization with DPP~\cite{elfeki2019gdpp} and HPSv2~\cite{hpsv2} objectives. We observe that our method improves diversity compared to $\text{i.i.d.}$ sampling with white noise. In addition, using pink noise simply at inference time without any optimization increases diversity for both $\text{i.i.d.}$ sampling and \cite{gi}.}
    \label{fig:example_generations_sdxl_pink}
\end{figure*}

\noindent \textbf{Qualitative Examples for SDXL-Turbo.} In \cref{fig:supp_example_generations}, we show example generations from SDXL-Turbo from the experiments in \cref{tab:supp_main_table} and \cref{tab:supp_pink_table}. We observe that our method produces greater visual diversity in terms of color, lighting, and pose across all prompts. We further observe that using pink noise initialization improves diversity even under $\text{i.i.d.}$ sampling and \cite{gi}. Example generations for different diversity objectives can be found in Fig. 5 of the main text. We also provide additional example generations using DPP and HPSv2 objectives for white and pink noise initializations in \cref{fig:example_generations_sdxl_pink}. 

\noindent \textbf{Qualitative Results for Flux.1 [schnell].}
We additionally test our optimization on a larger model, Flux.1 [schnell]. Using the best diversity objective from our ablations DPP, we generate results in \cref{fig:example_generations_flux}. Compared to $\text{i.i.d.}$ sampling and the default settings from \cite{gi}, we observe greater output diversity across multiple prompts, particularly in terms of object color, orientation, lighting, and also different backgrounds and positioning. We also provide additional examples optimizing with pink noise initialization in \cref{fig:example_generations_flux_pink}.

\begin{figure*}[t]
    \centering
    {\textit{\color{blue}{\comicfont \textbf{"A photo of a cat"}}}\\[1ex]}
        \begin{tabular}{p{0.48\textwidth} p{0.48\textwidth}}
        \includegraphics[width=1\linewidth]{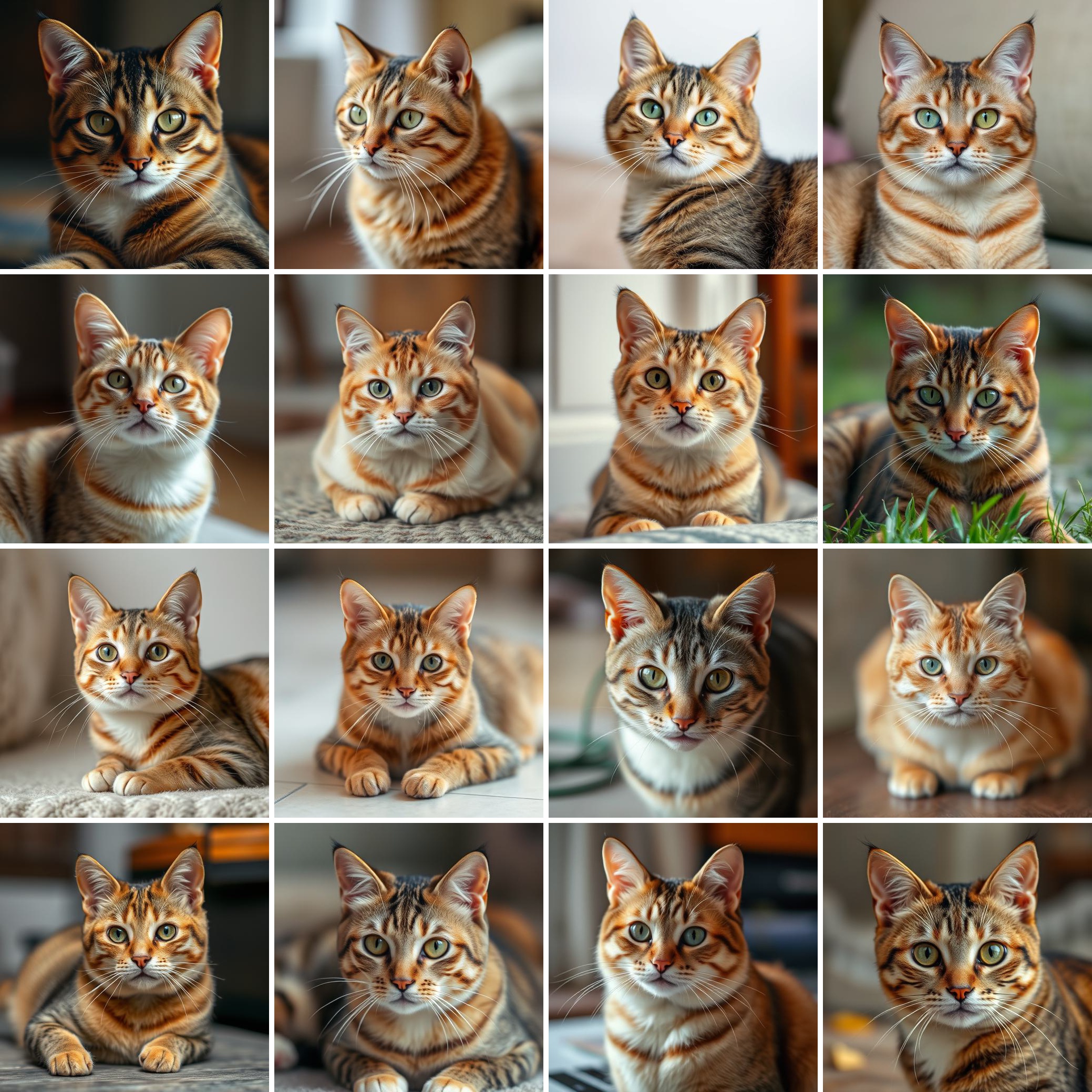} &
        \includegraphics[width=1\linewidth]{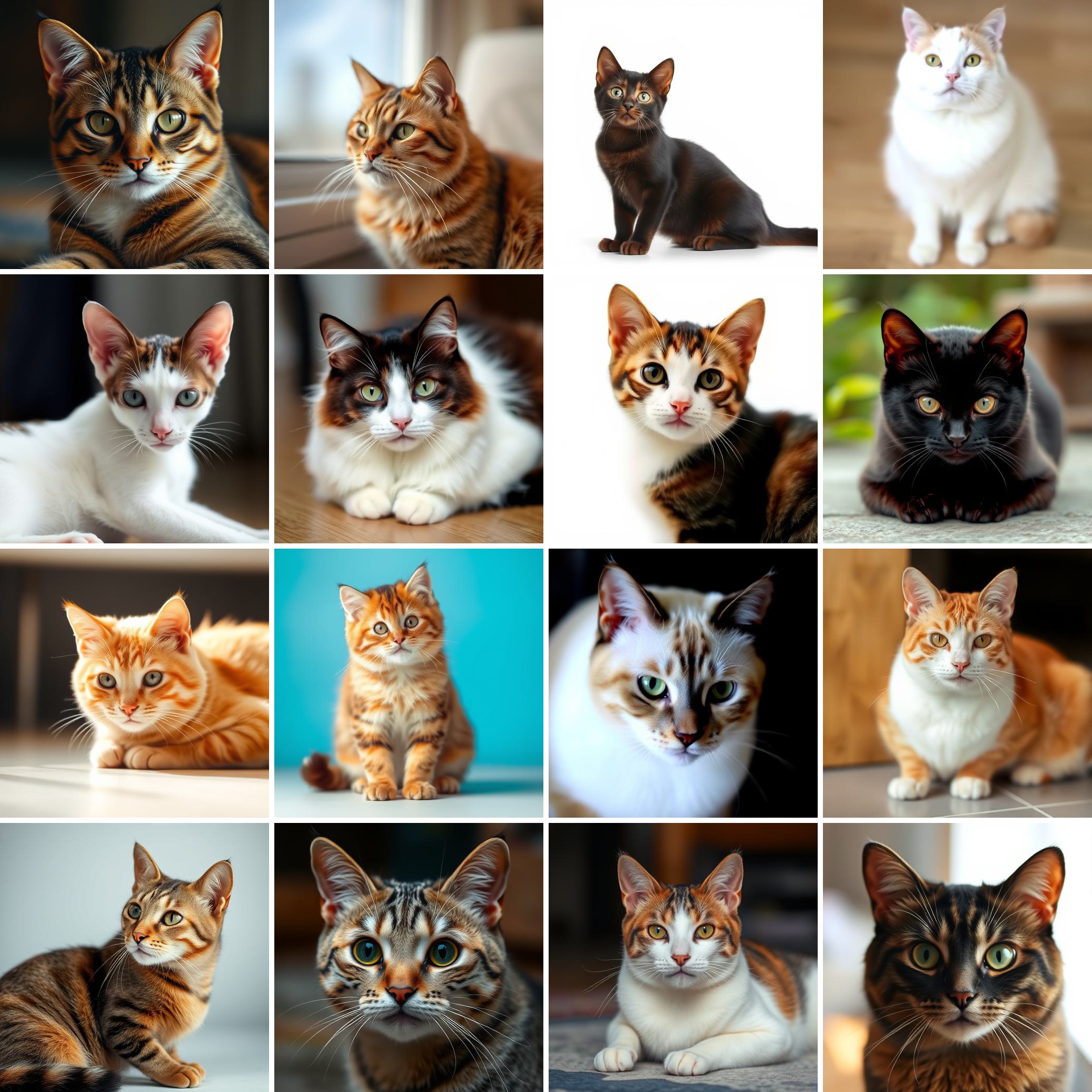} \\ [1.5ex]
    \end{tabular}
    \hfill
    {\textit{\color{blue}{\comicfont \textbf{"A photo of a teddy bear"}}}\\[1ex]}
        \begin{tabular}{p{0.48\textwidth} p{0.48\textwidth}}
        \includegraphics[width=1\linewidth]{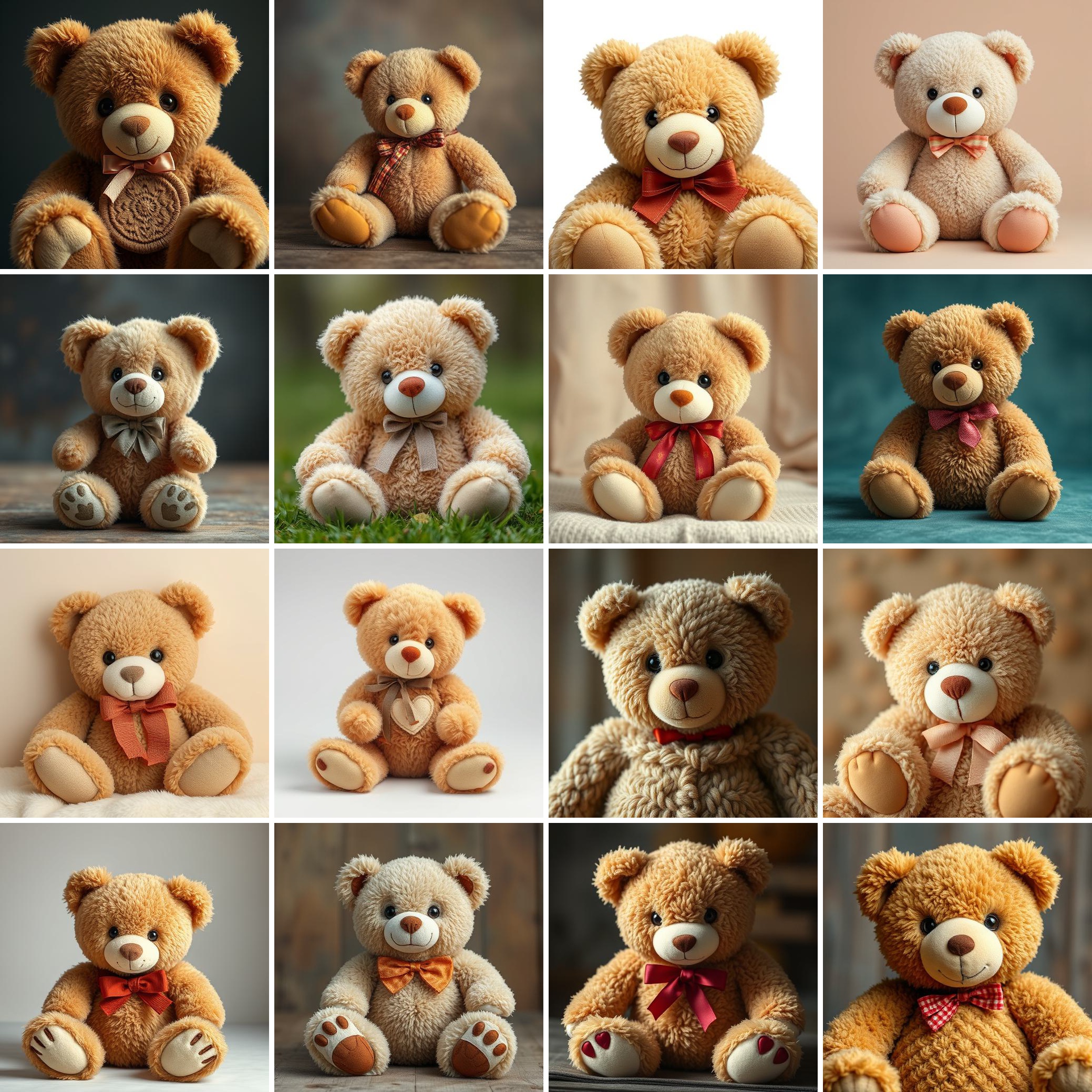} &
        \includegraphics[width=1\linewidth]{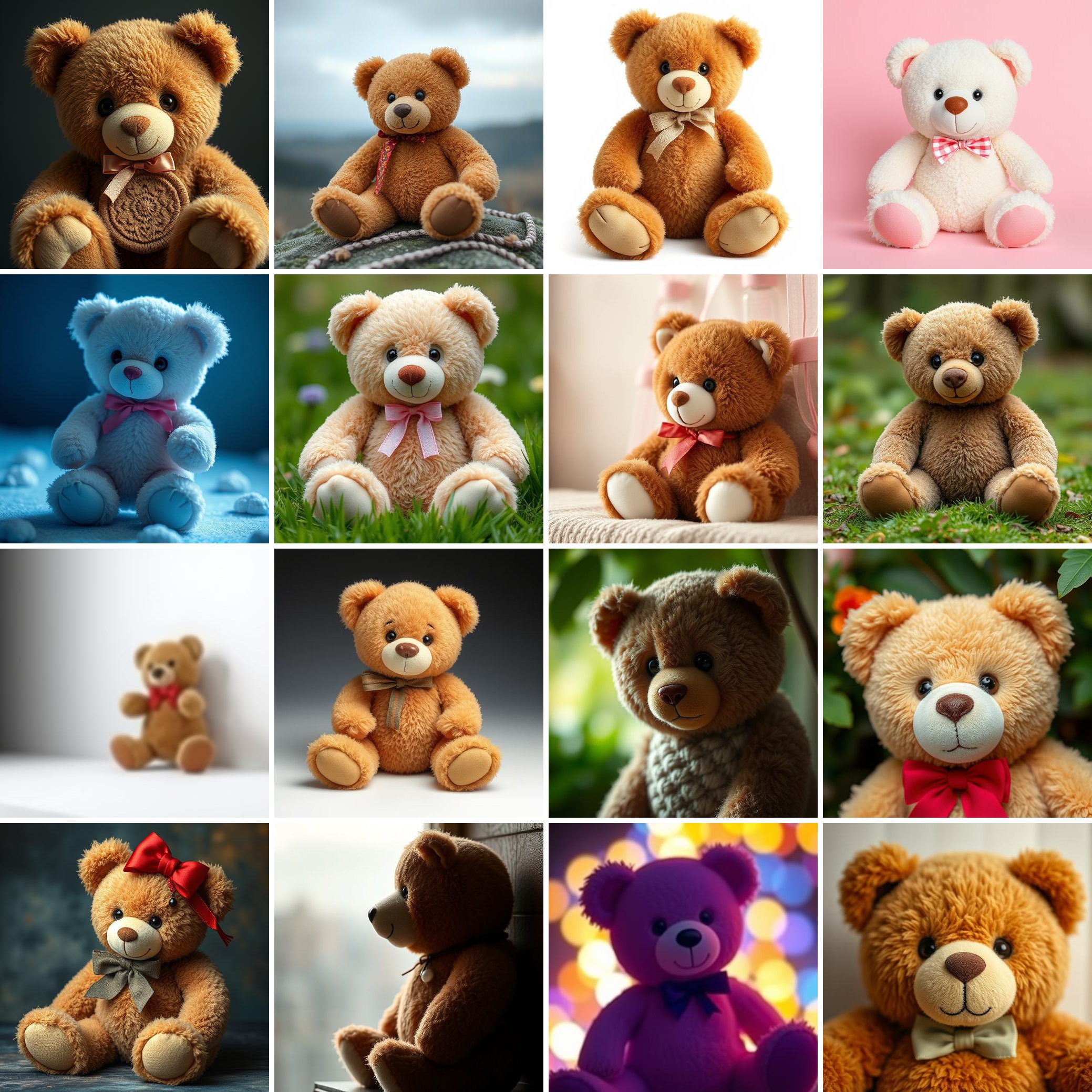} \\
    \end{tabular}
    \hfill
    \begin{tabular}{p{0.48\textwidth} p{0.48\textwidth}}
        \centering $\text{i.i.d.}$ &
        \centering \textbf{Ours} \\
    \end{tabular}
    \vspace{-0.5em}
    \caption{Our method scales to large, diverse image sets via sequential generation. For Flux.1 [schnell], our optimization yields improved diversity of generated image sets compared to $\text{i.i.d.}$ sampling and scales to larger sets such as the 16 shown here. 
    }
    \label{fig:supp_sequential_generation}
\end{figure*}

\begin{figure*}
    \centering
\includegraphics[width=\textwidth]{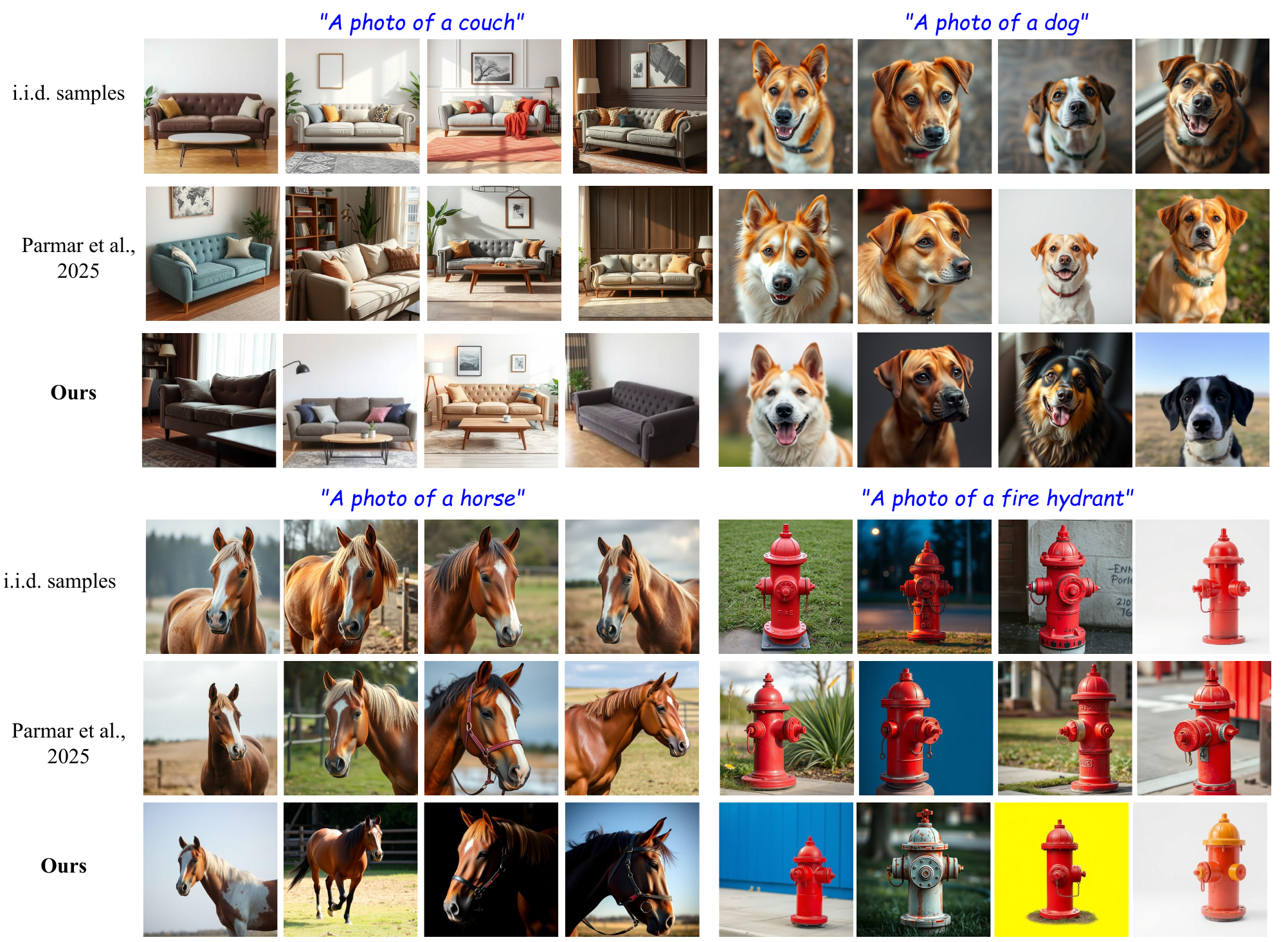}
    \caption{Image generations applying our method to Flux.1 [schnell]~\cite{flux} with white noise initialization. We achieve greater visual diversity compared to baselines while maintaining image quality.
    }
    \vspace{-0.8em}
\label{fig:example_generations_flux}
\end{figure*}

\begin{figure*}
    \centering
\includegraphics[width=0.98\textwidth]{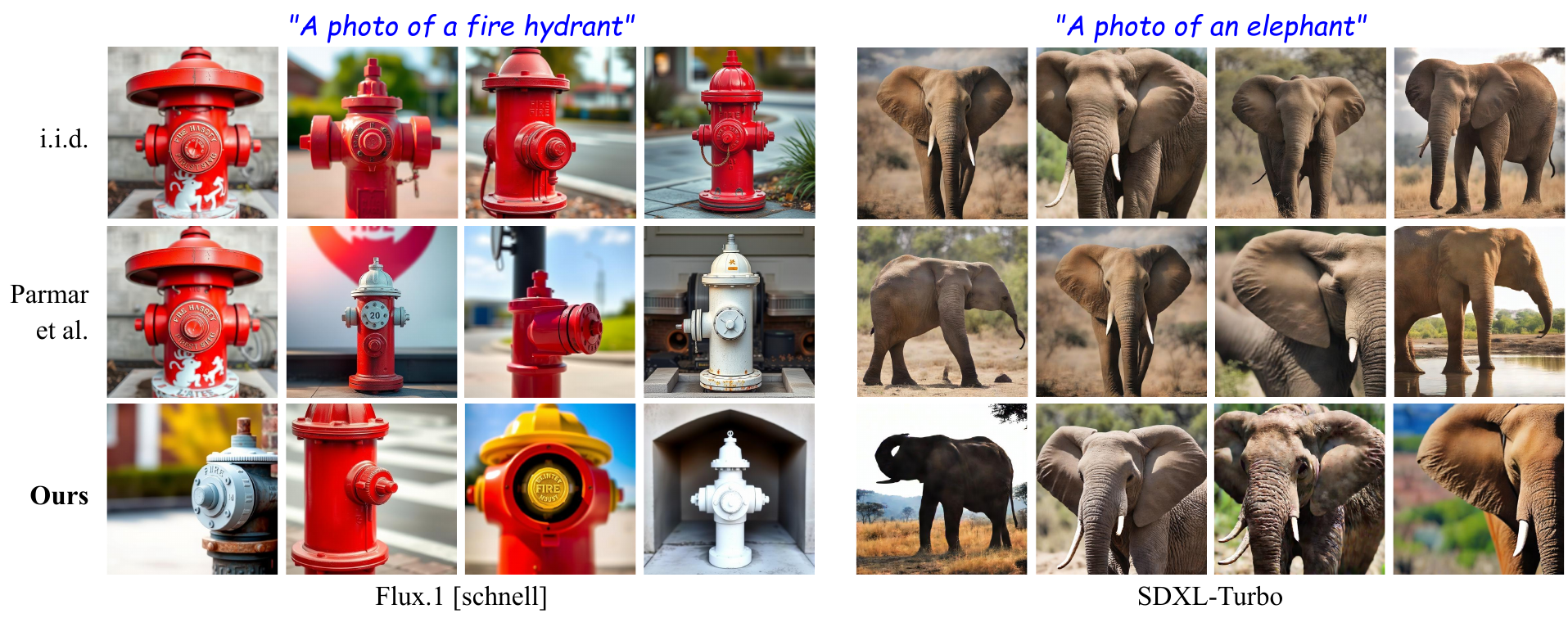}
    \caption{Image generations applying our method to Flux.1 [schnell]~\cite{flux} and SDXL-Turbo~\cite{sdxlturbo} with pink noise initialization. We achieve greater visual diversity compared to baselines while maintaining image quality.
    }
    \vspace{-0.8em}
\label{fig:example_generations_flux_pink}
\end{figure*}

 In \cref{fig:supp_sequential_generation}, we demonstrate that our method can be scaled to larger image sets such as $16$ generations even on larger models like Flux.1 [schnell]. Compared to $\text{i.i.d.}$ sampling, we again see greater diversity across different text prompts. Here, we use 25 iterations, a learning rate of $3.0$, $\lambda_{div}=15$ for the DPP diversity objective, $\lambda_{q}=1$ for a HPSv2 quality reward, and gradient clipping of $0.15$. 

\noindent \textbf{Quantitative Comparisons for Flux.1 [schnell].} For longer complex prompts, we provide quantitative results on DPG-Bench~\cite{hu2024ella}. We evaluate Flux.1 [schnell] in \cref{tab:dpg_table} and again demonstrate that our method improves diversity scores across multiple metrics. We use the same hyperparameters as Tab. 2 in the main paper.

We also provide additional baseline comparisons to guidance-based methods such as Particle Guidance~\cite{particle}, CADS~\cite{sadat2023cads}, and NegToMe~\cite{singh2024negative} (\cref{tab:baselines_table}).
In line with \cite{gi}, we observe that \cite{particle} does not significantly improve diversity. These methods all require multi-step models, so we use Flux.1 [schnell]. ~\cite{sadat2023cads} and \cite{singh2024negative} are more effective, but their quality-diversity trade-off results in worse image quality for higher diversity.

\begin{table}[h]
\centering
\caption{Output diversity on DPG-Bench~\cite{hu2024ella} with Flux.1 [schnell]~\cite{flux} using white initial noise optimized with DPP~\cite{elfeki2019gdpp} diversity objective.}
\vspace{-0.6em}
\resizebox{\columnwidth}{!}{%
\begin{tabular}{lcc|ccc}
\toprule
Method & DreamSim & Vendi (DINO) & HPSv2 & PickScore & FID \\
\midrule
$\text{i.i.d.}$ & 0.197$_{\pm 0.062}$ & 1.787$_{\pm 0.358}$ & 0.278$_{\pm 0.036}$ & 0.217$_{\pm 0.012}$ & 22.458 \\
\ours & 0.285$_{\pm 0.077}$ & 2.319$_{\pm 0.474}$ & 0.270$_{\pm 0.033}$ & 0.215$_{\pm 0.011}$ & 21.468 \\
\bottomrule
\bottomrule
\end{tabular}%
}
\label{tab:dpg_table}
\end{table}

\begin{table*}[t]
\centering
\caption{Impact of different optimization objectives for our pipeline with SDXL-Turbo on GenEval using white noise initializations. Our optimization pipeline does not hurt the overall image quality (HPSv2, CLIPScore, PickScore, FID) across different diversity objectives (the result on the metric that we optimized for is shown in brackets), despite only using a weakly weighted CLIP text-image objective as an additional reward to maintain adherence to the input prompt.}\label{tab:diversity_metrics}
\vspace{-.08in}
\resizebox{\textwidth}{!}{%
\begin{tabular}{lccccccc|cccc}
  \toprule
  \textbf{Objective} & DINO & DreamSim & LPIPS & Color & L2 & DPP & Vendi & HPSv2 & CLIPScore & PickScore & FID \\
      \midrule
      None (init) & 0.588$_{\pm 0.082}$ & 0.249$_{\pm 0.089}$ & 0.643$_{\pm 0.059}$ & 0.094$_{\pm 0.041}$ & 0.279$_{\pm 0.046}$ & 2.104$_{\pm 0.216}$ & 1.999$_{\pm 0.505}$ & 0.263$_{\pm 0.027}$ & 0.335$_{\pm 0.031}$ & 0.224$_{\pm 0.013}$ & 24.515 \\
      \midrule
DINO & (0.892$_{\pm 0.049}$) & 0.476$_{\pm 0.105}$ & 0.799$_{\pm 0.056}$ & 0.165$_{\pm 0.057}$ & 0.436$_{\pm 0.061}$ & 2.678$_{\pm 0.114}$ & 3.652$_{\pm 0.368}$ & 0.260$_{\pm 0.024}$ & 0.347$_{\pm 0.032}$ & 0.219$_{\pm 0.012}$ & 21.802 \\
DreamSim & 0.718$_{\pm 0.083}$ & (0.763$_{\pm 0.245}$) & 0.786$_{\pm 0.082}$ & 0.177$_{\pm 0.068}$ & 0.407$_{\pm 0.079}$ & 2.450$_{\pm 0.218}$ & 2.919$_{\pm 0.613}$ & 0.243$_{\pm 0.027}$ & 0.333$_{\pm 0.028}$ & 0.216$_{\pm 0.013}$ & 22.760 \\
LPIPS & 0.680$_{\pm 0.077}$ & 0.383$_{\pm 0.119}$ & (0.852$_{\pm 0.100}$) & 0.146$_{\pm 0.062}$ & 0.370$_{\pm 0.065}$ & 2.219$_{\pm 0.221}$ & 2.276$_{\pm 0.552}$ & 0.269$_{\pm 0.025}$ & 0.338$_{\pm 0.030}$ & 0.223$_{\pm 0.011}$ & 24.170 \\
Color & 0.661$_{\pm 0.076}$ & 0.401$_{\pm 0.117}$ & 0.726$_{\pm 0.069}$ & (0.376$_{\pm 0.156}$) & 0.408$_{\pm 0.080}$ & 2.241$_{\pm 0.216}$ & 2.330$_{\pm 0.552}$ & 0.259$_{\pm 0.027}$ & 0.346$_{\pm 0.032}$ & 0.215$_{\pm 0.014}$ & 23.756 \\
L2 & 0.684$_{\pm 0.065}$ & 0.362$_{\pm 0.091}$ & 0.768$_{\pm 0.056}$ & 0.145$_{\pm 0.052}$ & (0.492$_{\pm 0.081}$) & 2.237$_{\pm 0.213}$ & 2.318$_{\pm 0.538}$ & 0.268$_{\pm 0.024}$ & 0.335$_{\pm 0.033}$ & 0.208$_{\pm 0.012}$ & 25.686 \\
DPP & 0.787$_{\pm 0.043}$ & 0.477$_{\pm 0.098}$ & 0.778$_{\pm 0.054}$ & 0.170$_{\pm 0.061}$ & 0.444$_{\pm 0.058}$ & (2.772$_{\pm 0.000}$) & 4.000$_{\pm 0.001}$ & 0.261$_{\pm 0.025}$ & 0.368$_{\pm 0.035}$ & 0.219$_{\pm 0.012}$ & 22.062 \\
Vendi & 0.791$_{\pm 0.043}$ & 0.486$_{\pm 0.103}$ & 0.782$_{\pm 0.052}$ & 0.167$_{\pm 0.060}$ & 0.440$_{\pm 0.057}$ & 2.773$_{\pm 0.000}$ & (4.000$_{\pm 0.000}$) & 0.259$_{\pm 0.024}$ & 0.356$_{\pm 0.034}$ & 0.219$_{\pm 0.017}$ & 21.925 \\
  \bottomrule
  \bottomrule
\end{tabular}%
}
\vspace{-10pt}
\end{table*}

\begin{table*}[h]
\centering
\caption{Baseline comparisons to guidance-based methods on GenEval with Flux.1 [schnell]. Methods include Particle Guidance~\cite{particle}, CADS~\cite{sadat2023cads}, and NegToMe~\cite{singh2024negative}.}
\resizebox{0.8\textwidth}{!}{%
\begin{tabular}{lcc|cccc}
\toprule
Method & DreamSim & Vendi (DINO) & HPSv2 & PickScore & CLIPScore & FID \\
\midrule
$\text{i.i.d.}$ & 0.307$_{\pm 0.100}$ & 2.013$_{\pm 0.490}$ & 0.304$_{\pm 0.025}$ & 0.232$_{\pm 0.010}$ & 0.332$_{\pm 0.031}$ & 27.871 \\
Particle Guidance~\cite{particle} & 0.296$_{\pm 0.095}$ & 2.047$_{\pm 0.512}$ &0.304$_{\pm 0.024}$  & 0.231$_{\pm 0.001}$ & 0.331$_{\pm 0.032}$ & 27.450\\ 
Parmar et al.~\cite{gi} & 0.399$_{\pm0.104}$& 2.460$_{\pm 0.573}$ & 0.294$_{\pm 0.022}$  & 0.228$_{\pm0.009}$ & 0.324$_{\pm0.027}$ & 26.570 \\ 
CADS~\cite{sadat2023cads} & 0.363$_{\pm 0.117}$ & 2.365$_{\pm 0.611}$ & 0.295$_{\pm 0.028}$ & 0.228$_{\pm 0.001}$ & 0.323$_{\pm 0.031}$ & 25.570\\ 
NegToMe~\cite{singh2024negative} & 0.385$_{\pm 0.092}$ & 2.355$_{\pm 0.515}$ & 0.291$_{\pm 0.022}$ & 0.227$_{\pm 0.009}$ & 0.328$_{\pm 0.029}$ & 26.090\\
\ours &
0.446$_{\pm 0.116}$ & 2.753$_{\pm 0.587}$ & 0.293$_{\pm 0.025}$ & 0.229$_{\pm 0.009}$ & 0.329$_{\pm 0.029}$ & 26.703 \\
\bottomrule
\bottomrule
\end{tabular}%
}
\label{tab:baselines_table}
\end{table*}

\section{Quality-Diversity Relationship}\label{sec:supp_quality_diversity}

\begin{figure}[t]
    \centering
    \includegraphics[width=\linewidth]{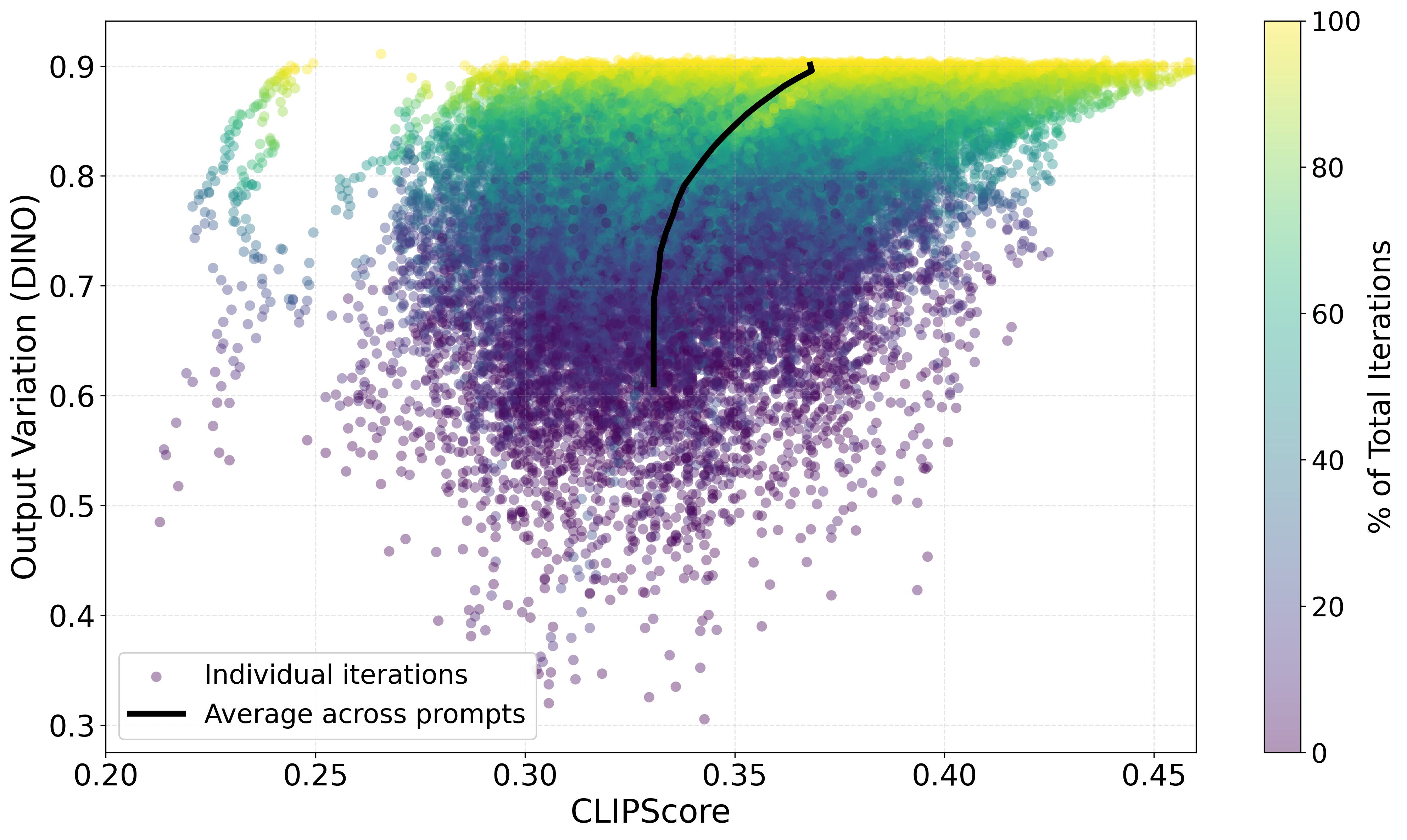}
    \caption{Scatter plot of CLIPScore and DINO diversity during optimization for SDXL-Turbo with white noise initialization on GenEval. Points are colored by iteration progress. The averaged trajectory (black) shows joint improvements in image quality and diversity, demonstrating that our method overcomes the quality–diversity tradeoff.}
    \label{fig:supp_clip_dino_tradeoff}
\end{figure}

The scatter plot in \cref{fig:supp_clip_dino_tradeoff} illustrates the relationship between image quality (measured by CLIPScore) and output diversity (DINO) throughout the optimization process for the white noise configuration on the GenEval dataset. The plot corresponds to the setup used for \cref{fig:scaling_curve}. Note that early stopping terminated optimization after 100 iterations or when the DINO diversity objective reached a threshold of $0.9$.

Each point in the plot represents a single iteration across all prompts, colored by the percentage of total iterations completed (darker points indicate early iterations, lighter points indicate later stages). The black line shows the averaged trajectory across all prompts, revealing that both CLIPScore and DINO diversity increase jointly during optimization. This demonstrates that our approach overcomes the quality-diversity tradeoff described in \cite{gi}. Our improved output variation does not come at the expense of prompt alignment.

\section{Noise Evolution Analysis}
Here, we provide further analysis of the change in noise latents across iterations. In \cref{fig:noise_evolution_raw}, we show the average noise change on the raw noise signal, measured by the L2 norm. The shaded regions around the lines indicate the standard deviation, showing the variability in noise change across different samples.
We observe that the L2 norm increases steadily over iterations for white noise initializations.

The average norm change is slightly lower for pink noise initializations compared to white noise (\cref{fig:noise_evolution_raw}). This confirms that using pink noise as initialization is favorable for our optimization.

\begin{figure}
    \centering
    \includegraphics[width=0.96\linewidth]{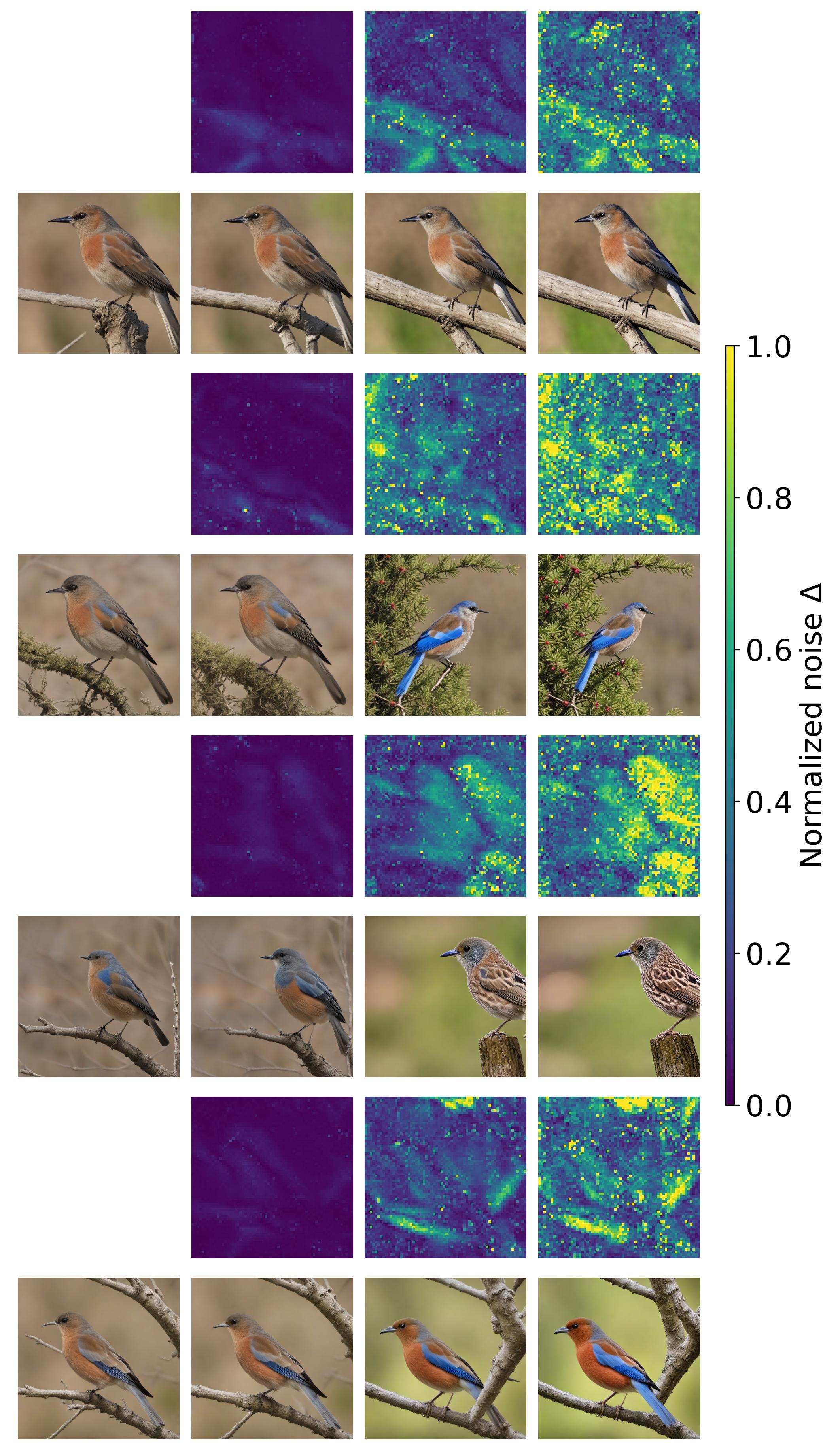}
    \caption{Noise evolution across optimization iterations for a set of four images. We show spatial heatmaps with the averaged L2 difference between the current noise latent and the initial white noise along with the corresponding generated images. Images were generated with SDXL-Turbo and the prompt: ``A photo of a bird''.}
    \label{fig:noise_evolution_bird}
    \vspace{-0.8em}
\end{figure}

\begin{figure}[h!]
    \centering
    {\hspace*{0.7cm}\textit{\color{blue}{\comicfont \textbf{"A photo of a bench"}}}\\[1ex]}
    \includegraphics[trim=45 20 30 135, clip, width=\linewidth]{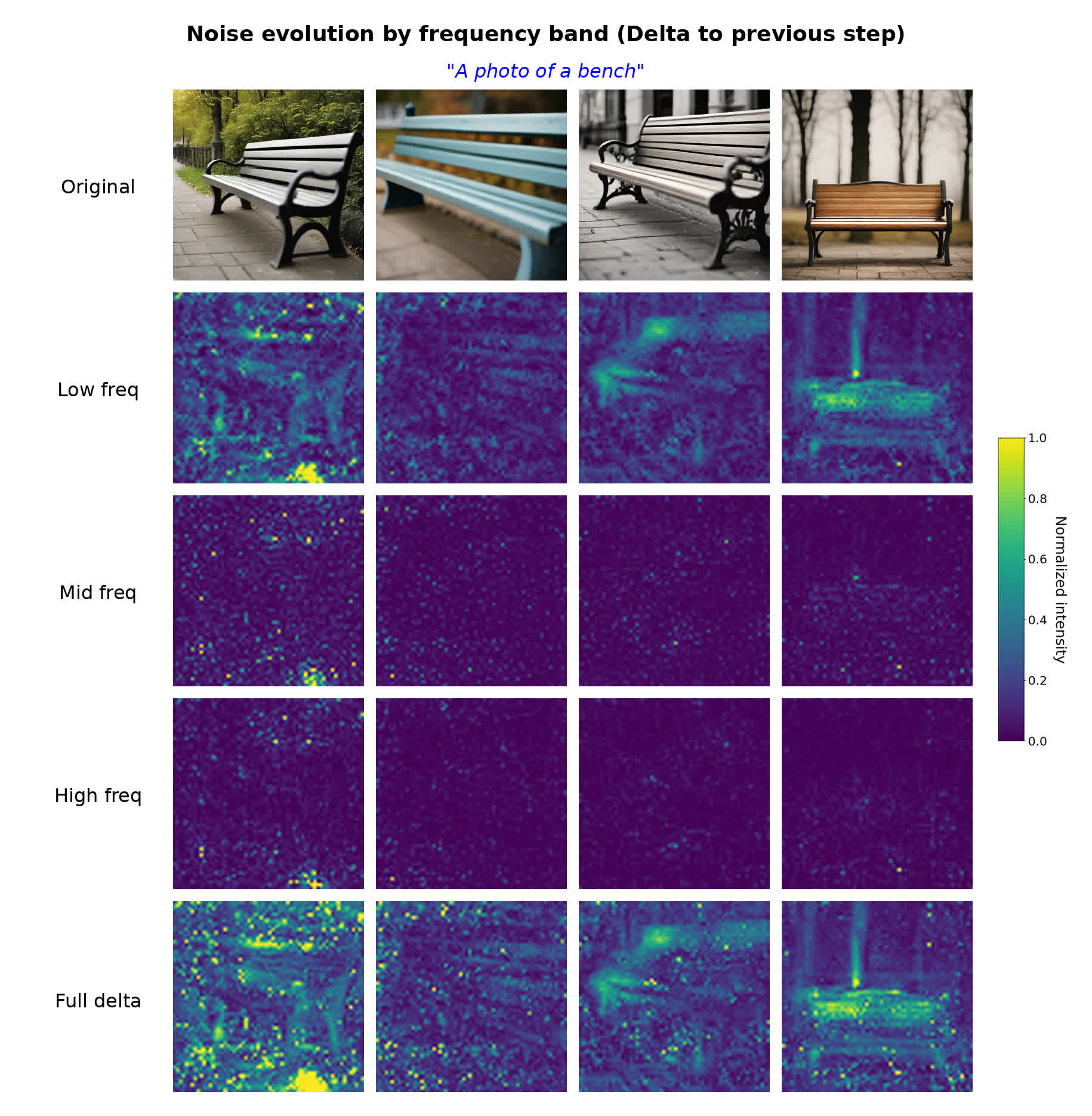}
    \caption{Example showing how the noise changes across optimization iterations in different frequency bands for SDXL-Turbo with white noise initialization and DINO diversity objective. We see that most of the change happens in the lowest third of the frequencies.
    }
    \label{fig:supp_video_frame_example_noise}
\end{figure}

\begin{figure}[t]
        \centering
        \includegraphics[width=\linewidth]{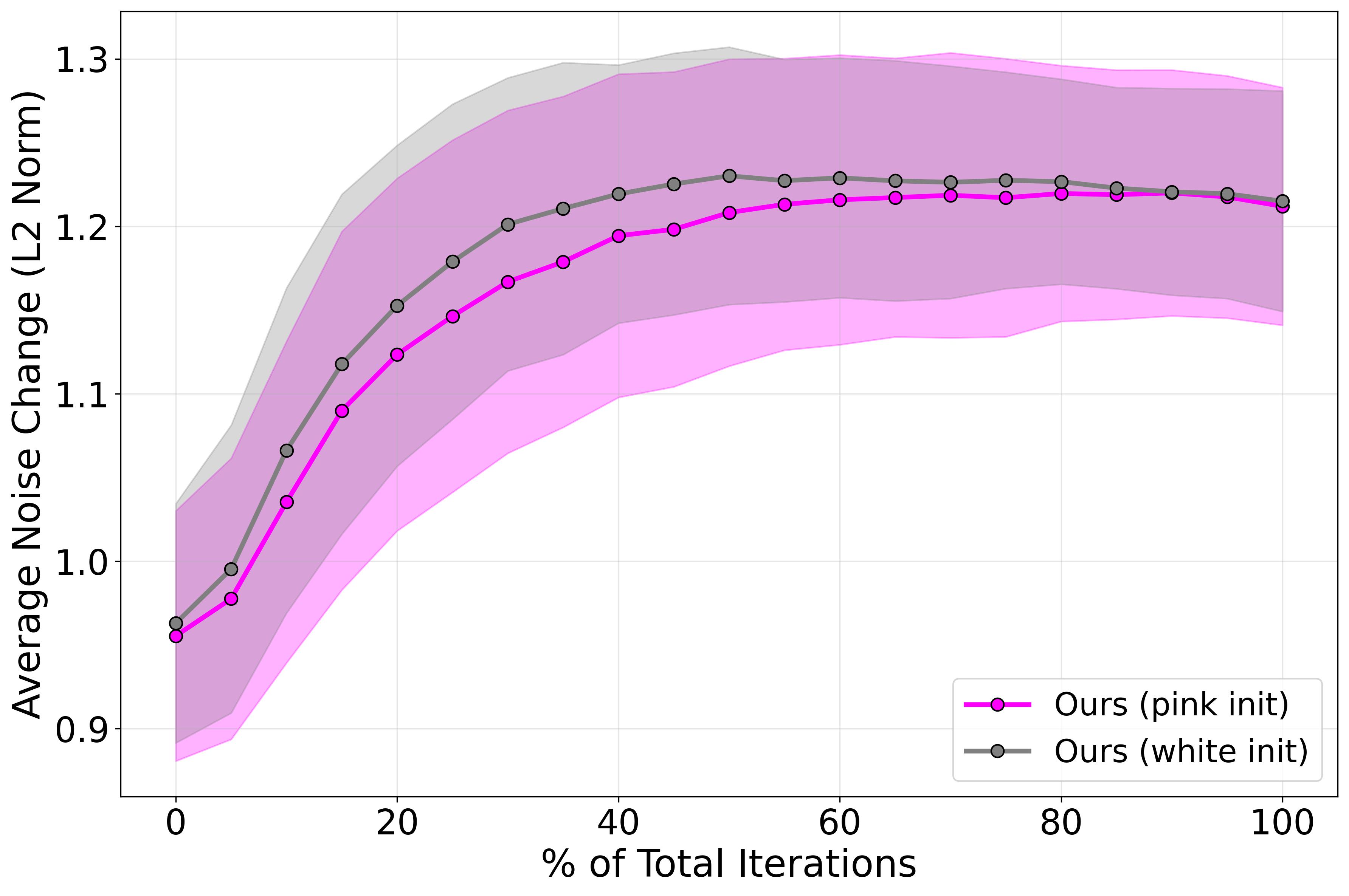}
        \caption{Noise change across iterations on raw noise signal measured as the L2 norm between subsequent iterations. White noise initialization results in slightly higher overall noise change across iterations than pink noise initialization.}
        \label{fig:noise_evolution_raw}
\end{figure}

We also analyze the spatial change in noise, both in general and decomposed into frequency bands (\cref{fig:noise_evolution_bird,fig:supp_video_frame_example_noise}) for SDXL-Turbo. The first column in \cref{fig:noise_evolution_bird} shows the images produced from randomly sampled white noise initializations. Subsequent columns show the intermediate outputs, with the final column displaying the images after optimization. For each iteration, we also visualize a heatmap of the noise change, computed as the averaged L2 difference between the current latent and its initial value. Early in the process the heatmaps remain dark, indicating minimal deviation from the original noise. As optimization proceeds, brighter regions emerge in areas where the noise undergoes substantial modification. These regions align with the parts of the image that change the most (e.g.\ altered bird species or rearranged branches).

Furthermore, we visualize the noise evolution decomposed into frequency bands in \cref{fig:supp_video_frame_example_noise}. This visualization demonstrates that the low frequency components of the noise are being modified most significantly during the optimization process.

\paragraph{Noise Delta Computation.}
For each optimization step $t$, let $\mathbf{z}_t \in \mathbb{R}^{N \times C \times H \times W}$ be the noise.  
We define the noise change as
   $ \Delta \mathbf{z}_t = \mathbf{z}_t - \mathbf{z}_{t-1}$,
with $\mathbf{z}_0$ the initial noise.
To visualize how the noise changes spatially, we compute
\begin{equation}
    M_t(h, w) = \sqrt{\sum_{c=1}^{C} \left( \Delta \mathbf{z}_t \right)_{c,h,w}^2 }.
\end{equation}
This results in a heatmap $M_t \in \mathbb{R}^{H \times W}$ showing the noise change at each location.

\paragraph{Frequency Band Decomposition.}
We decompose $M_t$ into three frequency bands. For this, we compute the 2D FFT:
\begin{equation*}
\mathcal{F}_t(u,v) = \mathcal{F}\{M_t\}, \quad P_t(u,v) = |\mathcal{F}_t(u,v)|^2,
\end{equation*}
where $(u,v)$ are frequency coordinates. The radial distance from the zero-frequency center is
\begin{equation}
r(u,v) = \sqrt{(u-u_c)^2 + (v-v_c)^2}, 
\end{equation}
and we define three frequency bins:
\begin{align*}
\text{Low: }& [0, r_{\max}/3),\\
\text{Mid: }& [r_{\max}/3, 2 r_{\max}/3), \\
\text{High: }& [2 r_{\max}/3, r_{\max}],
\end{align*}
for $r_{\max} = \sqrt{u_c^2 + v_c^2}$.

For each bin $b \in \{\text{low}, \text{mid}, \text{high}\}$, we apply a band-pass mask to the power spectrum:
\begin{equation}
P_t^{(b)}(u,v) = P_t(u,v) \cdot \mathcal{M}_b(u,v),
\end{equation}
and scale the original FFT to preserve phase:
\begin{equation}
\mathcal{F}_t^{(b)}(u,v) = \mathcal{F}_t(u,v) \cdot \sqrt{\frac{P_t^{(b)}(u,v)}{P_t(u,v) + \epsilon}}, \quad \epsilon = 10^{-10}.
\end{equation}
The spatial representation is obtained via the inverse FFT:
\begin{equation}
M_t^{(b)}(h,w) = \left|\mathcal{F}^{-1}\{\mathcal{F}_t^{(b)}\}\right|.
\end{equation}
We then normalize, so the frequency bands sum to the full magnitude:
\begin{equation}
    \tilde{M}_t^{(b)}(h,w) = M_t^{(b)}(h,w) \cdot \frac{M_t(h,w)}{\sum_{b'} M_t^{(b')}(h,w) + \epsilon}.
\end{equation}
This ensures $\sum_{b'} \tilde{M}_t^{(b')} = M_t$ at each pixel.

\paragraph{Visual Observations.} The noise evolution visualization confirms that most noise change happens in the low-frequency components. These changes directly correspond to spatial changes in the generations throughout the optimization steps. 
This observation along with the fact that natural images have a 1/f power spectrum inspires our exploration of noise initializations with stronger low-frequency components (e.g.\ pink noise).

\subsection{Pink Noise Example Generations}\label{sec:supp_noise_ablation}

\begin{figure*}[t]
    \centering
    {\textit{\color{blue}{\comicfont \textbf{"A photo of a bear"}}}\\[1ex]}
    \includegraphics[width=\linewidth]{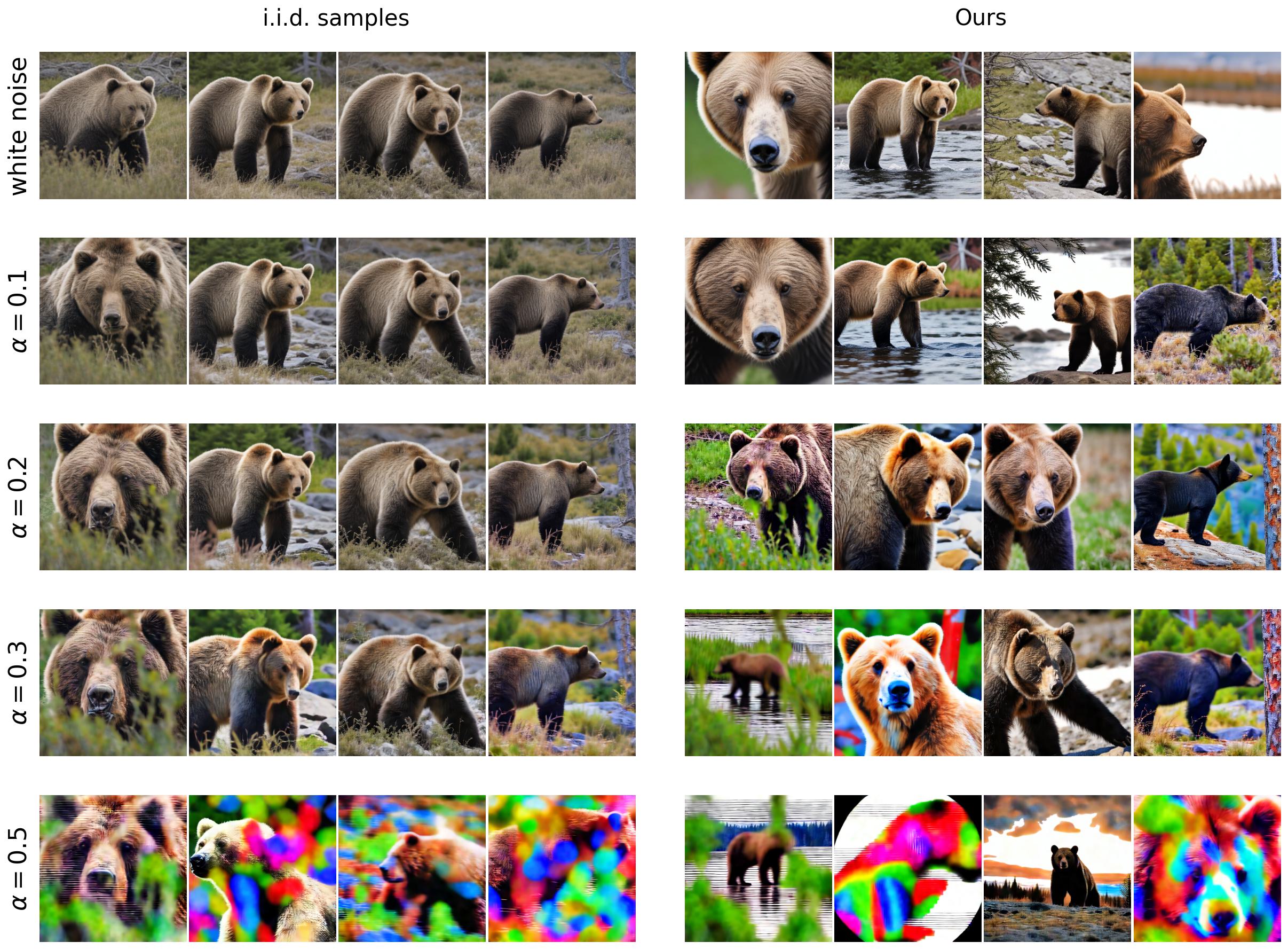}
    \caption{Effect of noise exponent values on image generation. Each row compares $\text{i.i.d.}$ samples from initial noise (left) with our outputs (right) for different $\alpha$ values. Results were obtained with SDXL-Turbo and noise optimization using DINO diversity and CLIPScore.}
    \label{fig:supp_noise_example_bear}
\end{figure*}

Higher $\alpha$ values (see main paper Eq. 6) generally lead to higher diversity scores. However, the image quality decreases with high noise exponents (see generations for $\alpha=0.3$ and $\alpha=0.5$ in \cref{fig:supp_noise_example_bear} which have patchy artefacts). Note that we use CLIPScore as the only image quality reward during optimization for \cref{tab:supp_pink_table}. However, additional rewards for image quality can easily be included in our pipeline. 

In our experiments, we use a noise exponent of 
$0.2$ (referred to as pink noise), which provides substantial gains in sample diversity and reduces the number of required iterations, while preserving image quality.

\section{User Study}\label{sec:supp_user_study}

\begin{table}[t]
\caption{Human preference win rates from a user study for our method against $\text{i.i.d.}$ sampling and \citet{gi} for PixArt-$\alpha$~\cite{pixartalpha}, SANA-Sprint-1.6B~\cite{sanasprint}, and SDXL-Turbo~\cite{sdxlturbo}.}
\centering
\resizebox{\linewidth}{!}{%
\begin{tabular}{lcc}
\toprule
\textbf{Method} & \textbf{Win \% vs i.i.d.} & \textbf{Win \% vs \cite{gi}} \\
\midrule
PixArt-$\alpha$~\cite{pixartalpha}       & 90.00 & 77.50 \\
SANA-Sprint-1.6B~\cite{sanasprint}      & 85.00 & 66.25 \\
SDXL-Turbo~\cite{sdxlturbo}            & 88.75 & 91.25 \\
\bottomrule
\bottomrule
\end{tabular}
}
\label{tab:win_rates}
\end{table}

\begin{table}[t]
\caption{Output diversity results on the single-object subset of GenEval for our proposed approach with the PixArt-$\alpha$, SANA-Sprint-1.6B, and SDXL-Turbo models using white noise initialization. Output diversity is measured with averaged pairwise DINO, DreamSim, and LPIPS scores.}
\centering
\resizebox{\linewidth}{!}{%
\begin{tabular}{lccc}
\toprule
\textbf{Method} &  DINO & DreamSim & LPIPS \\
\midrule
\textbf{PixArt-$\alpha$}~\cite{pixartalpha} & & &  \\
\midrule
$\text{i.i.d.}$ & 0.382$_{\pm 0.093}$ & 0.160$_{\pm 0.078}$ & 0.460$_{\pm 0.126}$  \\
\citet{gi} & 0.520$_{\pm 0.093}$ & 0.227$_{\pm 0.094}$ & 0.563$_{\pm 0.116}$\\
\ours & 0.731$_{\pm 0.077}$ & 0.370$_{\pm 0.117}$ & 0.691$_{\pm 0.096}$  \\
\midrule
\textbf{SANA-Sprint-1.6B}~\cite{sanasprint} &  &  &  \\
\midrule
 $\text{i.i.d.}$ &  0.494$_{\pm 0.091}$ & 0.219$_{\pm 0.081}$ & 0.631$_{\pm 0.070}$ \\
 \citet{gi} & 0.695$_{\pm 0.061}$ & 0.363$_{\pm 0.112}$ & 0.733$_{\pm 0.052}$\\
\ours & 0.752$_{\pm 0.065}$ & 0.485$_{\pm 0.109}$ & 0.795$_{\pm 0.058}$ \\
\midrule
\textbf{SDXL-Turbo}~\cite{sdxlturbo} & & &  \\
\midrule
$\text{i.i.d.}$ & 0.529$_{\pm 0.077}$ & 0.218$_{\pm 0.089}$ & 0.611$_{\pm 0.058}$  \\
\citet{gi} & 0.667$_{\pm 0.069}$ & 0.320$_{\pm 0.118}$ & 0.661$_{\pm 0.053}$\\
\ours & 0.808$_{\pm 0.047}$ & 0.450$_{\pm 0.131}$ & 0.768$_{\pm 0.046}$ \\
\bottomrule
\bottomrule
\end{tabular}
}
\label{tab:geneval_80_div}
\end{table}
We conduct a human user preference study to determine which methods produce more diverse outputs, similar to ~\citet{gi}. We compare our method to baselines such as $\text{i.i.d.}$ sampling and \citet{gi}, as well as across different target diversity objectives.

During the study, we show participants a 2x2 grid of images generated from our method and a comparison. We ask the user to select ``which grid of images has higher variety?''. For each pairing, we collect 10 user preferences to determine a per prompt win rate. User data is anonymized and crowdsourced.

We run trials across all single-object prompts in the GenEval benchmark~\cite{ghosh2023geneval} (prompts 1 to 80). For reference, we also report diversity scores for this subset in \cref{tab:geneval_80_div}. We count the number of wins across trials for each model to compute a final overall win percentage.
In the results in \cref{tab:win_rates}, we observe that our method shows the highest win rate across all three models. 

In addition, we compared our method across different diversity objectives (see main paper Fig.~6).

\begin{figure*}[h!]
    \centering
\hspace*{0.42cm}\begin{tabular}{p{0.463\textwidth} p{0.463\textwidth}}
        \centering $\text{i.i.d.}$ samples &
        \centering Ours \\
    \end{tabular}
   {\hspace*{0.7cm}\textit{\color{blue}{\comicfont \textbf{"A photo of a bench"}}}\\[1ex]}
    \includegraphics[trim=5 10 20 40, clip, width=0.87\textwidth]{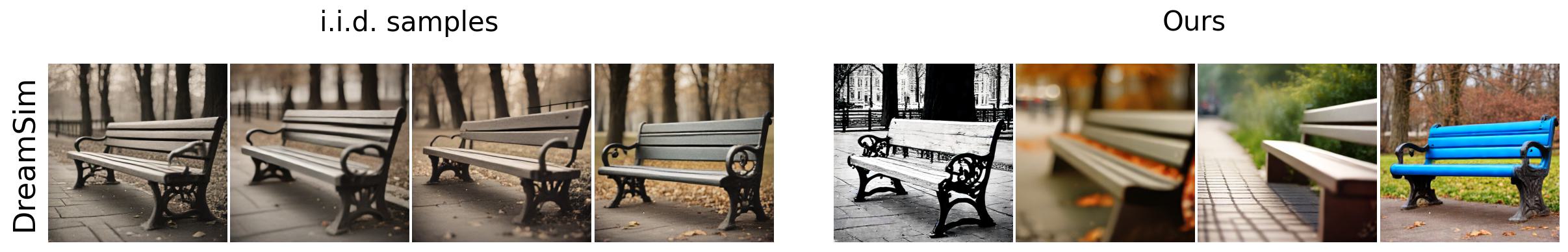}
    
       {\hspace*{0.7cm}\textit{\color{blue}{\comicfont \textbf{"A photo of a backpack"}}}\\[1ex]}
    \includegraphics[trim=5 10 20 0, clip, width=0.87\textwidth]{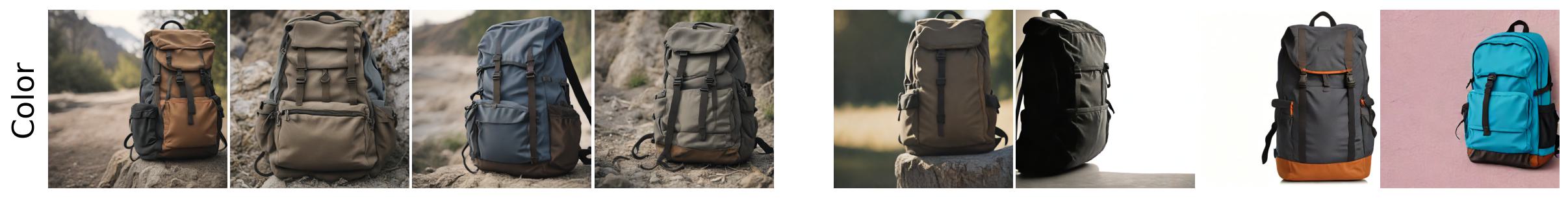}
    
       {\hspace*{0.7cm}\textit{\color{blue}{\comicfont \textbf{"A photo of a surfboard"}}}\\[1ex]}
    \includegraphics[trim=5 10 20 0, clip, width=0.87\textwidth]{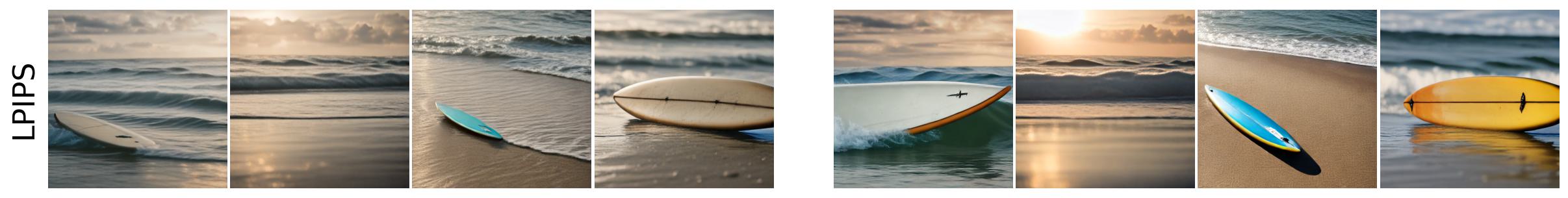}
    \caption{Failure cases of our method for different optimization objectives (SDXL-Turbo). Top row: Removing fine details through blurring one image increases perceptual distance without introducing meaningful diversity. Middle row: Overly simple compositions (e.g.\ plain backgrounds) lead to high color diversity scores as different solid colors maximize L2 color histogram distance effectively. Bottom row: LPIPS optimization fails to recover semantic content that is missing in the generation from the initial noise.}
    \label{fig:supp_failure_cases}
\end{figure*}

\section{Failure Cases}
Despite the effectiveness of our optimization approach, several failure modes can be observed. We visualize these in \cref{fig:supp_failure_cases}. When using DreamSim, the optimization sometimes produces blurry images as the method exploits perceptual distance which can remove high-frequency details (top row). 
Color histogram diversity tends to encourage plain backgrounds since uniform color regions efficiently maximize histogram L2 distances. LPIPS diversity exhibits a critical limitation: it does not recover semantic content missing from the initial noise visualization (e.g., if a surfboard is not generated at first, it remains absent), as LPIPS diversifies existing perceptual features rather than introducing new semantic elements. This could be recovered with a larger weighting of image quality and prompt adherence rewards in the optimization process.

\begin{figure*}[t]
    \centering
        \begin{minipage}{0.48\textwidth}
        \centering
    {\textit{\color{blue}{\comicfont \textbf{"A photo of a horse"}}}\\[1ex]}
        \includegraphics[width=\textwidth]{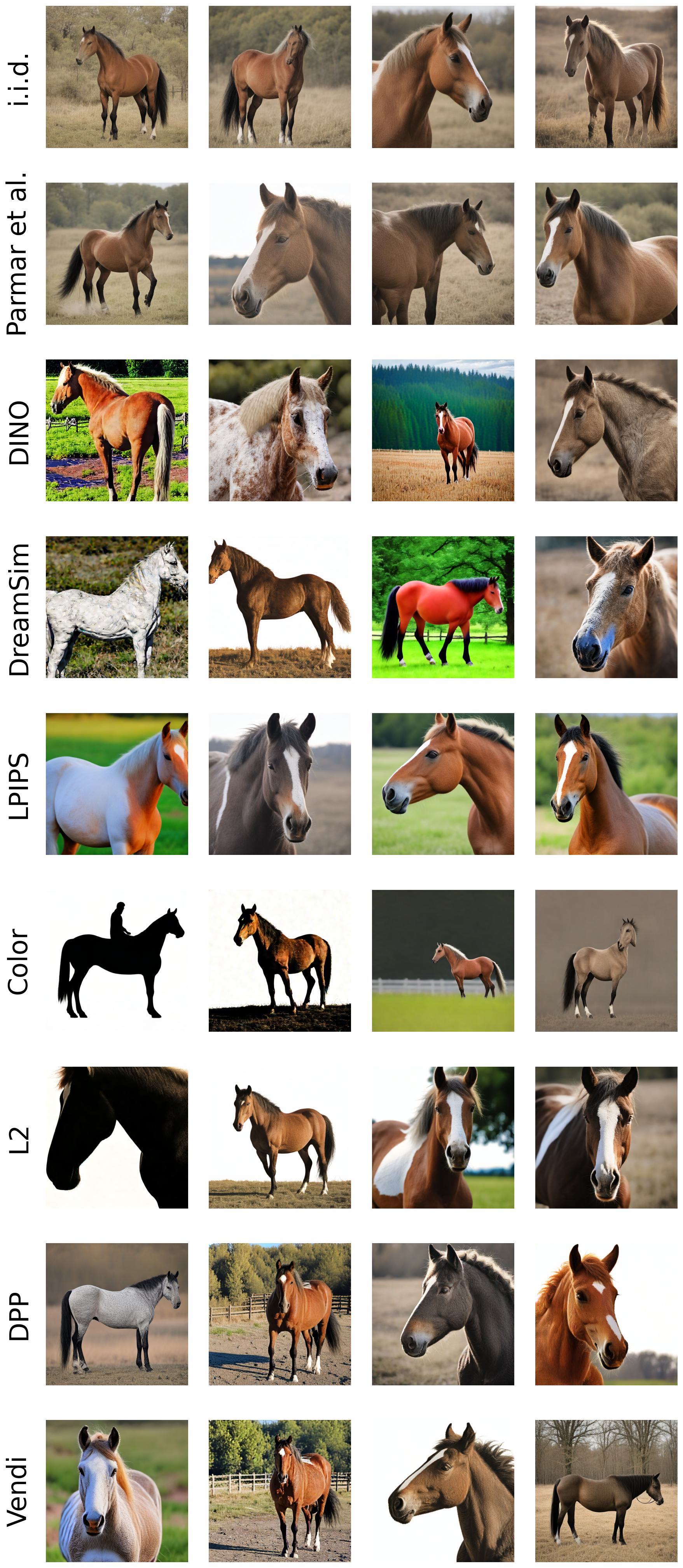}
    \end{minipage}
    \hfill
    \begin{minipage}{0.48\textwidth}
        \centering
 {\textit{\color{blue}{\comicfont \textbf{"A photo of a fire hydrant"}}}\\[1ex]}
        \includegraphics[width=\textwidth]{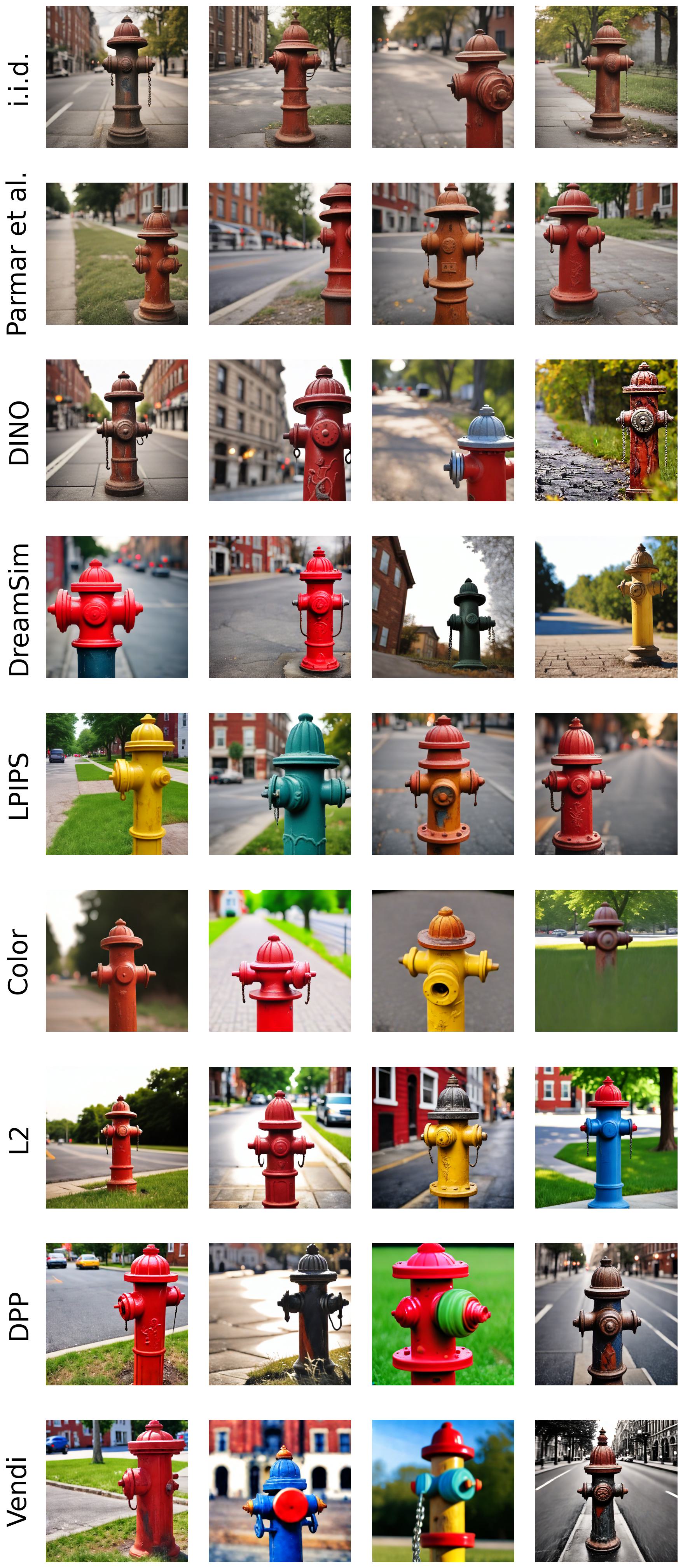}
    \end{minipage}

    \caption{Impact of diversity objectives on the resulting noise optimization and image generations compared to $\text{i.i.d.}$ sampled noise initialization and the search method proposed by \citet{gi}. Our approach results in more varied generations in terms of object pose, appearance, colors, and backgrounds (e.g. different horse breeds in different surroundings, and fire hydrants in different colors).}
    \label{fig:supp_objectives_1}
\end{figure*}

\begin{figure*}[t]
    \centering
        \begin{minipage}{0.48\textwidth}
        \centering
 {\textit{\color{blue}{\comicfont \textbf{"A photo of a handbag"}}}\\[1ex]}
        \includegraphics[width=\textwidth]{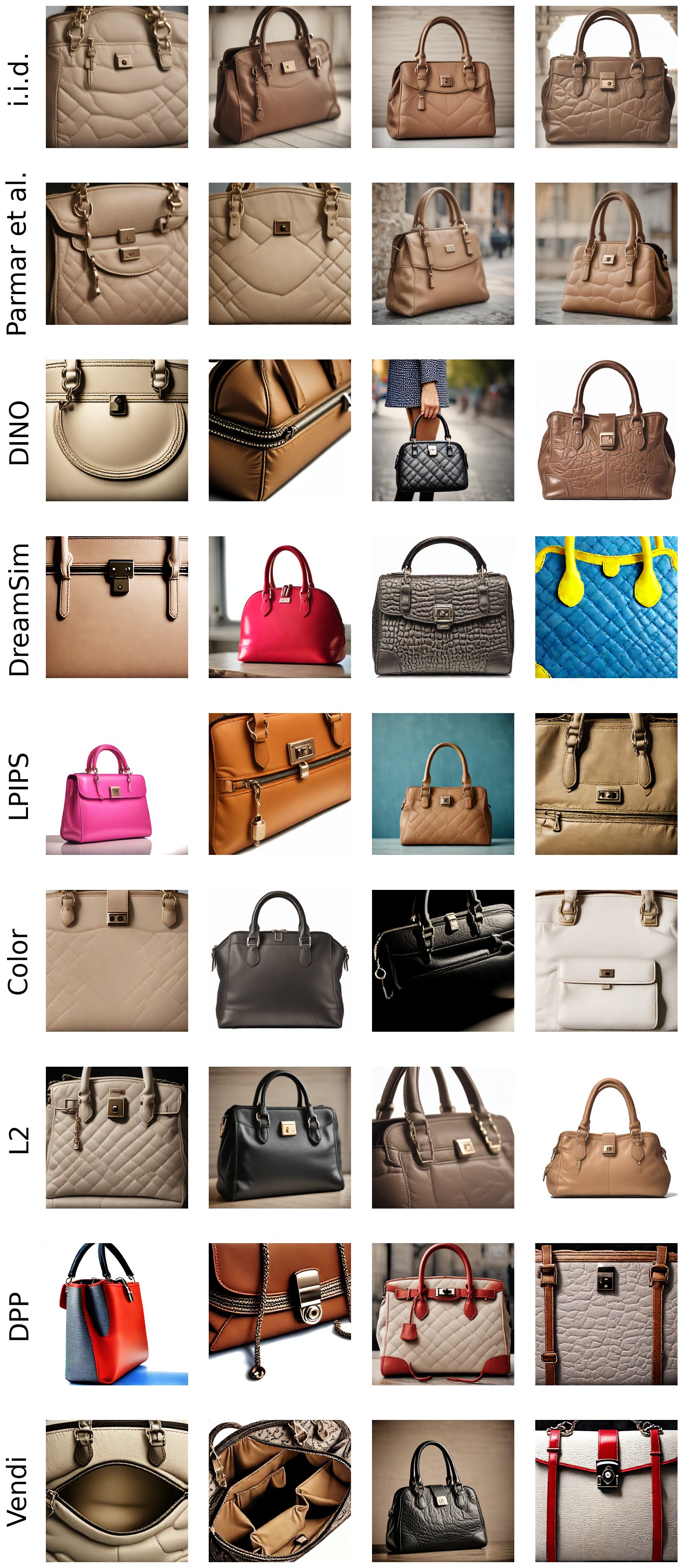}
    \end{minipage}
    \hfill
    \begin{minipage}{0.48\textwidth}
        \centering
         {\textit{\color{blue}{\comicfont \textbf{"A photo of a potted plant"}}}\\[1ex]}
        \includegraphics[width=\textwidth]{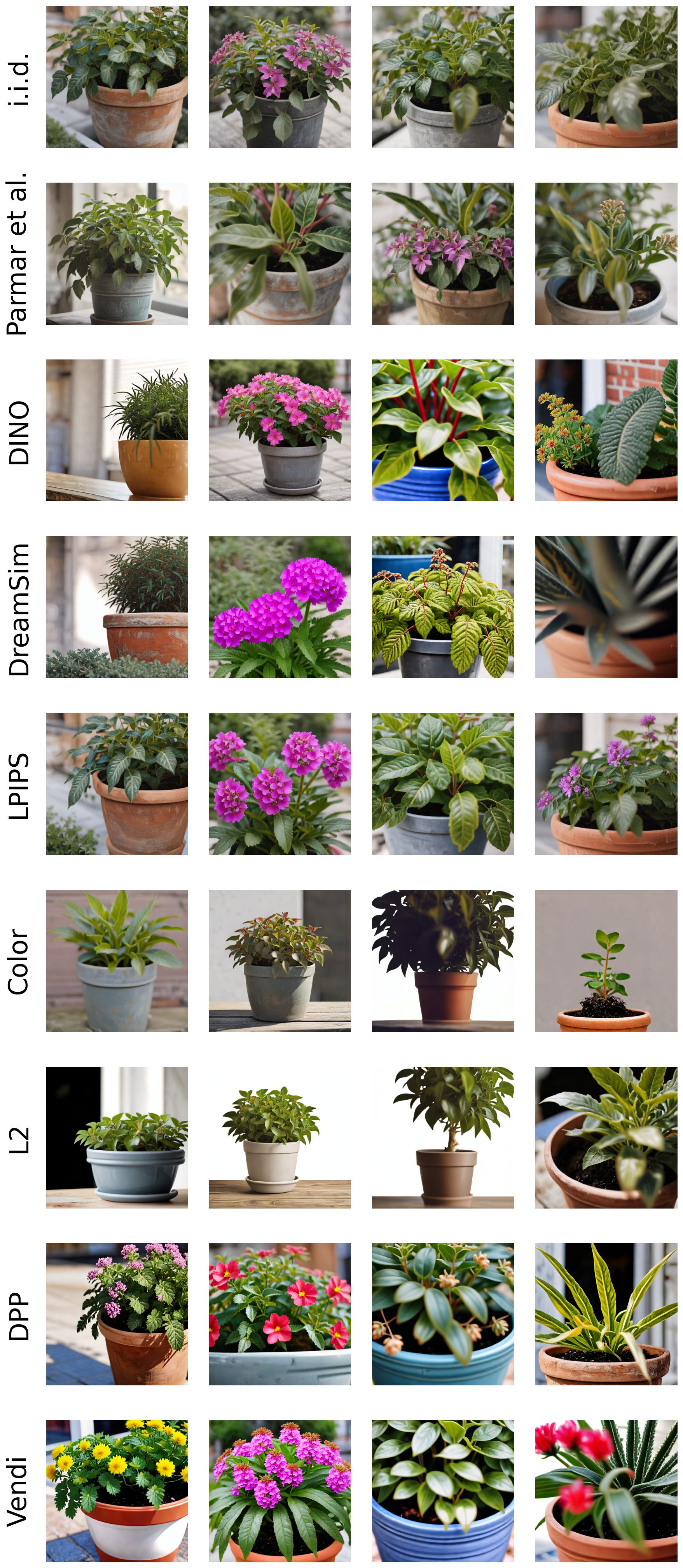}
    \end{minipage}

    \caption{Impact of diversity objectives on the resulting noise optimization and image generations compared to $\text{i.i.d.}$ sampled noise initialization and the search method proposed by \citet{gi}. The generated handbags and potted plants show larger variation in terms of handbag types and colors, and plant species.}
    \label{fig:supp_objectives_2}
\end{figure*}

\end{document}